%% file: main.tex
\documentclass[10pt,twocolumn,letterpaper]{article}

\usepackage{iccv}
\usepackage{times}
\usepackage{epsfig}
\usepackage{graphicx}
\usepackage{amsmath}
\usepackage{amssymb}

\usepackage{mathtools} 
\usepackage{bm}
\usepackage{algorithm}
\usepackage{algorithmic}
\usepackage{listings}
\usepackage{makecell}  % \Xhline
\usepackage{diagbox} % diagonal table cell
\usepackage{graphicx} % rotated text
\usepackage{multirow}
\usepackage{float}
\usepackage{subfig}
\usepackage{bbding}

\usepackage[table,dvipsnames]{xcolor}

\definecolor{detcolor}{gray}{.9}

\definecolor{bestcolor}{gray}{.9}
\newcommand{\bestcell}[1]{\cellcolor{bestcolor}{#1}}

\newcommand{\tablestyle}[2]{\setlength{\tabcolsep}{#1}\renewcommand{\arraystretch}{#2}\centering\footnotesize}
\newlength\savewidth\newcommand\shline{\noalign{\global\savewidth\arrayrulewidth
  \global\arrayrulewidth 1pt}\hline\noalign{\global\arrayrulewidth\savewidth}}

\newcolumntype{x}[1]{>{\centering\arraybackslash}p{#1pt}}
\newcolumntype{y}[1]{>{\raggedright\arraybackslash}p{#1pt}}
\newcolumntype{z}[1]{>{\raggedleft\arraybackslash}p{#1pt}}
\renewcommand{\paragraph}[1]{\vspace{1.25mm}\noindent\textbf{#1}}
\definecolor{deemph}{gray}{0.6}

\usepackage{etoolbox}
\makeatletter
\AfterEndEnvironment{algorithm}{\let\@algcomment\relax}
\AtEndEnvironment{algorithm}{\kern2pt\hrule\relax\vskip3pt\@algcomment}
\let\@algcomment\relax
\newcommand\algcomment[1]{\def\@algcomment{\footnotesize#1}}
\renewcommand\fs@ruled{\def\@fs@cfont{\bfseries}\let\@fs@capt\floatc@ruled
  \def\@fs@pre{\hrule height.8pt depth0pt \kern2pt}%
  \def\@fs@post{}%
  \def\@fs@mid{\kern2pt\hrule\kern2pt}%
  \let\@fs@iftopcapt\iftrue}
\makeatother
\usepackage[pagebackref=true,breaklinks=true,letterpaper=true,colorlinks,bookmarks=false]{hyperref}

\iccvfinalcopy % *** Uncomment this line for the final submission

\def\iccvPaperID{3887} % *** Enter the ICCV Paper ID here

\begin{document}

%%%%%%%%% TITLE
\title{Diffusion-Based 3D Human Pose Estimation with Multi-Hypothesis Aggregation}

\author{Wenkang Shan$^{1}$ \quad
Zhenhua Liu$^{2}$ \quad
Xinfeng Zhang$^{3}$ \\ Zhao Wang$^{1}$ \quad Kai Han$^{2}$ \quad Shanshe Wang$^{1}$ \quad Siwei Ma$^{1}$ \quad Wen Gao$^{1}$\\
$^{1}$National Engineering Research Center of Visual Technology, Peking University\\
$^{2}$Huawei Noah’s Ark Lab \quad
$^{3}$University of Chinese Academy of Sciences\\
{\tt\small \{wkshan,zhaowang,sswang,swma,wgao\}@pku.edu.cn}
\\
{\tt\small\{liu.zhenhua,kai.han\}@huawei.com \quad xfzhang@ucas.ac.cn}}

% \and

\maketitle
%\ificcvfinal\thispagestyle{empty}\fi

%%%%%%%%% ABSTRACT
\begin{abstract}
In this paper, a novel Diffusion-based 3D Pose estimation (D3DP) method with Joint-wise reProjection-based Multi-hypothesis Aggregation (JPMA) is proposed for probabilistic 3D human pose estimation. On the one hand, D3DP generates multiple possible 3D pose hypotheses for a single 2D observation. It gradually diffuses the ground truth 3D poses to a random distribution, and learns a denoiser conditioned on 2D keypoints to recover the uncontaminated 3D poses. The proposed D3DP is compatible with existing 3D pose estimators and supports users to balance efficiency and accuracy during inference through two customizable parameters. On the other hand, JPMA is proposed to assemble multiple hypotheses generated by D3DP into a single 3D pose for practical use. It reprojects 3D pose hypotheses to the 2D camera plane, selects the best hypothesis joint-by-joint based on the reprojection errors, and combines the selected joints into the final pose. The proposed JPMA conducts aggregation at the joint level and makes use of the 2D prior information, both of which have been overlooked by previous approaches. Extensive experiments on Human3.6M and MPI-INF-3DHP datasets show that our method outperforms the state-of-the-art deterministic and probabilistic approaches by 1.5\% and 8.9\%, respectively. Code is available at \url{https://github.com/paTRICK-swk/D3DP}.

%The proposed method has the advantage of high flexibility, allowing customizing the parameters in the inference phase to balance performance and speed, avoiding complex loss function designs.
\end{abstract}

%%%%%%%%% BODY TEXT
\section{Introduction}

Monocular 3D human pose estimation aims to locate the 3D positions of human body joints from 2D images or videos. It plays a crucial role in various applications, such as human-computer interaction, metaverse, and self-driving. This task can be decomposed into two steps: first estimating the 2D locations of human joints using off-the-shelf 2D keypoint detectors (\eg, CPN~\cite{chen2018cascaded}, OpenPose~\cite{cao2017realtime}); and then mapping these 2D locations to their corresponding 3D positions. In this work, we focus on the latter step, also known as the 2D-to-3D lifting process, following recent approaches\cite{ci2019optimizing, pavllo20193d, zhang2022mixste, shan2022p}.

\begin{figure}[t]
\begin{center}
\includegraphics[width=\linewidth]{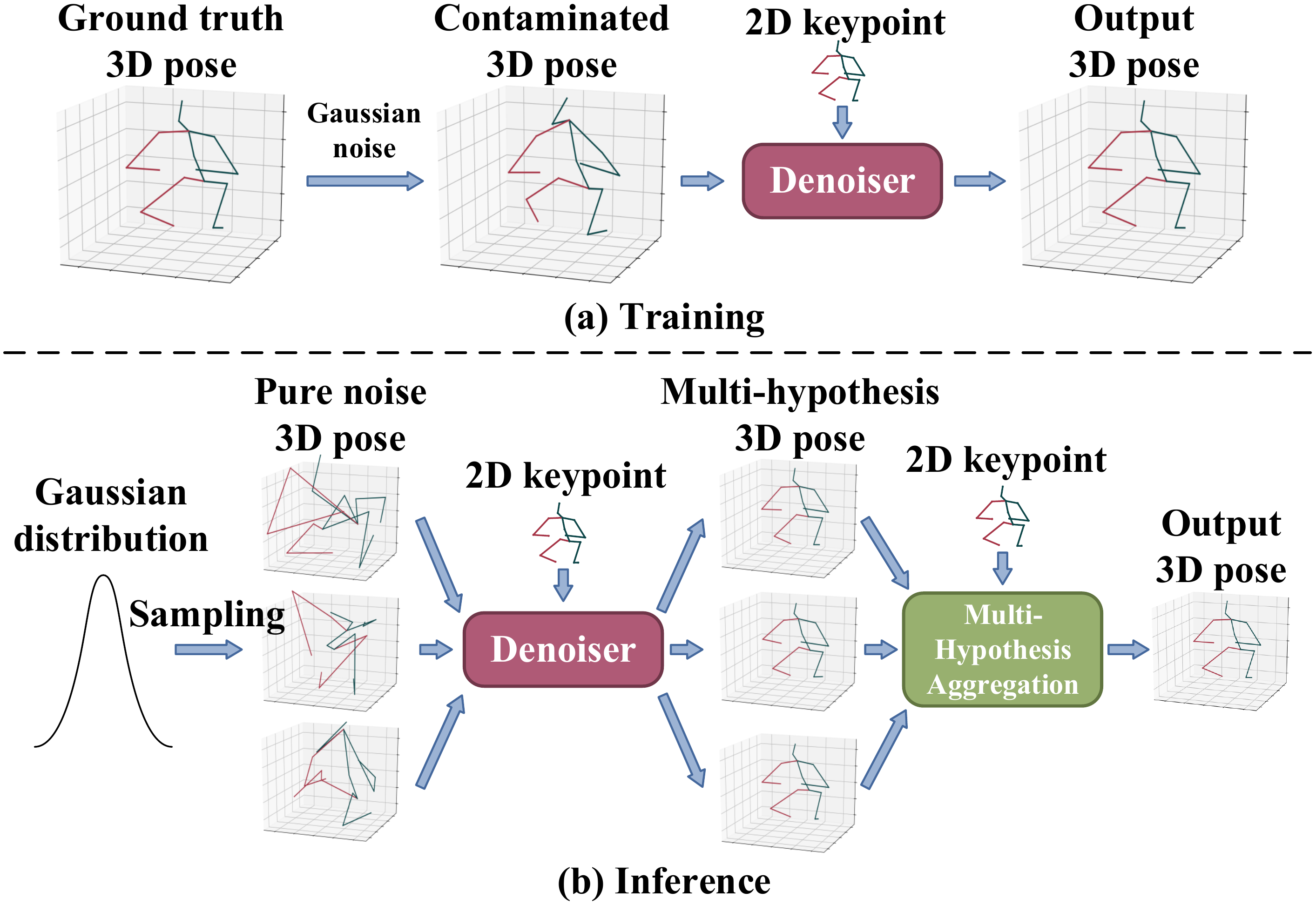}
\end{center}
\vspace*{-0.4cm}
\caption{Thumbnail of our D3DP approach with multi-hypothesis aggregation. (a) Training: we gradually destroy the ground truth 3D pose and reverse this process by a denoiser to obtain a clean 3D prediction. (b) Inference: multiple Gaussian noises are passed through the denoiser, which is conditioned on 2D keypoints, to generate plausible 3D pose hypotheses. Then, the proposed multi-hypothesis aggregation method is utilized to yield the final output.}
\vspace*{-0.5cm}
\label{fig:thumb}
\end{figure}

Existing works can be divided into two categories: \textit{deterministic} and \textit{probabilistic} approaches. Deterministic approaches~\cite{zeng2020srnet,martinez2017simple,zheng20213d,shan2022p,liu2020attention} are designed to produce a single, definite 3D pose for each image, which is practical in real-world applications. Probabilistic approaches~\cite{wehrbein2021probabilistic,li2019generating,jahangiri2017generating,sharma2019monocular,oikarinen2021graphmdn} represent the 2D-to-3D lifting as a probability distribution and produce a set of possible solutions for each image, which allows for uncertainty and ambiguity in the lifting process. This paper focuses on probabilistic methods but combines the advantages of deterministic methods, \ie, we aggregate multiple pose hypotheses into a single, higher-quality 3D pose for practical use.

Previous probabilistic approaches use generative models (\eg, generative adversarial networks~\cite{goodfellow2020generative} or normalizing flows~\cite{rezende2015variational}) to predict multiple 3D pose hypotheses. Despite the great progress achieved, they still have the following two disadvantages: 1) They either (i) rely on special network designs and have poor compatibility~\cite{li2022mhformer,wehrbein2021probabilistic,li2020weakly,sohn2015learning}, or (ii) cannot specify the number of hypotheses based on actual needs~\cite{li2022mhformer,li2019generating,oikarinen2021graphmdn}. 2) Although these methods generate multiple possible hypotheses, an individual 3D pose is still needed for practical use. They generally average over multiple hypotheses at the pose level to obtain the final output~\cite{sharma2019monocular,li2019generating}, which does not take into account the differences between joints and the prior distribution of 2D keypoints. Consequently, the averaged result is much worse than the best of all hypotheses.

To address the first problem, we exploit Denoising Diffusion Probabilistic Models (DDPMs)~\cite{ho2020denoising} for 3D human pose estimation and introduce a novel approach called Diffusion-based 3D Pose estimation (D3DP). As shown in Fig.~\ref{fig:thumb}, our approach involves adding varying levels of noise to the ground truth 3D poses and learning a denoiser to predict the original data during training. During inference, we generate initial 3D poses by sampling pure noise from a Gaussian distribution. These poses are then sent to the denoiser, which is conditioned on 2D keypoints, to predict multiple 3D pose hypotheses. Compared to previous approaches, our method is superior in two aspects. (i) The denoiser has the compatibility to use existing 3D human pose estimators as the backbone, which enables an easy transformation from a deterministic approach into a probabilistic version. (ii) The number of hypotheses can be customized during inference. Besides, the proposed method progressively refines the generated results, which supports an additional adjustable parameter (number of iterations). More hypotheses and iterations bring better results at the expense of greater computational overhead. These two parameters help to strike a balance between performance and efficiency under limited computing resources.

To address the second problem, we focus on how to aggregate multiple 3D pose hypotheses into a single, more accurate 3D prediction. It is observed that the upper-bound performance of aggregation at the joint level is much higher than that at the pose level (Section~\ref{sec:ablation}), as the former conducts aggregation at a finer granularity. This observation drives us to propose a \textit{joint-wise reprojection-based} multi-hypothesis aggregation (JPMA) method, which reprojects 3D pose hypotheses back to the 2D camera plane and conducts the joint-level selection of the best hypothesis based on the reprojection errors. The selected joints are assembled into a complete 3D pose as the final output. The proposed method uses \textit{joints} as granularity, allowing a more diverse combination of hypotheses. In addition, the marginal distribution of 2D keypoints is taken into account, thus aiding the modeling of the posterior distribution of the final 3D pose. Compared to previous methods that use the averaged pose, JPMA integrates joint-level geometric priors into the multi-hypothesis aggregation, and hence improves the prediction performance.

Our contributions can be summarized as follows:
\begin{itemize}
\setlength{\itemsep}{0pt}
\setlength{\parsep}{0pt}
\setlength{\parskip}{0pt}
\setlength{\topsep}{0pt}
\setlength{\partopsep}{0pt}
\item We propose a diffusion-based 3D human pose estimation (D3DP) method, which is compatible and customizable.
\item When multiple hypotheses are combined into a single prediction, we observe that the upper bound of performance for aggregation at the joint level is significantly higher than that at the pose level. This observation drives us to conduct joint-by-joint aggregation. To the best of our knowledge, we are the first to tackle this problem from the \textit{joint} perspective.
\item We propose a joint-wise reprojection-based multi-hypothesis aggregation (JPMA) method, which leverages the 2D prior at the joint level to improve the accuracy of the final 3D prediction. The proposed method proves to be more effective than existing pose-level aggregation methods.
\item Our method outperforms both the state-of-the-art deterministic and probabilistic approaches on 3D human pose estimation benchmarks.
\end{itemize}

\section{Related Work}

\begin{figure*}[t]
\begin{center}
\includegraphics[width=\linewidth]{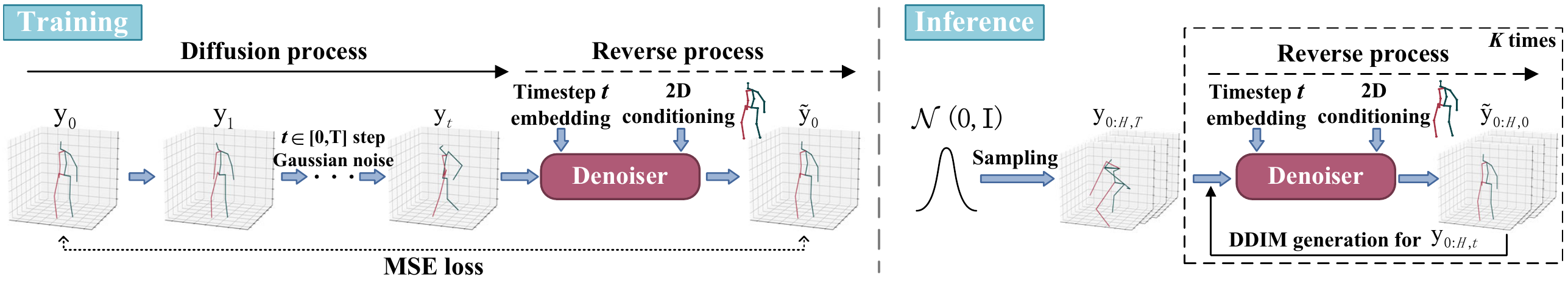}
\end{center}
\vspace*{-0.3cm}
\caption{Overview of the proposed D3DP method. Left: Training. $t$-step Gaussian noise is added to the ground truth 3D pose $\bm{y}_0$, resulting in the noisy pose $\bm{y}_t$. $\bm{y}_t$ is then fed to a denoiser conditioned on 2D keypoints $\bm{x}$ and timestep $t$ to yield the final prediction $\widetilde{\bm{y}}_0$. Right: Inference. $H$ samples are drawn from a Gaussian distribution to initialize 3D poses $\bm{y}_{0:H,T}$, which are utilized to yield the noiseless 3D pose hypotheses $\widetilde{\bm{y}}_{0:H,0}$. Besides, we can iterate the above reverse process $K$ times to refine the final results by sending DDIM-generated 3D poses $\bm{y}_{0:H,t}$ with different levels of noise to the denoiser.}
\vspace*{-0.2cm}
\label{fig:overview}
\end{figure*}

\subsection{Diffusion Model}
Diffusion models~\cite{sohl2015deep} are a family of generative models that gradually deconstruct observed data by adding noise, and then recover the original data by reversing this process. Denoising Diffusion Probabilistic Models (DDPMs)~\cite{ho2020denoising} establish a connection between diffusion models and denoising score matching, which sparks recent interest in this approach. DDPMs have shown remarkable progress in numerous areas, including image generation~\cite{ho2022cascaded,nichol2021glide,batzolis2021conditional,rombach2022high,saharia2022palette,saharia2022image}, cross-modal generation~\cite{avrahami2022blended,fan2022frido,huang2022prodiff,kim2022guided,levkovitch2022zero}, graph generation~\cite{vignac2022digress,niu2020permutation,jo2022score,yan2023swingnn}, semantic segmentation~\cite{baranchuk2021label,brempong2022denoising}, object detection~\cite{chen2022diffusiondet}, etc. For example, SR3~\cite{saharia2022image} adapts DDPMs to super-resolution. DiffusionDet~\cite{chen2022diffusiondet} applies diffusion models to object detection to recover bounding boxes from noise. Pix2Seq-D~\cite{chen2022generalist} leverages diffusion models based on analog bits for panoptic segmentation. 

3D human pose estimation suffers from intrinsic depth ambiguity, making probabilistic generation methods suitable for this task. Since diffusion models are well known for their ability to generate high-fidelity samples, we apply DDPMs to 3D human pose estimation. 

\subsection{Probabilistic 3D Human Pose Estimation}
Early probabilistic methods~\cite{simo2012single,lee2004proposal} rely on heuristic approaches to sample multiple plausible 3D poses that could potentially produce similar 2D projections. Recently, approaches based on mixture density networks (MDN)~\cite{bishop1994mixture}, normalizing flows (NF)~\cite{rezende2015variational}, conditional
variational autoencoder (CVAE)~\cite{sohn2015learning} and generative adversarial networks (GAN)~\cite{goodfellow2020generative} have been proposed. Li and Lee~\cite{li2019generating} introduce MDN to generate pose hypotheses, and Oikarinen \textit{et al.}~\cite{oikarinen2021graphmdn} further combine MDN with graph neural networks that incorporate graph-structured information. Sharma \textit{et al.}~\cite{sharma2019monocular} use CVAE to obtain diverse 3D pose samples. Wehrbein \textit{et al.}~\cite{wehrbein2021probabilistic} and Li \textit{et al.}~\cite{li2020weakly} exploit NF and GAN for this task, respectively. Besides, MHFormer~\cite{li2022mhformer} uses a Transformer~\cite{vaswani2017attention} to generate three features to represent three individual hypotheses.

To overcome the drawbacks of previous methods, such as bad compatibility (MHFormer, NF, GAN, MDN, CVAE), and a non-adjustable number of hypotheses (MHFormer, MDN), we present D3DP, a straightforward and compatible architecture that can be easily integrated with existing 3D pose estimators with minimal modifications. Our method supports a custom number of iterations $K$ and hypotheses $H$ during inference to balance performance and efficiency. Note that concurrent methods~\cite{holmquist2022diffpose,gong2023diffpose,choi2022diffupose} also use diffusion models for this task, but they either don't use any aggregation method or use a pose-level method (average). In contrast, we propose a joint-level aggregation method, called JPMA, to achieve a more accurate prediction.

\subsection{Multi-Hypothesis Aggregation}
Probabilistic 3D pose estimation methods typically generate multiple hypotheses, but how to synthesize them into a single 3D pose is still an open research question. Many methods~\cite{wehrbein2021probabilistic,li2019generating,oikarinen2021graphmdn,li2020weakly} simply average all pose hypotheses, which often produces unsatisfactory results. Other approaches explore more sophisticated ways of aggregating multiple hypotheses. Sharma \textit{et al.}~\cite{sharma2019monocular} use depth ordinal relations to score and weight-average the candidate 3D poses. Oikarinen \textit{et al.}~\cite{oikarinen2021graphmdn} choose the 3D pose with the highest mixture coefficient outputted by MDN. However, these works ignore the differences between joints. Besides, the information contained in 2D keypoints is not exploited. 

Thus, our JPMA method is proposed to use the errors between the reprojected hypotheses and the input 2D keypoints as guidance for joint-level hypothesis selection. Although other approaches~\cite{pavllo20193d,wandt2019repnet,cheng2019occlusion,chen2019unsupervised,kundu2020kinematic} have already used the 3D-to-2D reprojection method in deterministic 3D human pose estimation, our method marks the pioneering effort to apply it to probabilistic estimation. Specifically, reprojection is incorporated into the multi-hypothesis aggregation (MHA), which provides a new perspective on this problem.

\section{Method}

Given the input 2D keypoints $\bm{x} = \{p^{\text{2d}}_n\}_{n=0}^N, p^{\text{2d}}_n \in \mathbb{R}^{J\times 2}$, our goal is to predict the corresponding 3D positions of all joints $\widetilde{\bm{y}} = \{p^{\text{3d}}_n\}_{n=0}^N, p^{\text{3d}}_n \in \mathbb{R}^{J\times 3}$. $N, J$ are the number of frames and human joints in each frame, respectively. Since diffusion models can generate high-quality samples, we apply it to probabilistic 3D human pose estimation and propose D3DP method, which generates multiple 3D pose hypotheses for a single 2D observation. Then, these hypotheses are used by the proposed JPMA method to yield a single, accurate 3D pose for practical use.

\subsection{Diffusion-Based 3D Pose Estimation (D3DP)}

D3DP uses two Markov chains: 1) a \textit{diffusion process} that gradually perturbs data to noise, and 2) a \textit{reverse process} that reconstructs the uncontaminated data by a denoiser. For more information about diffusion models, please refer to Appendix A. The training and inference phases of D3DP are described in detail below.

\paragraph{Training.} As shown in Fig.~\ref{fig:overview} (left), we first sample a timestep $t \sim U(0,T)$, where $T$ is the maximum number of timesteps. Then the ground truth 3D pose $\bm{y}_0$ is diffused to the corrupted pose $\bm{y}_t$ by adding $t$-step independent Gaussian noise $\epsilon \sim \mathcal{N}(0, \bm{I})$. Following DDPMs~\cite{ho2020denoising}, this process can be formulated as
\begin{equation}
\setlength{\abovedisplayskip}{3pt}
\setlength{\belowdisplayskip}{3pt}
    \begin{split}
        q(\bm{y}_t \mid \bm{y}_0) &\coloneqq \sqrt{\bar{\alpha}_t} \bm{y}_0  + \epsilon \sqrt{1 - \bar{\alpha}_t}
    \end{split}
    \label{eq:forward}
\end{equation}

\noindent where $\bar{\alpha}_t \coloneqq \prod_{s=0}^{t} \alpha_s$ and $\alpha_t \coloneqq 1 - \beta_t$. $\beta_t$ is the cosine noise variance schedule. When $T$ is large enough, the distribution of $q(\bm{y}_T)$ is nearly an isotropic Gaussian distribution. 

Subsequently, $\bm{y}_t$ is sent to a denoiser $\mathfrak{D}$ conditioned on 2D keypoints $\bm{x}$ and timestep $t$ to reconstruct the 3D pose $\widetilde{\bm{y}}_0$ without noise:
\begin{equation}
\setlength{\abovedisplayskip}{0pt}
\setlength{\belowdisplayskip}{0pt}
    \begin{split}
        \widetilde{\bm{y}}_0 = \mathfrak{D}\left(\bm{y}_t, \bm{x}, t\right)
    \end{split}
    \label{eq:denoiser}
\end{equation}

\noindent The entire framework is supervised by a simple MSE loss:
\begin{equation}
\setlength{\abovedisplayskip}{1pt}
\setlength{\belowdisplayskip}{1pt}
    \begin{split}
        \mathcal{L} = ||\bm{y}_0 - \widetilde{\bm{y}}_0||_{2}
    \end{split}
    \label{eq:mse_loss}
\end{equation}

\paragraph{Inference.} Since the degraded data is well approximated by a Gaussian distribution after the diffusion process, we can obtain $H$ initial 3D poses $\bm{y}_{0:H,T}$ by sampling noise from a unit Gaussian. As shown in Fig.~\ref{fig:overview} (right), feasible 3D pose hypotheses $\widetilde{\bm{y}}_{0:H,0}$ are predicted by passing $\bm{y}_{0:H,T}$ to the denoiser $\mathfrak{D}$ (Eq.~\ref{eq:denoiser}).

Thereafter, $\widetilde{\bm{y}}_{0:H,0}$ are used to generate the noisy 3D poses $\widetilde{\bm{y}}_{0:H,t'}$ as inputs to the denoiser for the next timestep via DDIM~\cite{song2020denoising}, which can be formulated as
\begin{equation}
\setlength{\abovedisplayskip}{2pt}
\setlength{\belowdisplayskip}{2pt}
    \begin{split}
        \boldsymbol{y}_{0:H,t'}=\sqrt{\bar{\alpha}_{t'}} \cdot \widetilde{\bm{y}}_{0:H,0}+\sqrt{1-\bar{\alpha}_{t'}-\sigma_t^2} \cdot \epsilon_t+\sigma_t \epsilon
    \end{split}
    \label{eq:ddim}
\end{equation}

\noindent where $t, t'$ are the current and next timesteps, respectively. The initial $t=T$. $\epsilon \sim \mathcal{N}(0, \bm{I})$ is standard Gaussian noise independent of $\bm{y}_{0:H,t}$ and
\begin{equation}
\setlength{\abovedisplayskip}{2pt}
\setlength{\belowdisplayskip}{2pt}
    \begin{split}
        \epsilon_t&=\left(\bm{y}_{0:H,t}-\sqrt{\bar{\alpha}_t} \cdot \widetilde{\bm{y}}_{0:H,0}\right)/\sqrt{1-\bar{\alpha}_t}\\
        \sigma_t&=\sqrt{\left(1-\bar{\alpha}_{t'}\right) / \left(1-\bar{\alpha}_t\right)} \cdot \sqrt{1-\bar{\alpha}_t / \bar{\alpha}_{t'}}
    \end{split}
    \label{eq:ddim-param}
\end{equation}
\noindent where $\epsilon_t$ is the noise at timestep $t$ (derived from Eq.~\ref{eq:forward}). $\sigma_t$ controls how stochastic the diffusion process is.

Then, we can regenerate $\widetilde{\bm{y}}_{0:H,0}$ using $\bm{y}_{0:H,t'}$ as inputs to the denoiser. This procedure will be iterated $K$ times. Since we start from T at the beginning, the timestep of each iteration can be written as $t=T\cdot \left(1-k/K\right), k\in [0,K)$. The detailed algorithm is provided in Appendix C.

In this way, the proposed method supports a customizable number of hypotheses $H$ by sampling multiple times from a Gaussian distribution. Moreover, the denoiser is trained only once but can be used multiple times to progressively refine the final predictions during inference. This refinement process supports an additional adjustable parameter (number of iterations $K$), which controls the diversity and quality of the generated hypotheses. Thus, we can specify an arbitrary number of $H,K$ during inference (both are set to 1 during training), which allows users to balance performance and efficiency. This property solves the problem of a non-adjustable number of hypotheses faced by previous methods~\cite{li2022mhformer,li2019generating,oikarinen2021graphmdn}.

\subsection{Joint-Wise Reprojection-Based \\Multi-Hypothesis Aggregation (JPMA)}
Previous probabilistic 3D pose estimation methods~\cite{jahangiri2017generating,rezende2015variational,li2019generating} mainly focus on the generation of 3D pose hypotheses, with little attention to how to aggregate multiple hypotheses to produce a single, high-fidelity prediction. These approaches typically report the error of the best 3D pose (\ie, closest to the ground truth) to highlight the superiority of their methods. However, this metric only demonstrates the upper bound of the performance and is not feasible in real-world applications because the ground truth is not available. Therefore, the averaged 3D pose is used as the final prediction, while its performance is far worse than the best result. A few works~\cite{oikarinen2021graphmdn,sharma2019monocular} explore multi-hypothesis aggregation methods that are more effective than average for testing on in-the-wild videos. However, they use \textit{poses} as granularity and ignore the distinctions between joints. Also, they do not establish the correspondence between 3D predictions and the 2D observation. Hence, further improvements are needed to achieve more accurate and reliable results.

\begin{figure}[t]
\begin{center}
\includegraphics[width=\linewidth]{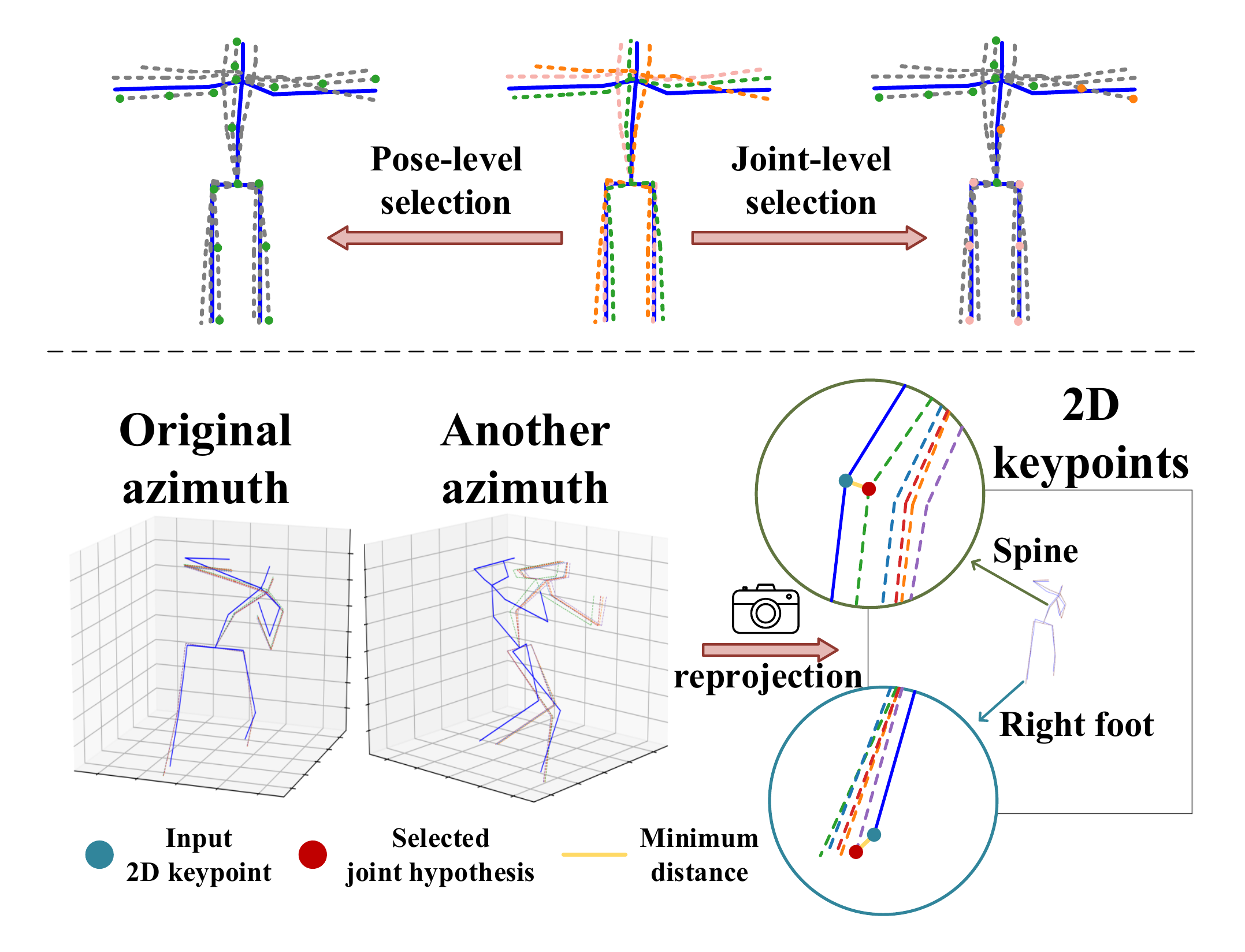}
\end{center}
\vspace*{-0.4cm}
\caption{Top: Comparison between the pose-level (existing methods) and joint-level (proposed method) selections. The selected joint is drawn in the color of the hypothesis it belongs to. Bottom: Our JPMA method. 3D pose hypotheses are reprojected to the 2D camera plane and compared with the input 2D keypoints. For each joint, we select the best hypothesis that has the minimum distance to the 2D input. Dashed line: predicted 3D pose hypotheses. Each color represents an individual hypothesis. Solid blue line: ground truth 3D poses.}
\vspace*{-0.2cm}
\label{fig:JPMA}
\end{figure}

To alleviate these problems, we first validate the upper-bound performance of aggregation methods at two levels by choosing the best hypothesis that is closest to the ground truth. In the case of pose level, the best pose is selected as the final prediction. In the case of joint level, the best joint is selected and all the selected joints are combined into the final prediction. Experimental results show that the latter has a higher upper-bound performance than the former (Section~\ref{sec:ablation}), as the joint-level aggregation allows more flexible combinations of hypotheses. An intuitive comparison between the upper bounds of joint-level and pose-level selections (aggregations) is shown in Fig.~\ref{fig:JPMA} (top). When the pose-level selection is used, the green hypothesis (best pose) will be used as the results for all joints. However, the performance of the same hypothesis may vary across joints. Thus, the joint-level selection decomposes and recombines the hypotheses at the joint level instead of viewing each 3D pose as a whole. Only a fraction of joints (those with better performance) in each hypothesis will be used by the joint-level selection, and then these selected joints are assembled into an independent 3D pose. In this way, the advantage of each hypothesis is maximized. Moreover, the input 2D keypoints can be used as geometric priors to guide the model to choose the most likely 3D pose $\widetilde{\bm{y}}_0$ with the maximum posterior probability $q(\widetilde{\bm{y}}_0|\bm{x},\widetilde{\bm{y}}_{0:H,0})$ as the final prediction during inference. Although 2D keypoints contain no depth information about 3D poses, they indicate the possible locations of human joints in 3D space (\ie, the joint should be on the ray from the camera's optical center to the 2D keypoint). Thus, 2D keypoints still play an important role in multi-hypothesis aggregation.

Based on the above two ideas, we propose a joint-wise reprojection-based multi-hypothesis aggregation (JPMA) method, which is illustrated in Fig.~\ref{fig:JPMA} (bottom). We use known or estimated intrinsic camera parameters (4 for pinhole camera and 9 for distorted pinhole camera) to reproject 3D hypotheses $\widetilde{\bm{y}}_{0:H,0}$ to the 2D camera plane. Then, we calculate the distance between each hypothesis and the input 2D keypoint $\bm{x}$ for each joint, and select the joint hypothesis with the minimum distance (reprojection error). This process allows us to choose distinct hypotheses for different joints (green hypothesis for the spine and purple hypothesis for the right foot) and combine all the selected joints into the final prediction $\widetilde{\bm{y}}_0$, which can be formulated as:
\begin{equation}
\setlength{\abovedisplayskip}{5pt}
\setlength{\belowdisplayskip}{5pt}
    \begin{split}    
        \widetilde{\bm{y}}_0^{(i)} = \widetilde{\bm{y}}_{h',0}^{(i)}, \quad h'=\underset{h \in [0,H]}{\arg \min }||\mathcal{P}\left(\widetilde{\bm{y}}_{h,0}^{(i)}\right)-\bm{x}^{(i)}||_{2}
    \label{eq:JPMA}
    \end{split}
\end{equation}
\noindent where $i$ is the index of joints. $\mathcal{P}(\cdot)$ is the reprojection function, which is detailed in Appendix B.

\subsection{Architecture}
The previous approaches~\cite{wehrbein2021probabilistic,li2022mhformer,li2020weakly,sohn2015learning} rely on special network designs, for example, MHFormer~\cite{li2022mhformer} needs to use the cross-hypothesis interaction module and NF~\cite{wehrbein2021probabilistic} requires the construction of bijective transformations. Therefore, they cannot keep up with the latest 3D human pose estimation methods. While the proposed D3DP method is compatible with up-to-date deterministic 3D pose estimators by using them as the backbone of the denoiser. In this paper, we use MixSTE~\cite{zhang2022mixste}, a mixed spatial temporal Transformer-based method, as the backbone. We can easily transform a deterministic pose estimation approach into a probabilistic version by making minor modifications, which are introduced in the next two paragraphs.

\paragraph{2D conditioning.}
Since our goal is to estimate 3D poses from 2D keypoints, using only noisy 3D poses $\bm{y}_{0:H,t}$ as inputs to the denoiser is not sufficient. Therefore, 2D keypoints $\bm{x}$ are used as additional information to guide the denoising process. We implement several schemes for fusing 2D keypoints and noisy 3D poses, such as cross attention, concatenation, etc. According to the experimental results (Appendix D.1), we choose to directly concatenate $\bm{x}$ and $\bm{y}_{0:H,t}$, and send them to the denoiser as inputs.

\paragraph{Timestep embedding.}
As the denoiser needs to handle 3D poses with different levels of noise, it is necessary to provide information about the current timestep $t$, which represents the number of times the Gaussian noise is added. Following DDPMs, we use a sinusoidal function to transform $t$ into a timestep embedding and add it after the input embedding in the same way as the Transformer positional embedding~\cite{vaswani2017attention}.

These two modifications bring a negligible amount of computation. In addition, JPMA mainly consists of a reprojection process (without network parameters), so its amount of computation can also be ignored. Therefore, the computational overhead of the proposed D3DP with JPMA (when $H$=$K$=1) is essentially the same as that of the backbone network.

\begin{table*}[t]
\centering
\input{table/h36m_new.tex}
\vspace{-0.3cm}
\caption{Results on Human3.6M in millimeters under MPJPE. $N,H,K$: the number of input frames, hypotheses, and iterations of the proposed D3DP. ($\ddagger$) - Our implementation. ($\sharp$) - Not feasible in real-world applications. (*) - Use CPN ~\cite{chen2018cascaded} as the 2D keypoint detector to generate the inputs. \textcolor{red}{Red}: Best. \textcolor{blue}{Blue}: Second best. \colorbox{bestcolor}{Gray}: our method.}
\label{tab:h36m}
\vspace{-0.3cm}
\end{table*}

\section{Experiments}
\subsection{Datasets and Evaluation Metrics}
\paragraph{Human3.6M (H36M)}~\cite{ionescu2013human3} is the largest and most commonly used indoor dataset. It contains 15 activities (\eg, walking) performed by 11 actors. Videos are captured by 4 synchronized and calibrated cameras at 50Hz. Following~\cite{li2022mhformer,pavllo20193d,oikarinen2021graphmdn}, our model is trained on 5 subjects (S1, S5, S6, S7, S8) and evaluated on 2 subjects (S9, S11). For evaluation metrics, we report the mean per joint position error (MPJPE), which computes the mean Euclidean distance between estimated and ground truth 3D joint positions in millimeters. Besides, Procrustes MPJPE (P-MPJPE) is also reported, which computes MPJPE after the estimated poses align to the ground truth using a rigid transformation.

\paragraph{MPI-INF-3DHP (3DHP)}~\cite{mehta2017monocular} is a recently popular dataset consisting of indoor and outdoor scenes. The training set contains 8 activities performed by 8 actors, and the test set covers 7 activities. Following~\cite{shan2022p}, we use the valid frames provided by the official for testing. For evaluation metrics, we report MPJPE, percentage of correct keypoint (PCK) within 150mm range, and area under curve (AUC).

\paragraph{3DPW}~\cite{vonMarcard2018} is the first dataset in the wild that includes video footage taken from a moving phone camera. We train our method on Human3.6M and evaluate on this dataset to show the qualitative results.

\subsection{Implementation Details}
\begin{figure}[t]
\begin{center}
\includegraphics[width=\linewidth]{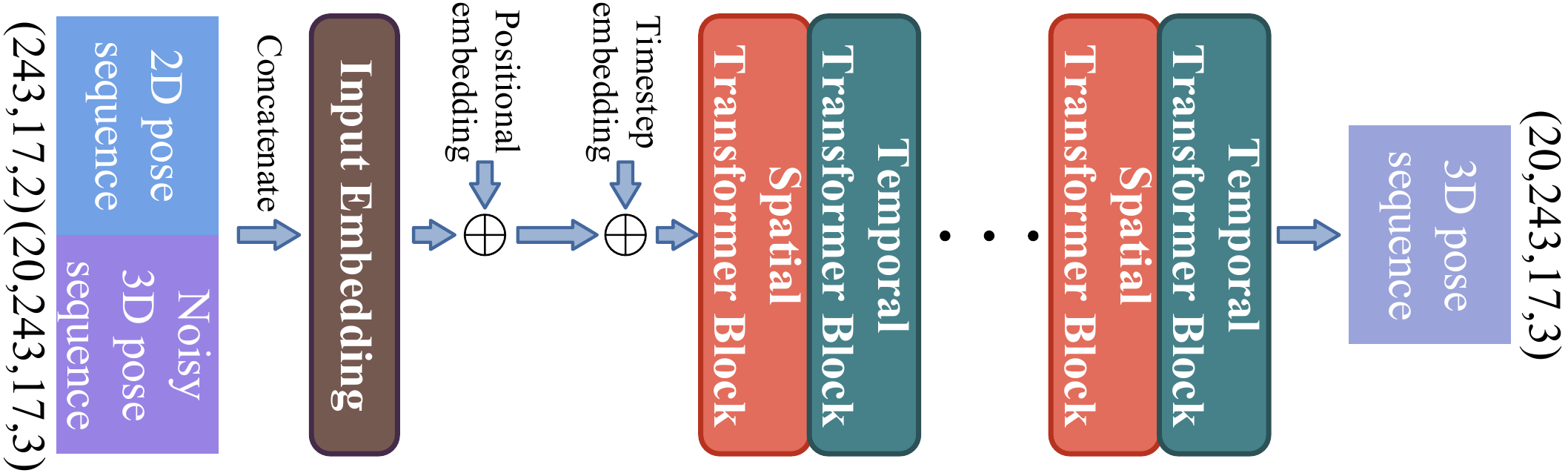}
\end{center}
\vspace*{-0.4cm}
\caption{Detailed architecture of the denoiser. The tensor sizes are shown in parentheses. The 2D pose sequence (243,17,2) is expanded, repeated, and then concatenated with the noisy 3D pose sequence (20,243,17,3) to get the input sequence (20,243,17,5). 20, 243, 17 are the number of hypotheses $H$, frames $N$, and human joints $J$ in each frame, respectively.}
\vspace*{-0.3cm}
\label{fig:denoiser_arc}
\end{figure}

We use MixSTE~\cite{zhang2022mixste} as the backbone of the denoiser, which makes use of factorized attention by combining multiple spatial and temporal attention layers. It contains 16 alternating spatial and temporal Transformer layers~\cite{vaswani2017attention} with the channel size set to 512. We modify MixSTE to fit the diffusion framework as follows: concatenating 2D keypoints and noisy 3D poses as inputs; adding timestep embedding after the input embedding in the first layer. The detailed architecture of the denoiser is shown in Fig.~\ref{fig:denoiser_arc}.

The proposed method is implemented in PyTorch~\cite{paszke2019pytorch} using AdamW~\cite{loshchilov2017decoupled} optimizer with the momentum parameters as $\beta_1,\beta_2=0.9,0.999$ and the weight decay as 0.1. We train our model for 400 epochs and the initial learning rate is $6e^{-5}$ with a shrink factor of 0.993 after each epoch. We set the batch size to 4 and each sample contains a pose sequence of 243 frames ($N$=243). During training, the number of hypotheses and iterations $H,K$ are set to 1,1, respectively. During inference, they are set to 20, 10. The maximum number of timesteps $T$ is set to 1000. All experiments are carried out on two GeForce RTX 3080 Ti GPUs.

\begin{figure*}
\begin{center}
\small
\setlength{\tabcolsep}{2pt}
\begin{tabular}{cccccc}
\multicolumn{2}{c}{\small H36M}&\multicolumn{2}{c}{\small 3DHP}&\multicolumn{2}{c}{\small 3DPW}\\
%{\small H36M} & {\small 3DHP} & {\small 3DPW} \\
{\includegraphics[width=0.12\textwidth]{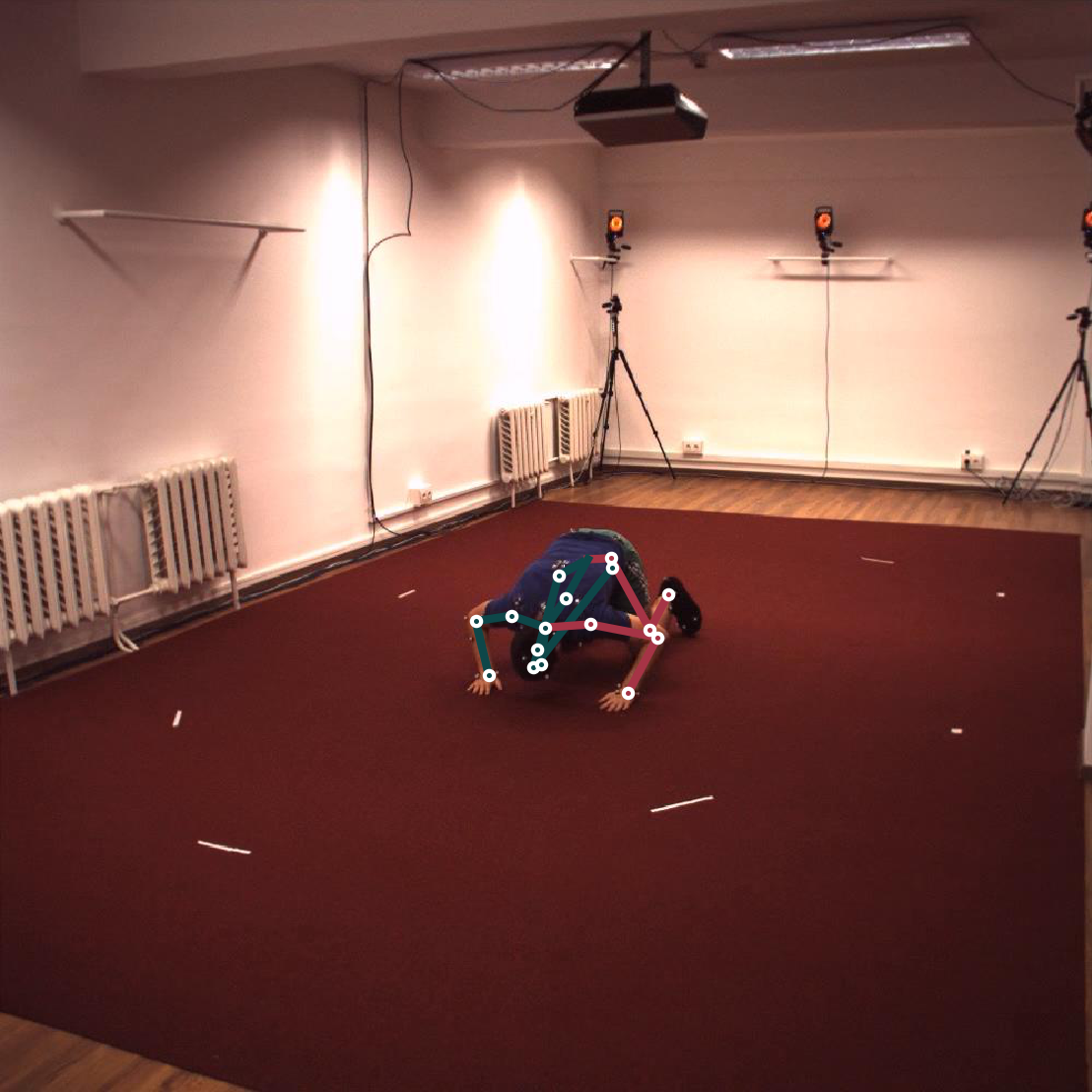}} &
{\includegraphics[width=0.18\textwidth]{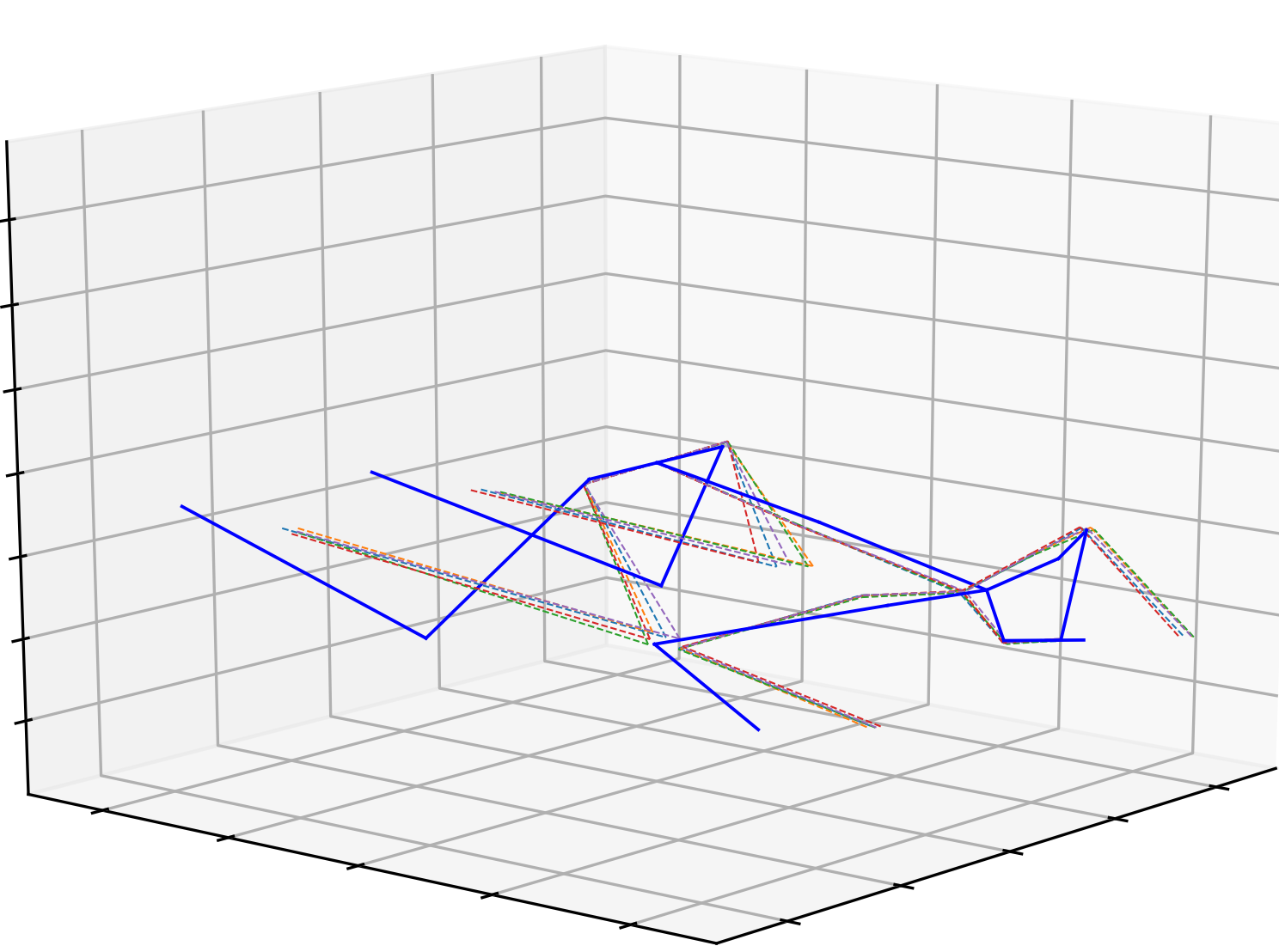}} &
{\includegraphics[width=0.12\textwidth]{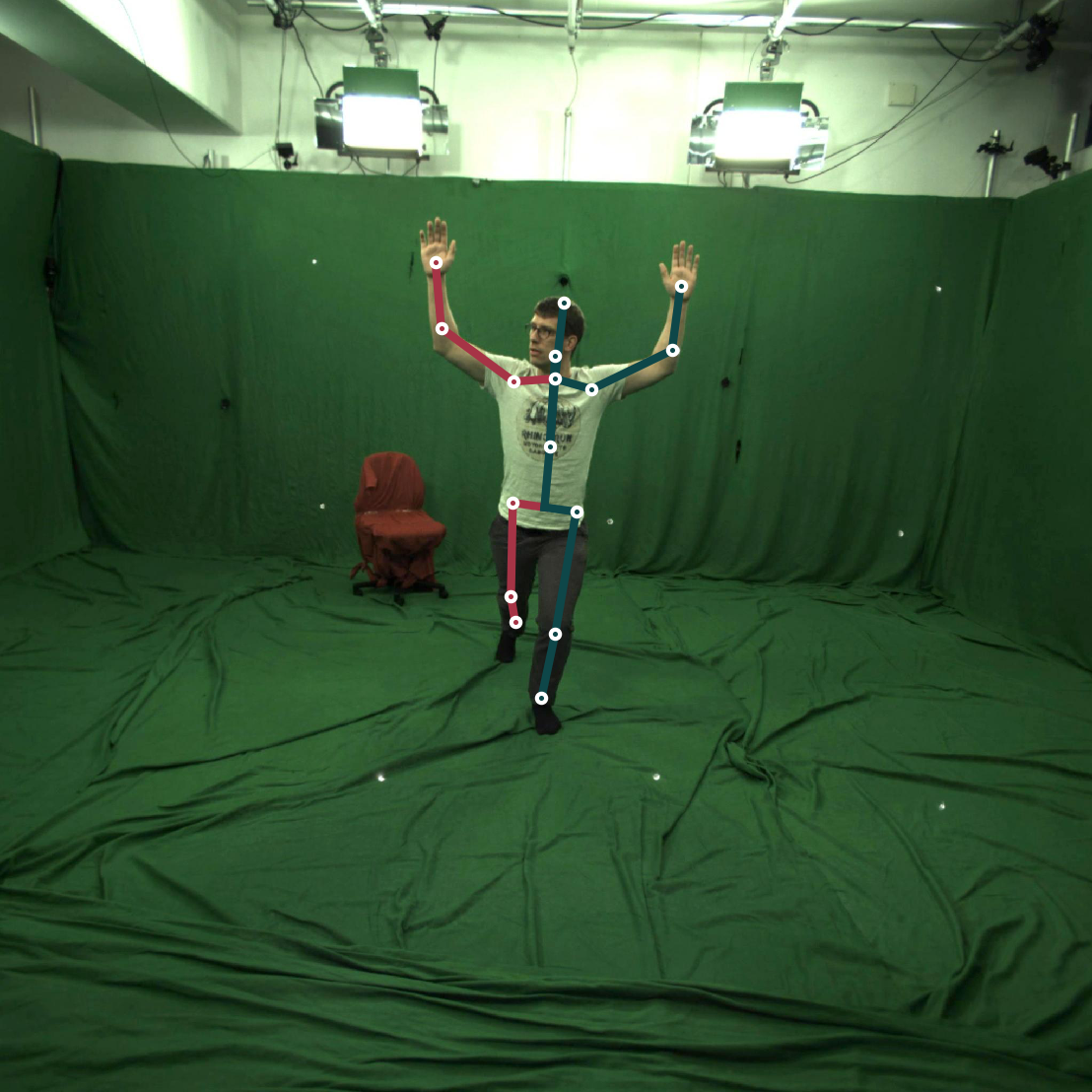}} &
{\includegraphics[width=0.18\textwidth]{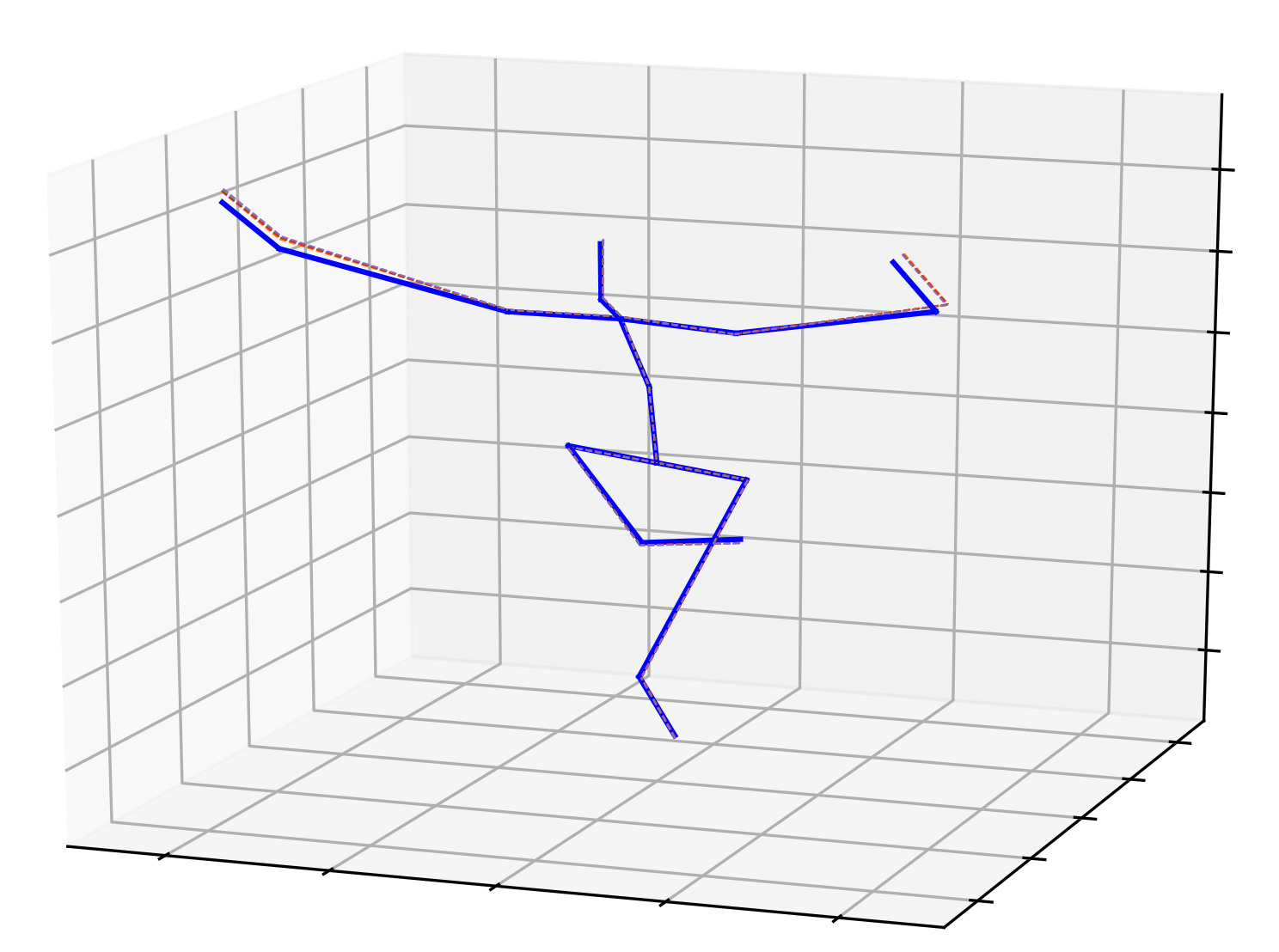}} &
{\includegraphics[width=0.12\textwidth]{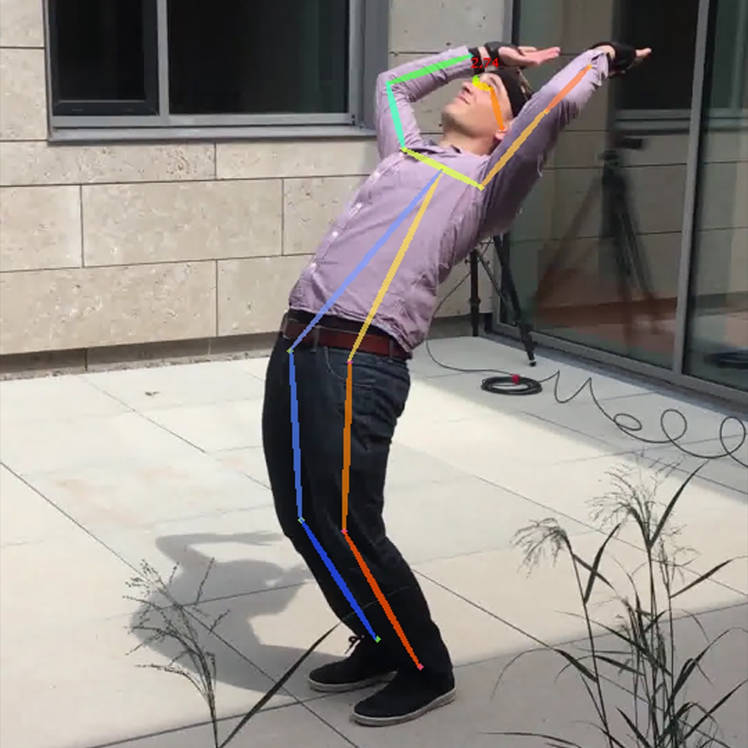}} &
{\includegraphics[width=0.18\textwidth]{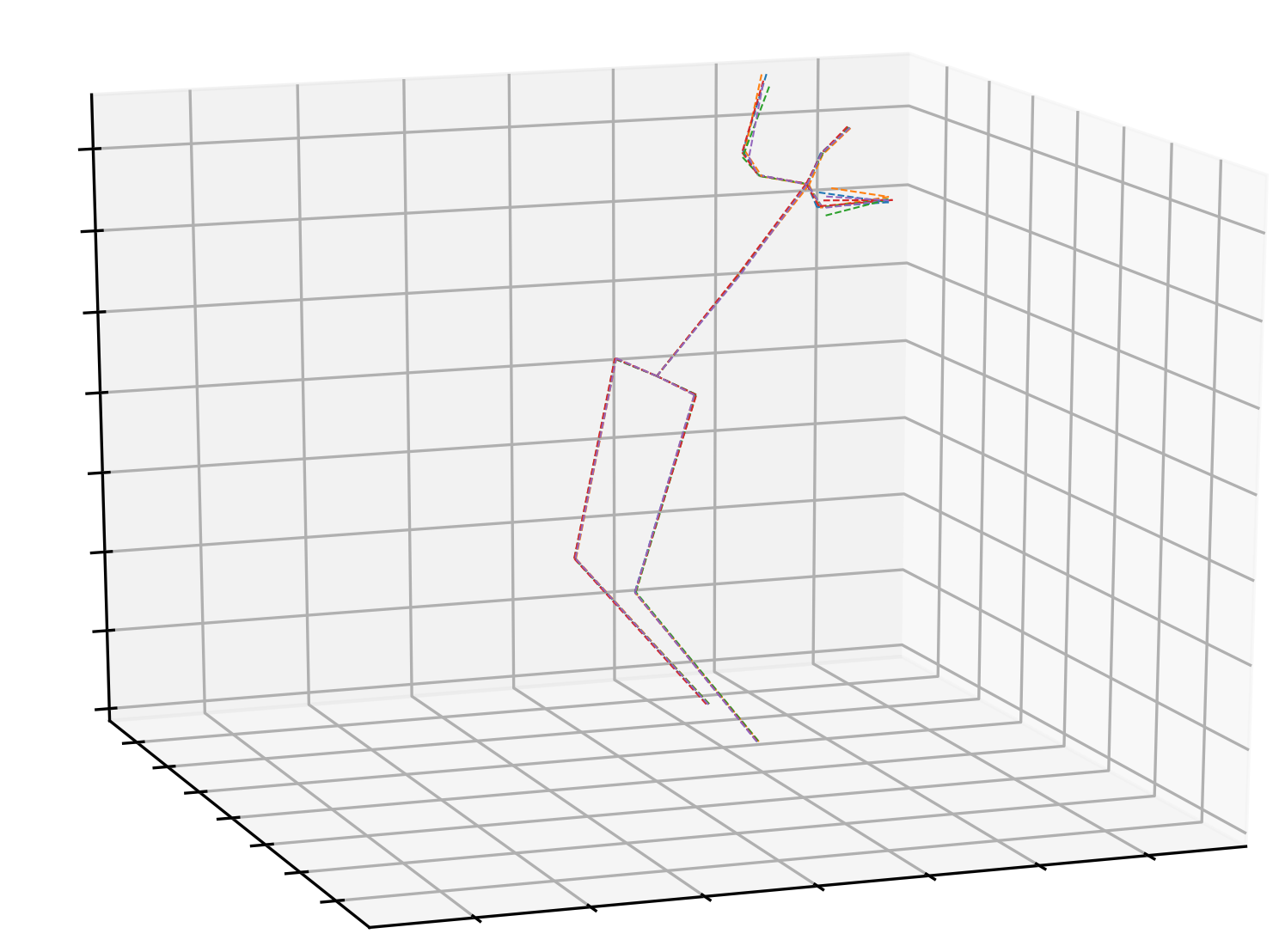}} \\
[-2pt]

{\includegraphics[width=0.12\textwidth]{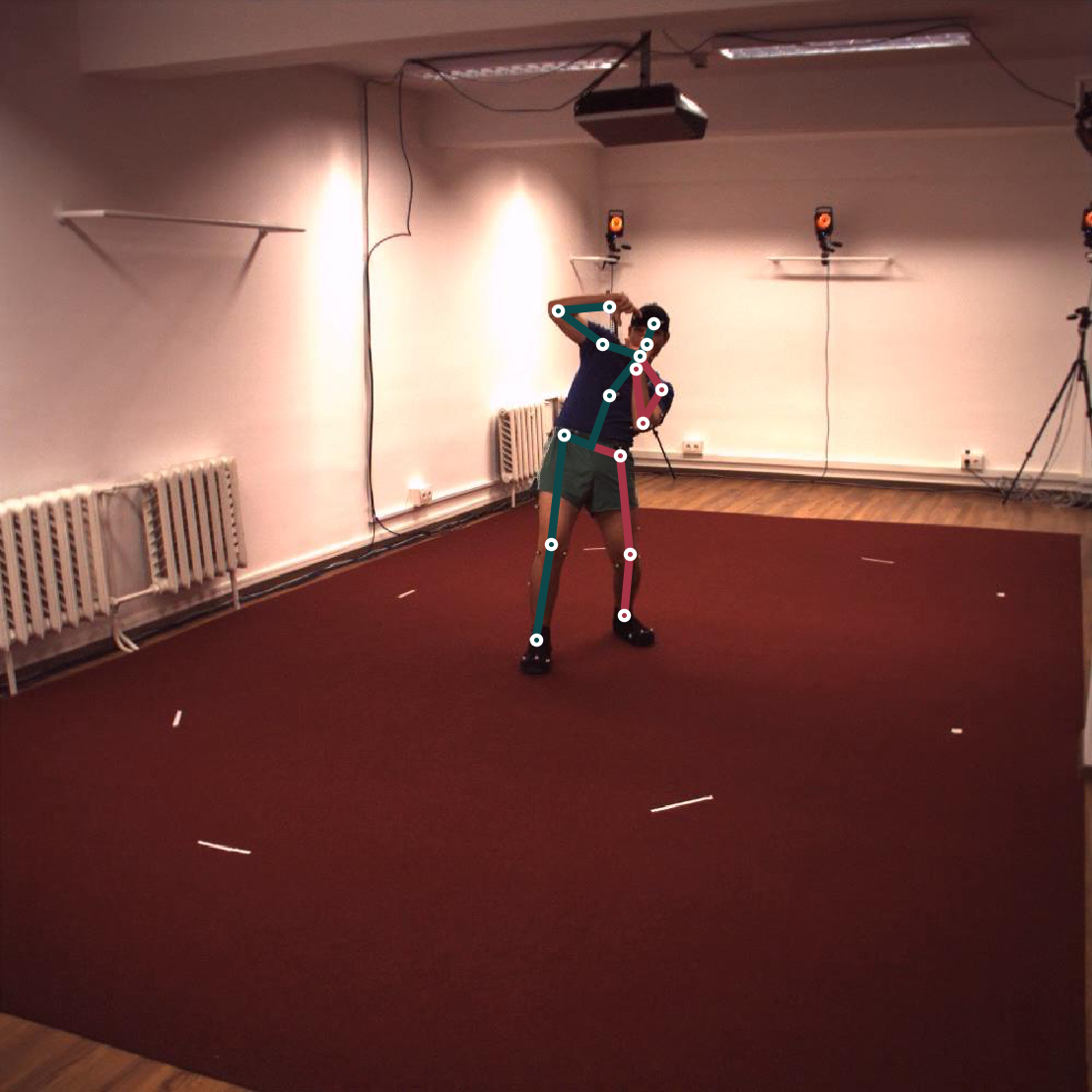}} &
{\includegraphics[width=0.18\textwidth]{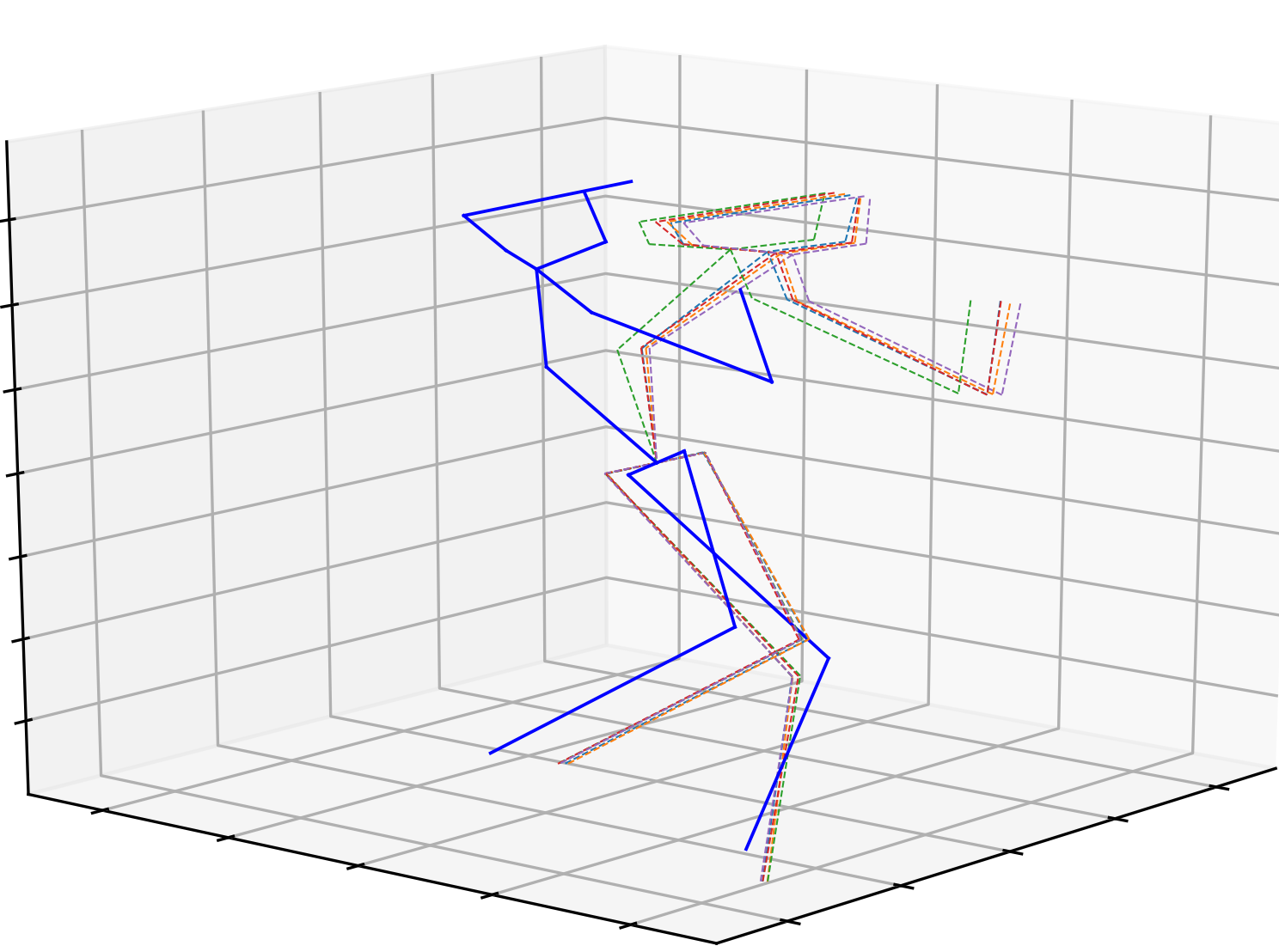}} &
{\includegraphics[width=0.12\textwidth]{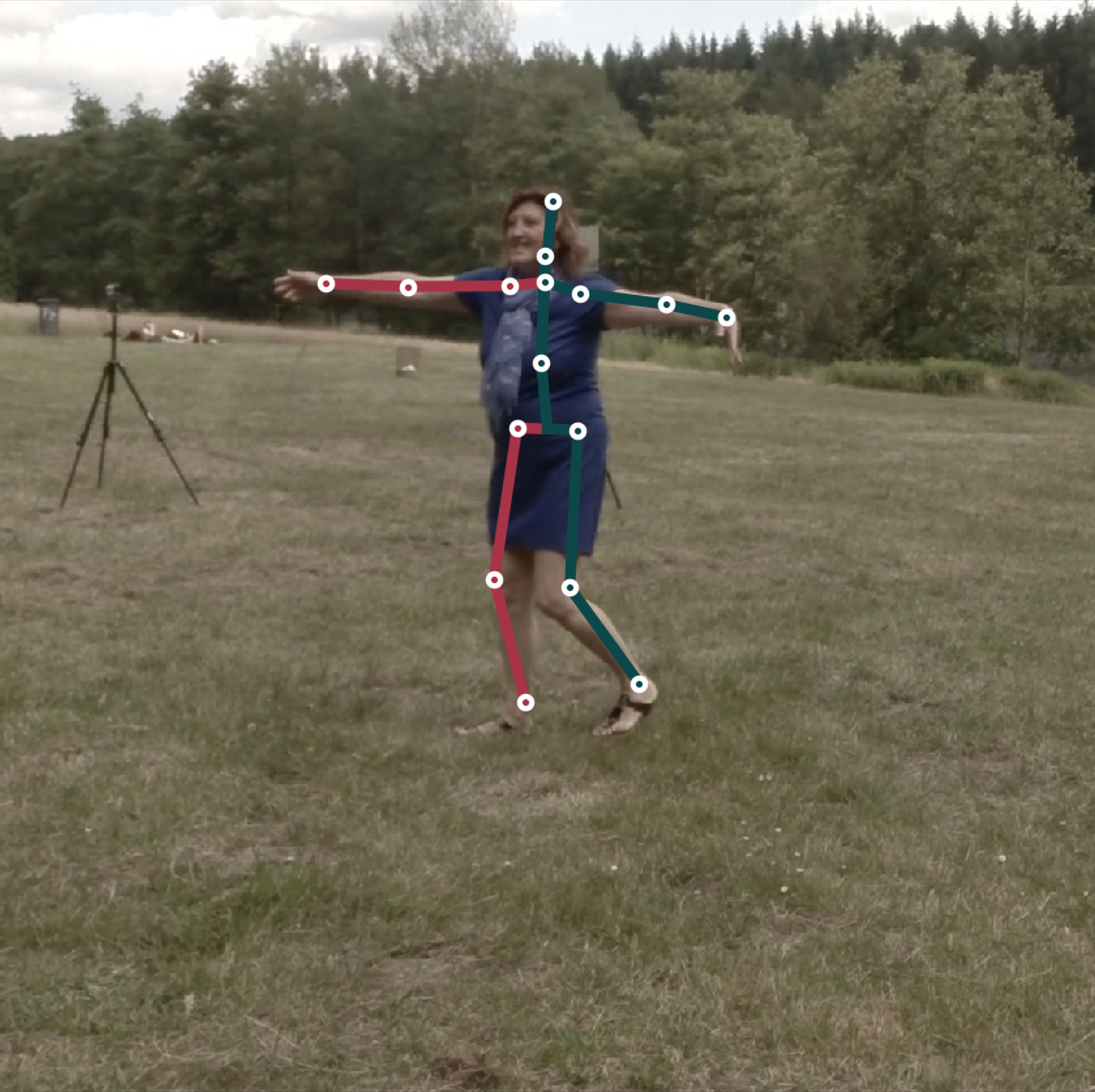}} &
{\includegraphics[width=0.18\textwidth]{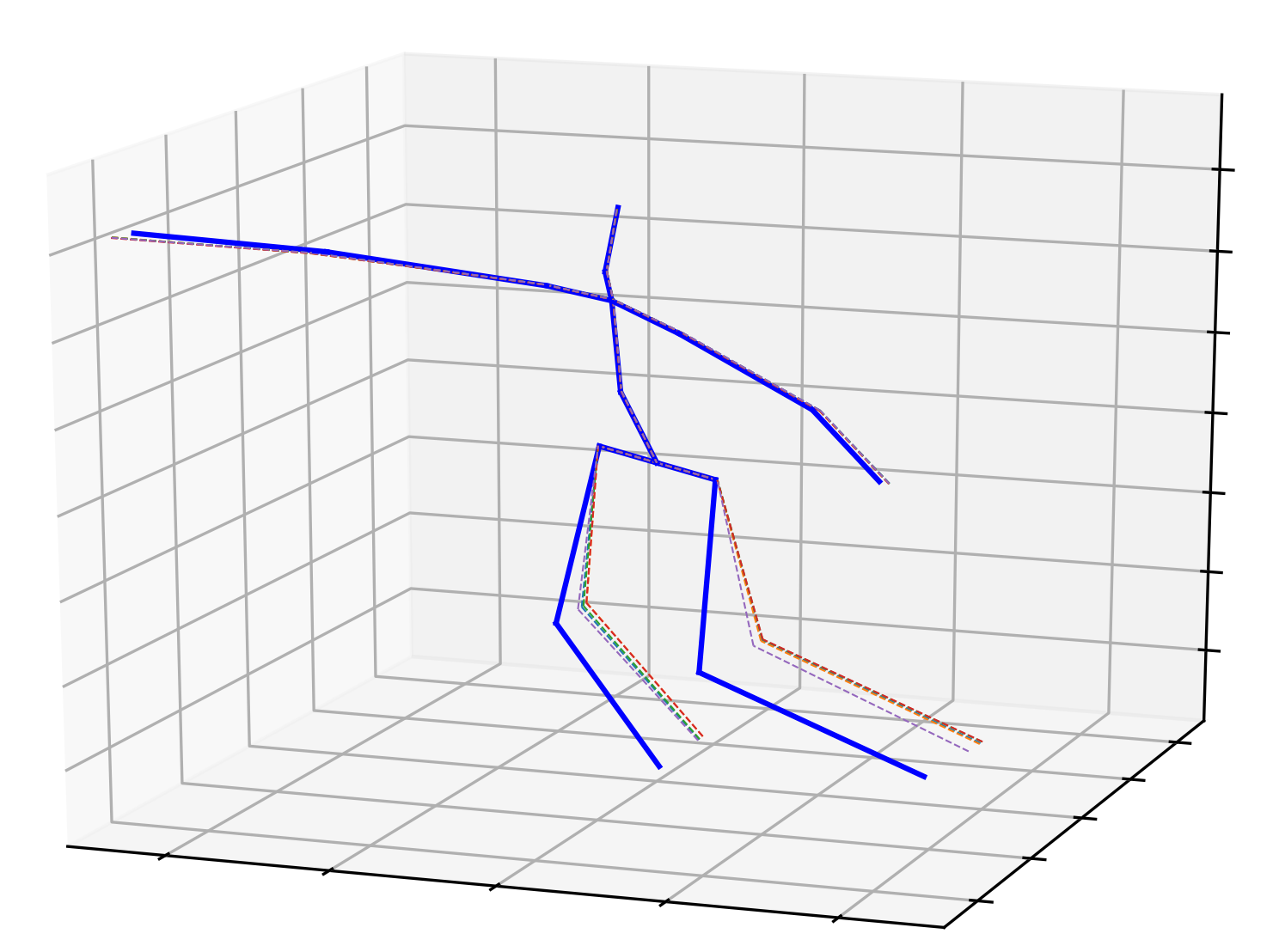}} &
{\includegraphics[width=0.12\textwidth]{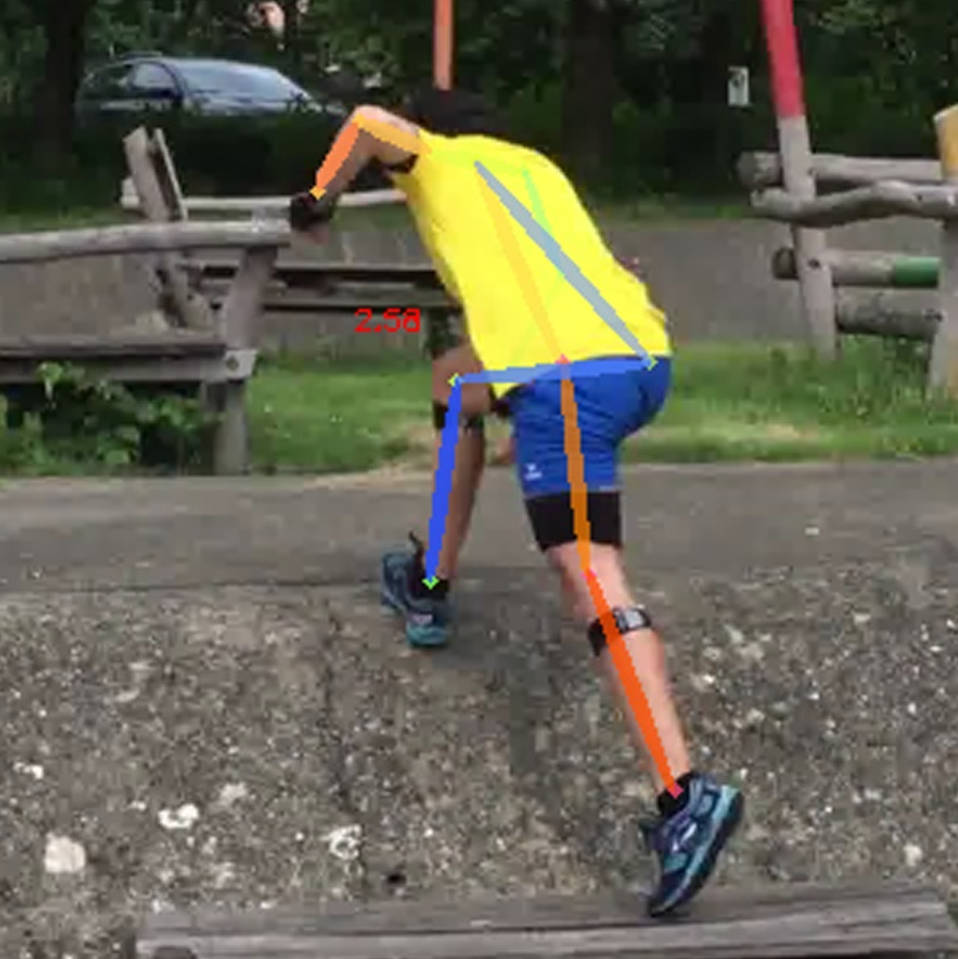}} &
{\includegraphics[width=0.18\textwidth]{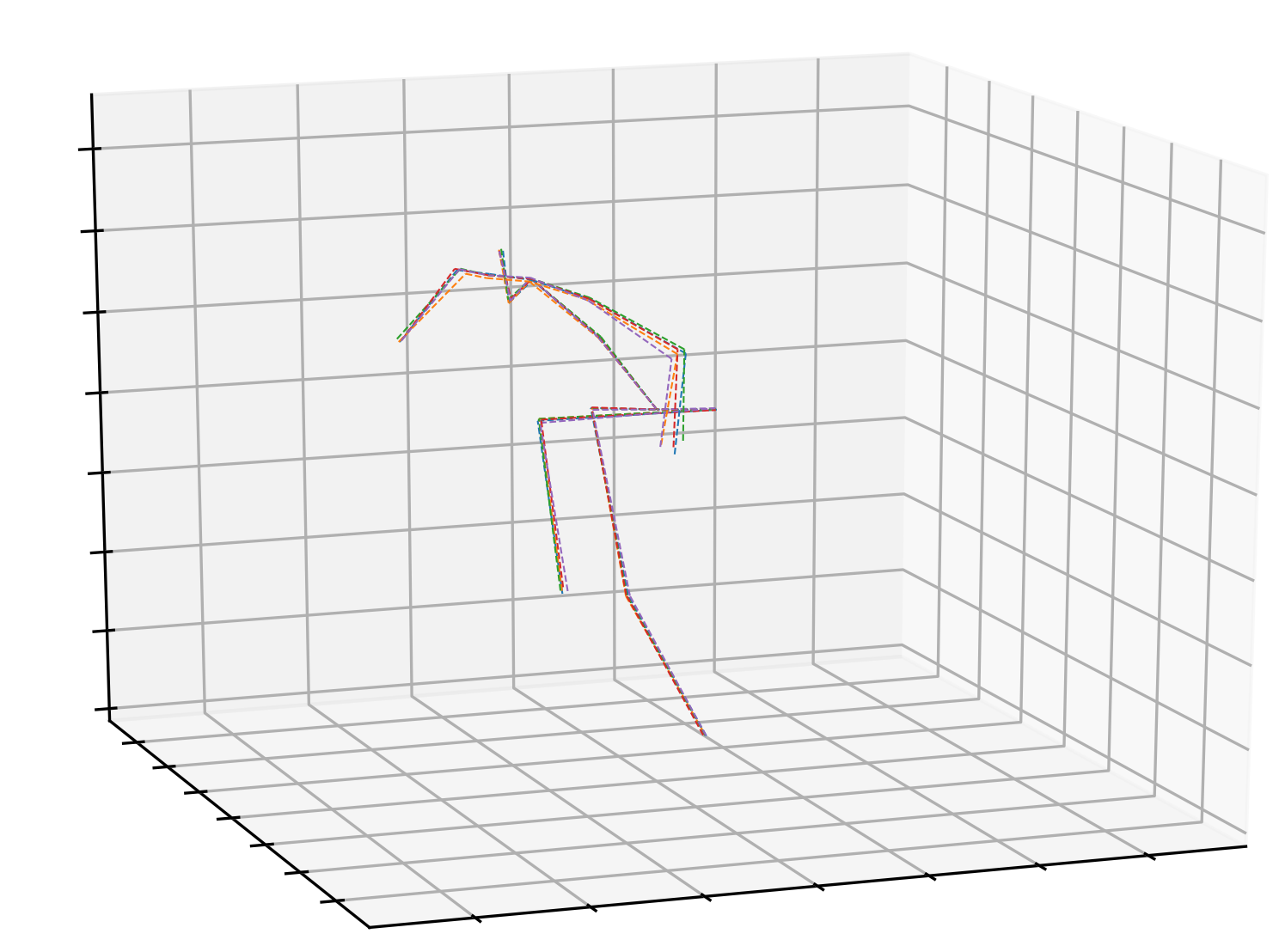}} \\

\end{tabular}
%\vspace*{-0.2cm}
\end{center}
\vspace*{-0.4cm}
\caption{Qualitative results on three datasets. Dashed line: predicted 3D pose hypotheses. Each color represents an individual hypothesis. Solid blue line: ground truth 3D poses.}
\vspace*{-0.3cm}
\label{fig:qualitative}
\end{figure*}

\subsection{Quantitative Results}
% All experiments are conducted using MixSTE~\cite{zhang2022mixste} as the backbone of the denoiser. Implementation details are provided in Appendix C.

\paragraph{Human3.6M.}
As shown in Table~\ref{tab:h36m} (top), we compare the proposed D3DP with the state-of-the-art deterministic 3D human pose estimation methods on Human3.6M. For fairness, we set $H$=1 to produce a single prediction. Our D3DP ($N$=243) achieves 40.0mm under MPJPE, outperforming DUE~\cite{zhang2022uncertainty} ($N$=300) by 0.6mm, even though we use fewer frames. Moreover, our method exceeds MixSTE~\cite{zhang2022mixste} (our backbone) by 1 mm under MPJPE. Although our method aims to generate multiple 3D pose hypotheses, it still improves performance in the single-hypothesis case. This result demonstrates that D3DP (\ie, conditional denoising from corrupted 3D poses) helps the network generalize better to the test set, thus improving the prediction accuracy.

Additionally, D3DP is compared with other probabilistic methods, as shown in Table~\ref{tab:h36m} (bottom). We report the results according to the following four experimental settings.
\begin{itemize}
\setlength{\itemsep}{0pt}
\setlength{\parsep}{0pt}
\setlength{\parskip}{0pt}
\setlength{\topsep}{0pt}
\setlength{\partopsep}{0pt}
\item \textbf{P-Agg}: Use pose-level aggregation methods, which treat each 3D pose as the smallest unit to compose the final prediction, such as depth ordinal guidance~\cite{sharma2019monocular}, mean~\cite{oikarinen2021graphmdn} and mean-shift~\cite{li2020weakly}.
\item \textbf{J-Agg} (proposed): Use joint-level aggregation methods (our JPMA) to yield the final prediction. Each joint is treated as the smallest unit.
\item \textbf{P-Best}$\sharp$: Select the 3D pose hypothesis that is closest to the ground truth. It is the upper bound of P-Agg performance.
\item \textbf{J-Best}$\sharp$ (proposed): Select the joint hypothesis that is closest to the ground truth, and then combine the selected joints into the final 3D pose. It is the upper bound of J-Agg performance.
\end{itemize}
\vspace{-0.1cm}
\noindent ($\sharp$) refers to the settings that are not feasible for in-the-wild videos. When testing under traditional P-Best and P-Agg (average is used in our experiments), our approach has a significant improvement over NF~\cite{wehrbein2021probabilistic} (4.8mm) and MHFormer~\cite{li2022mhformer} (3.1mm), respectively. However, the increase of $H$ (1$\rightarrow$20) does not significantly improve the performance of the proposed D3DP method under P-Best (40.0$\rightarrow$39.5mm). This result prompts us to propose the new setting J-best, under which a noticeable improvement (40.0$\rightarrow$35.4mm) can be observed. It is because the same hypothesis may perform differently depending on which joint is being considered. This observation motivates us to propose J-Agg (using our JPMA) to exploit the joint-level differences between hypotheses. Compared to P-Agg, the performance under J-Agg is boosted (39.9$\rightarrow$39.5mm). This performance is the same as P-Best (39.5mm). More results in terms of P-MPJPE or using 2D ground truth as inputs are provided in Appendix E.

\begin{table}[t]
\centering
\input{table/3dhp_new.tex}
\vspace{-0.3cm}
\caption{Results on MPI-INF-3DHP under three evaluation metrics using ground truth 2D keypoints as inputs. ($\triangledown$) - They train and evaluate on 3D poses scaled to the height of the universal skeleton used by H36M, while we use the ground truth 3D poses.}
\label{tab:3dhp}
\vspace{-0.3cm}
\end{table}
\paragraph{MPI-INF-3DHP.}
Table~\ref{tab:3dhp} shows comparisons between our method and others on 3DHP dataset. D3DP reduces the MPJPE by 5.2mm on top of MixSTE~\cite{zhang2022mixste} in the single-hypothesis case ($H$=1). When $H$ is increased to 20, the performance under J-Agg using JPMA is similar to P-Best and superior to P-Agg. Our method performs best when evaluated under J-Best. These results exhibit the same pattern as those on Human3.6M, which demonstrates the strong generalization capability of our method.

\subsection{Qualitative Results}

Fig.~\ref{fig:qualitative} shows the qualitative results of D3DP on three datasets with $H,K$ set to 5. For the cases of 2D keypoint misdetection (top left: right knee), occlusion (top left: lower body, bottom right: right arm), and depth ambiguity (bottom left: upper body), our method generates pose hypotheses with high variance to reflect the uncertainty in 3D pose estimation. For simple poses (top/bottom middle), D3DP is inclined to generate a deterministic prediction. For rare poses (top right), although the predicted 3D hypotheses are not accurate, they are anatomically plausible and diverse in the arm region to approximate the correct result. More results are provided in Appendix F.

\begin{table}[t]
\centering
\input{table/ablation_short}

\vspace*{-0.4cm}
\caption{Ablation experiments on Human3.6M. $H$=20, $K$=10.}
\vspace*{-0.2cm}
\end{table}

\begin{table}[t]
\centering
\input{table/ablation_speed}

\vspace{-0.3cm}
\caption{Accuracy\&speed. MACs are averaged over each frame. The proposed method uses the J-Agg setting.}
\label{tab:speed}
\vspace{-0.3cm}
\end{table}

\subsection{Ablation Study}
\label{sec:ablation}

We conduct ablation experiments on Human3.6M to validate the impact of each design in our method. More results are provided in Appendix D.

\paragraph{Effectiveness of each component.} As shown in Table~\ref{tab:component}, we begin with MixSTE~\cite{zhang2022mixste} as the backbone (first row), and then combine it with the diffusion models, resulting in our D3DP (second row), which achieves 1mm improvement in the single-hypothesis case. This result reveals that a simple combination of diffusion models and existing deterministic pose estimators brings a large performance gain. Besides, the network is equipped with the ability to generate multiple pose hypotheses, which fits the property of depth uncertainty in 3D pose estimation. Under P-Agg setting, MPJPE slightly decreases (40.0$\rightarrow$39.9mm) when $H$ is raised to 20. The error is further reduced to 39.5mm by using our JPMA (under J-Agg) method. This result shows that multi-hypothesis aggregation at the joint level, which allows fine-grained hypothesis selection, is superior to that at the pose level.

\paragraph{Multi-hypothesis aggregation.} We compare pose-level and joint-level multi-hypothesis aggregation methods in Table~\ref{tab:MHA}, including 1) averaging over all hypotheses; 2) using a 4-layer MLPs to generate scores for each pose/joint, which are used for a weighted average of hypotheses; 3) using our reprojection-based method at the pose level (PPMA, $\text{4}^\text{th}$ row) and joint level (JPMA, $\text{5}^\text{th}$ row). For fairness, we implement the same approach (MLPs/reprojection-based) at both levels. The results show that joint-level approaches are superior to pose-level approaches. Besides, PPMA outperforms both MLPs and average at the pose level and JPMA outperforms MLPs at the joint level, which shows that the 2D prior is effective in multi-hypothesis aggregation. Thus, our JPMA, which conducts aggregation at the joint level using 2D priors, achieves the best performance.

\paragraph{Accuracy and speed.}
Table~\ref{tab:speed} shows the accuracy and inference speed. Experiments are conducted on a single NVIDIA RTX 3080 Ti GPU. For previous methods (first four rows), MACs of inference are twice as high as those of training, due to the data augmentation applied. D3DP ($\text{5}^\text{th}$ row) makes a few modifications (add additional inputs and timestep embedding) to MixSTE (our backbone, $\text{4}^\text{th}$ row), and adds a parameter-free diffusion process. This results in a small increase in the number of parameters, essentially unchanged MACs, a slight decrease in FPS, and a 1mm performance improvement. Moreover, we can adjust $H,K$ to trade computational budgets for performance during inference. When $H,K$ increases, MACs and FPS roughly show a linear increase and decrease, respectively. 

\begin{figure}
\begin{center}
\small
\setlength{\tabcolsep}{2pt}
\begin{tabular}{c}
%{\small H36M} & {\small 3DHP} & {\small 3DPW} \\
{\includegraphics[width=\linewidth]{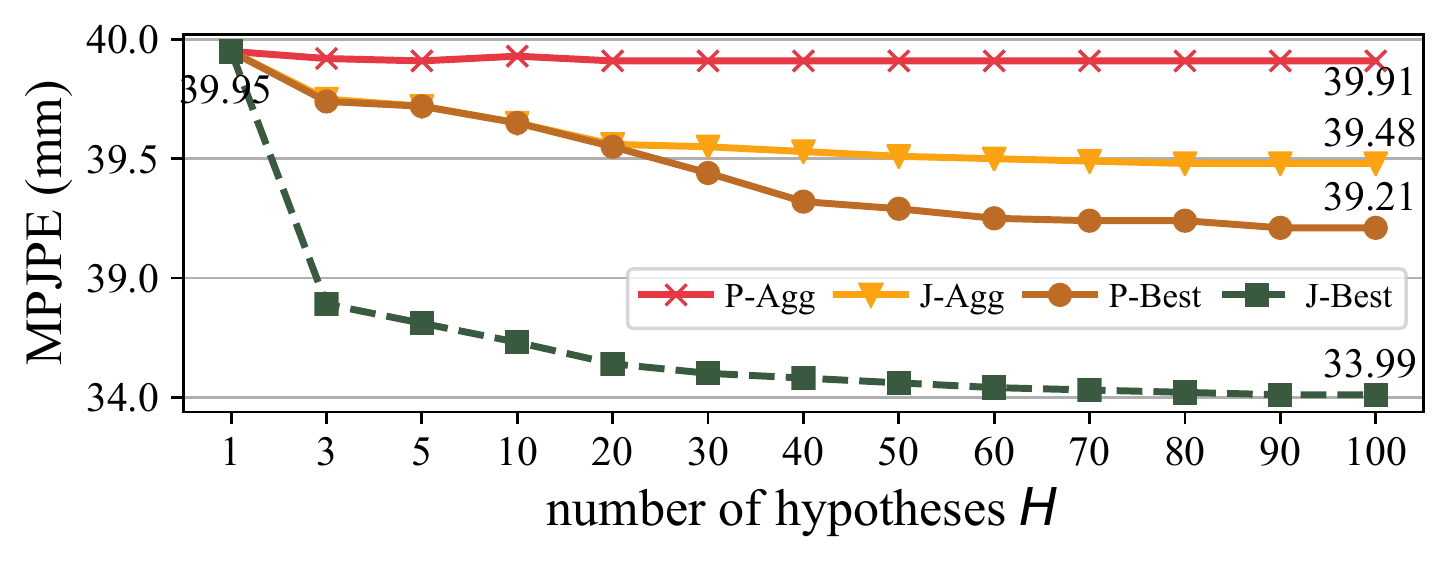}}\\[-7pt]
{\includegraphics[width=\linewidth]{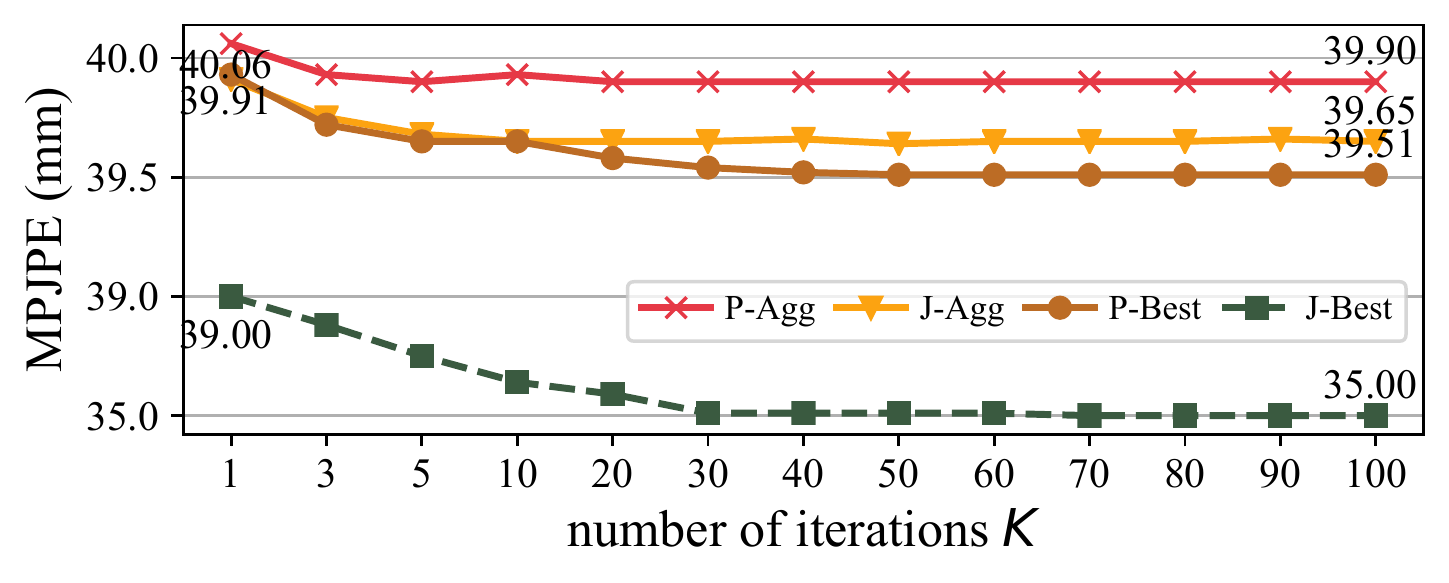}} 
 \\
\end{tabular}
%\vspace*{-0.2cm}
\end{center}
\vspace*{-0.5cm}
\caption{Ablation experiments on the number of hypotheses and iterations $H,K$. Top: $K$=10. Bottom: $H$=10.}
\vspace*{-0.2cm}
\label{fig:hk}
\end{figure}

\paragraph{Number of hypotheses and iterations.}
Fig.~\ref{fig:hk} (Top) illustrates how performance fluctuates with the number of hypotheses under the four settings. They all start with the same performance ($H$=1) and then gradually diverge. 1) P-Best and J-Best: The error under J-Best is much lower than that under P-Best, which indicates that joint-level aggregation (J-Agg) has a higher upper-bound performance than pose-level aggregation (P-Agg). 2) P-Agg and J-Agg: Average and JPMA are used as the pose-level (P-Agg) and joint-level (J-Agg) methods, respectively. MPJPE under P-Agg stays around 39.9mm. It is because as $H$ increases, the mean of all hypotheses remains roughly the same, while the variance rises (Appendix D.3). This observation indicates that the average operation counteracts the advantage of increased diversity of hypotheses, leading to no improvement in performance. The error under J-Agg decreases as $H$ rises, which demonstrates that J-Agg makes better use of the diversity than P-Agg to achieve performance gains when $H$ increases. The performance of J-Agg is very close to the upper-bound performance of P-Agg (\ie, P-Best) when $H<20$. J-Agg is practical, but P-Best is not because the latter is agnostic in real-world applications, which highlights the advantages of J-Agg. Besides, there is still a performance gap between J-Agg (39.48mm) and J-Best (33.99mm), which shows that JPMA serves as a baseline method with considerable room for improvement.

Fig.~\ref{fig:hk} (Bottom) illustrates how performance fluctuates with the number of iterations. The general trend of the performance curves for the four settings is similar to that of the above experiment. Performance under P-Agg does not vary substantially with $K$. J-Agg and P-Best diverge after $K$=10. J-Best outperforms the other three settings by a wide margin. The performance of all settings saturates at $K$=30. 

%\left(\frac{\boldsymbol{x}_t-\sqrt{1-\alpha_t} \epsilon_\theta^{(t)}\left(\boldsymbol{x}_t\right)}{\sqrt{\alpha_t}}\right)

\section{Conclusions}
This paper presents D3DP with JPMA, a diffusion-based probabilistic method with joint-wise reprojection-based multi-hypothesis aggregation, for 3D human pose estimation. D3DP, which is compatible and customizable, generates multiple pose hypotheses. It progressively adds noise to clean 3D poses and reverses this process to reconstruct the original signals. To aggregate the generated pose hypotheses into a single 3D prediction, JPMA is proposed, which conducts joint-level aggregation based on reprojection errors. Compared with existing pose-level aggregation methods, JPMA exploits the merits of each joint and thus achieves exceptional performance. Experimental results on two benchmarks show that our method surpasses the state-of-the-art probabilistic and deterministic approaches.

\appendix
%\begin{appendices}
%\section*{Acknowledgment}

\section{Preliminary: Diffusion Model}
Following the notation of~\cite{nichol2021improved,ho2020denoising,chen2022diffusiondet}, we provide a detailed review of the diffusion models, which consist of a diffusion process and a reverse process.

\textbf{Diffusion process.} The diffusion process $q$ generates corrupted samples $\bm{y}_1$, $\bm{y}_2$, ..., $\bm{y}_T$ by adding different levels of Gaussian noise to the original signal $\bm{y}_0$ at each timestep $t \in [0, T]$, which can be formulated as 
\begin{equation}
    \begin{split}
        q(\bm{y}_{1:T}|\bm{y}_0) &\coloneqq \prod^T_{t=1}q(\bm{y}_t | \bm{y}_{t-1}) \\
        q(\bm{y}_t | \bm{y}_{t-1}) &\coloneqq \mathcal{N}(\bm{y}_t; \sqrt{1-\beta_t}\bm{y}_{t-1}, \beta_t \bm{I})
    \end{split}
    \label{eq:diffusion}
\end{equation}
% \vspace{-0.7cm}
% \begin{alignat}{2}
%     q(\bm{y}_{1:T}|\bm{y}_0) &\coloneqq \prod^T_{t=1}q(\bm{y}_t | \bm{y}_{t-1}) \\
%     q(\bm{y}_t | \bm{y}_{t-1}) &\coloneqq \mathcal{N}(\bm{y}_t; \sqrt{1-\beta_t}\bm{y}_{t-1}, \beta_t \bm{I})
% \end{alignat}

\noindent where $\beta_t$ is the noise variance schedule~\cite{nichol2021improved} and $\bm{I}$ denotes the identity matrix. Following Ho~\etal~\cite{ho2020denoising}, we can rewrite Eq.~\ref{eq:diffusion} in a form that requires no iteration to yield $\bm{y}_t$ from $\bm{y}_0$:
\begin{equation}
    \begin{split}
        q(\bm{y}_t|\bm{y}_0) &\coloneqq \mathcal{N}(\bm{y}_t; \sqrt{\bar{\alpha}_t}\bm{y}_0, (1 - \bar{\alpha}_t)\bm{I}) \\
        &\coloneqq \sqrt{\bar{\alpha}_t} \bm{y}_0  + \epsilon \sqrt{1 - \bar{\alpha}_t}, \epsilon \sim \mathcal{N}(0, \bm{I})
    \end{split}
    \label{eq:diffusion1}
\end{equation}
\noindent where $\bar{\alpha}_t \coloneqq \prod_{s=0}^{t} \alpha_s$ and $\alpha_t \coloneqq 1 - \beta_t$. $\epsilon$ denotes the Gaussian noise. With a sufficiently large $T$ and reasonable noise schedule $\beta_t$, the distribution of $q(\bm{y}_T )$ is nearly an isotropic Gaussian distribution $\mathcal{N}(0, \bm{I})$. 

\textbf{Reverse process.} There are two ways to implement the reverse process $p$ using a neural network parameterized by $\theta$. One is by iterating $p_{\theta}(\bm{y}_{t-1} | \bm{y}_t)$ multiple times until $\bm{y}_0$, and the other is by getting $\bm{y}_0$ in one step directly from $p_{\theta}(\bm{y}_0 | \bm{y}_t)$.

In the first case, it is found that the posterior $q(\bm{y}_{t-1} | \bm{y}_t, \bm{y}_0)$ can be formulated as a Gaussian distribution as well using the Bayes' theorem:
\begin{equation}
    \begin{split}
        q(\bm{y}_{t-1}|\bm{y}_t, \bm{y}_0) &= \mathcal{N}(\bm{y}_{t-1};\tilde{\mu}(\bm{y}_t, \bm{y}_0), \tilde{\beta}_t \bm{I})
    \label{eq:posterior}
    \end{split}
\end{equation}

\noindent where
\begin{equation}
    \begin{split}
        \tilde{\beta}_t &\coloneqq \frac{1-\bar{\alpha}_{t-1}}{1-\bar{\alpha}_t} \beta_t
        \\
        \tilde{\mu}_t(\bm{y}_t, \bm{y}_0) &\coloneqq
        \frac{\sqrt{\bar{\alpha}_{t-1}}\beta_t}{1-\bar{\alpha}_t}\bm{y}_0 + \frac{\sqrt{\alpha_t}(1-\bar{\alpha}_{t-1})}{1-\bar{\alpha}_t} \bm{y}_t     
    \label{eq:posterior-param}
    \end{split}
\end{equation}
\noindent are the variance and mean of the Gaussian distribution, respectively. However, Eq.~\ref{eq:posterior} depends on $\bm{y}_0$ to measure the posterior, which is unknown in advance. Instead, we approximate $q(\bm{y}_{t-1}|\bm{y}_t, \bm{y}_0)$ through 
\begin{equation}
    \begin{split}
        p_{\theta}(\bm{y}_{t-1}|\bm{y}_t) &\coloneqq \mathcal{N}(\bm{y}_{t-1};\mu_{\theta}(\bm{y}_t, t), \Sigma_{\theta}(\bm{y}_t, t)) \label{eq:ptheta}
    \end{split}
\end{equation}
\noindent where 
\begin{equation}
    \begin{split}
        \Sigma_{\theta}(\bm{y}_t, t) &=\tilde{\beta}_t \bm{I}\\
        \mu_\theta(\bm{y}_t, t)  &= \frac{1}{\sqrt{\alpha_t}}\left( \bm{y}_t - \frac{\beta_t}{\sqrt{1-\bar\alpha_t}} \epsilon_\theta(\bm{y}_t, t) \right)
    \label{eq:ptheta-param}
    \end{split}
\end{equation}

\noindent where $\epsilon_\theta(\bm{y}_t, t)$ is the noise predicted by the neural network, which is supervised by
\begin{equation}
    \begin{split}    
        \mathcal{L} = \mathbb{E}_{\bm{y}_{t},t, \epsilon}[||\epsilon_{\theta}(\bm{y}_{t},t)- \epsilon||^{2} ]
    \label{eq:loss-eps}
    \end{split}
\end{equation}
\noindent where $\epsilon \sim \mathcal{N}(0, \bm{I})$.

In the second case, we can directly predict the clean data from a learned network $f_\theta(\bm{y}_t, t)$, which is supervised by 
\begin{equation}
    \begin{split}    
        \mathcal{L} = \mathbb{E}_{\bm{y}_{t},t}[||f_\theta(\bm{y}_t, t) - \bm{y}_0||^{2}]
    \label{eq:loss-x0}
    \end{split}
\end{equation}

The difference between the above two cases lies in the prediction target of the network, which in the first case is the noise $\epsilon$ at each timestep, and in the second case is the original signal $\bm{y}_0$. We choose to predict $\bm{y}_0$ in this work.

\section{Preliminary: Projective Geometry}
We refer to the process of a real-world object being photographed by a camera as the first \textit{projection}. The obtained image is passed through the 3D pose estimator to predict multiple plausible 3D hypotheses. These hypotheses are projected a second time to the camera plane by JPMA, which is called \textit{reprojection}. We introduce the reprojection function $\mathcal{P}(\cdot)$ in JPMA under two camera models: pinhole camera and distorted pinhole camera.

\paragraph{Pinhole camera.} We denote the 3D coordinate of a human joint in the camera coordinate system as $\left(X,Y,Z\right)$, and its reprojection to the camera plane as a 2D coordinate $\left(u,v\right)$. Then, the reprojection under a pinhole camera (with 4 intrinsic camera parameters) can be formulated by a perspective transformation:
\begin{equation}
    \begin{split} 
        X'&=X/Z, \quad Y'=Y/Z\\
        u&=f_x \cdot X'+c_x\\
        v&=f_y \cdot Y'+c_y
    \label{eq:pinhole}
    \end{split}
\end{equation}
where $f_x,f_y$ are the focal lengths expressed in pixel units and $(c_x,c_y)$ is the principal point that is usually at the image center.

\paragraph{Distorted pinhole camera.} Real lenses will bring distortion to the ideal pinhole camera model, resulting in a distorted pinhole camera (with 9 intrinsic camera parameters). The distortion can be categorized into two types: 1) radial distortion, which is caused by flawed radial curvature of a lens and can be approximated by three parameters $k_1,k_2,k_3$; 2) tangential distortion, which arises mainly from the tilt of a lens with respect to the image sensor array and can be approximated by two parameters $p_1,p_2$. When these two types of distortion are applied, Eq.~\ref{eq:pinhole} is extended as:
\begin{equation}
    \begin{split} 
        X'&=X/Z, \quad Y'=Y/Z\\
        X_\text{d}&=X' \cdot (d_\text{r}+d_\text{t})+p_1 r^2\\
        Y_\text{d}&=Y' \cdot (d_\text{r}+d_\text{t})+p_2 r^2\\
        u&=f_x \cdot X_\text{d}+c_x\\
        v&=f_y \cdot Y_\text{d}+c_y
    \label{eq:dpinhole}
    \end{split}
\end{equation}
where $d_\text{r}=1+k_1 r^2+k_2 r^4+k_3 r^6, r=\sqrt{X'^2+Y'^2}$ and $d_\text{t}=2p_1X'^2+2p_2Y'^2$, following the formulation in~\cite{cho2021camera}.

For calibrated cameras, we use the ground truth intrinsic camera parameters for JPMA. Otherwise, we use a network to estimate the parameters, which is discussed in Section~\ref{sec:ablation_cam}.

\begin{algorithm}[t]
\input{algorithms/training}
\end{algorithm}

\begin{algorithm}[t]
\input{algorithms/inference}
%\vspace{-1cm}
\end{algorithm}

\begin{table*}[t]
\centering
\input{table/ablation_D3DP.tex}
\vspace*{-0.2cm}
\caption{Ablation experiments of D3DP on Human3.6M to justify the selection of each component. $H$=1, $K$=1. Default settings are shown in \colorbox{bestcolor}{gray}. }
\vspace*{-0.4cm}
\end{table*}

\section{Algorithm}
%We present the specific procedure in training and testing phases.

Algorithm~\ref{alg:train} provides the pseudo-code of D3DP training procedure. First, we scale the ground truth 3D pose $\bm{y}_0$ to control the signal-to-noise ratio. Then, a diffusion process is constructed from the scaled signal to the noisy 3D pose. Next, we train a denoiser to reverse this process by using the noisy 3D pose, 2D keypoints, and the timestep to predict the clean 3D pose $\widetilde{\bm{y}}_0$. The entire framework is supervised by an MSE loss: $\mathcal{L} = ||\bm{y}_0 - \widetilde{\bm{y}}_0||_{2}$. By using this simple optimization objective, we avoid issues of sensitivity to hyperparameters (\eg, balance factor between different loss functions) and training instability encountered in previous methods~\cite{wehrbein2021probabilistic,li2020weakly}.
% \begin{equation}
%     \begin{split}    
%         \mathcal{L} = ||\bm{y}_0 - \widetilde{\bm{y}}_0||^{2}.
%     \label{eq:loss-x0}
%     \end{split}
% \end{equation}

Algorithm~\ref{alg:sample} provides the pseudo-code of D3DP inference procedure. We start by sampling $H$ initial 3D poses from a Gaussian distribution and use the number of iterations $K$ to determine the timestep for each iteration. In each iteration, noisy 3D poses are sent into the denoiser to estimate the uncontaminated 3D poses, which are used by DDIM to derive noisy inputs for the denoiser in the next iteration.

In addition, we combine D3DP with a common data augmentation scheme (\ie, horizontal flipping) in deterministic 3D pose estimation~\cite{zheng20213d,shan2022p,zhang2022mixste,pavllo20193d}, and propose \textit{diffusion-flipping}. Specifically, we horizontally flip input 2D keypoints as well as noisy 3D poses. The denoiser is then employed to produce the flipped predictions, which are flipped back and averaged with the original predictions to obtain the final 3D poses. The above steps are repeated in each iteration. Compared with previous approaches, the proposed method extends the data augmentation from once to $K$ times, which reduces the cumulative error and increases the accuracy of predictions.

\begin{table}[t]
\begin{center}
  %\centering
%\renewcommand\tabcolsep{8.0pt}
%\vspace{-0.4cm}
%\resizebox{\columnwidth}{!}{\begin{tabular}{x{30}|x{7}x{7}x{7}x{7}x{7}}
\begin{tabular}{c|ccccc}
\hline
Iteration&1&2&3&4&5\\
\hline
noise $\epsilon_t$&48.6&43.5&41.0&40.5&40.0\\
original data $\bm{y}_0$&\textbf{39.9}&\textbf{39.9}&\textbf{39.8}&\textbf{39.8}&\textbf{39.7}\\
\hline
\end{tabular}
\end{center}
\vspace{-0.4cm}
\caption{Performance of two regression targets after each iteration. MPJPE$\downarrow$ is reported. H=K=5. J-Agg is used.}
\label{tab:reg_tar}
\end{table}

\section{Additional Ablation Study}

\subsection{Components of D3DP}

We conduct more ablation experiments on Human3.6M to study the design of D3DP in detail.

\paragraph{Regression target.} We implement two alternatives: predicting the noise $\epsilon_t$ at each timestep of the reverse process or predicting the original 3D data $\bm{y}_0$. As shown in Table~\ref{tab:reg}, the latter achieves better results.

Note that the difference between our work and other concurrent diffusion-based methods~\cite{holmquist2022diffpose,gong2023diffpose,choi2022diffupose}  mainly lies in the regression target. The regression target of our model is the original 3D data $\bm{y}_0$, while theirs is the noise $\epsilon_t$ at each timestep. Table~\ref{tab:reg} shows our method outperforms theirs. Further experiments (Table~\ref{tab:reg_tar}) reveal that predicting $\bm{y}_0$ yields good performance even in early iterations, while predicting $\epsilon_t$ does not. This is because in early iterations, when the input $\bm{y}_t$ is extremely noisy, it is more effective to predict the original signal $\bm{y}_0$ directly than to obtain $\bm{y}_0$ by predicting the noise $\epsilon_t$ and then subtracting it from $\bm{y}_t$. This property is valuable for real-time processing. For example, when $K$ is fixed and computational resources are inadequate, the algorithm is required to produce predictions after the first iteration. Our method of predicting $\bm{y}_0$ still achieves satisfactory results, while theirs of predicting $\epsilon_t$ does not.

\begin{figure}[t]
\begin{center}
\includegraphics[width=\linewidth]{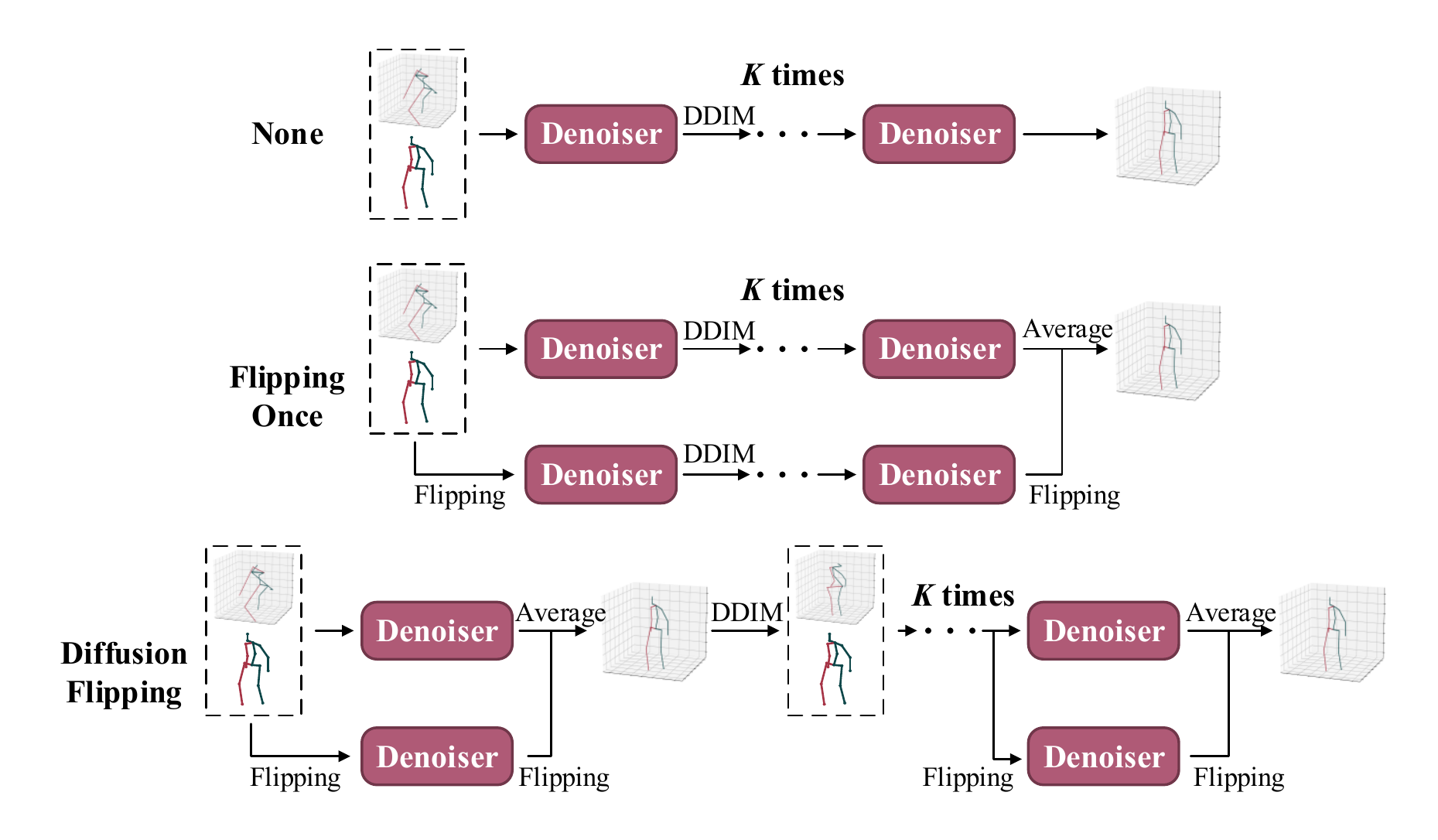}
\end{center}
\vspace*{-0.4cm}
\caption{Detailed architectures of three data augmentation approaches.}
%\vspace*{-0.4cm}
\label{fig:da}
\end{figure}

\paragraph{Location of the timestep embedding.} We add the timestep embedding to the network in a similar way as the positional embedding~\cite{vaswani2017attention}. Table~\ref{tab:timestep} shows that adding it to the first layer of the network performs the same as all layers, hence the former is chosen for simplicity. Experimental results demonstrate that timestep embedding is crucial to the denoising process.

\paragraph{Data augmentation.} Three data augmentation approaches are compared in Table~\ref{tab:da}, including 1) no augmentation; 2) flipping-once, which flips the input, conducts denoising for $K$ times, and flips the prediction again. The flipped prediction is then averaged with the unflipped prediction in the original branch to yield the final output; 3) diffusion-flipping, which applies the \textit{flip-denoise-flip} process to each timestep ($K$ times). The detailed architectures of these three approaches are shown in Fig.~\ref{fig:da}. Our diffusion-flipping achieves the best results because it averages the 3D poses of the original and flipped branches at each timestep, preventing the accumulation of errors. Other concurrent diffusion-based methods~\cite{holmquist2022diffpose,gong2023diffpose,choi2022diffupose} don't use any augmentation or use the flipping-once method. Therefore, our model is more effective than theirs.

\paragraph{2D conditioning.} 
As shown in Table~\ref{tab:2Dcond}, we evaluate multiple fusion methods (concatenation, addition, and cross attention~\cite{vaswani2017attention}) in two fusion types (input fusion and embedding fusion). For embedding fusion, two additional spatio-temporal Transformer layers are used to extract 2D features, after which these features are fused into the denoiser. The best fusion approach is concatenating noisy 3D poses and 2D conditions on the input side, which provides a fast and effective way to modify existing 3D pose estimators to fit the diffusion framework.

\paragraph{Maximum number of timesteps.} 
Table~\ref{tab:max_timesteps} indicates that the best performance can be achieved by setting an appropriate maximum number of timesteps. When $T$ is too small, we cannot diffuse the ground truth 3D poses to a Gaussian distribution during training, so the denoiser has trouble recovering a clean pose from Gaussian noise during inference. When $T$ is too large, excessive samples become pure noise after diffusion. Then, the training process of the denoiser is affected and the denoiser cannot generalize well to 3D poses with varying levels of noise (\ie, 3D poses at different timesteps) during inference.

\paragraph{Compatibility.}  The denoiser has the compatibility to use existing 3D human pose estimators as the backbone. We run our pipeline on a few other 3D estimators~\cite{shan2022p,zheng20213d,li2022exploiting,pavllo20193d}. Table~\ref{tab:backbone} shows that our approach achieves performance gains over different backbone networks, which verifies the compatibility and versatility.

\begin{table}[t]\small
\begin{center}
  %\centering
%\renewcommand\tabcolsep{10.0pt}
%\vspace{-0.4cm}
\resizebox{\linewidth}{!}{\begin{tabular}{c|cccc}
\hline
Backbone&P-STMO~\cite{shan2022p}&PoseFormer~\cite{zheng20213d}&STE~\cite{li2022exploiting}&TCN~\cite{pavllo20193d}\\
\hline
w/o diffusion&43.7&47.6&44.6&46.4\\
w/ diffusion&\textbf{42.2}&\textbf{45.7}&\textbf{43.8}&\textbf{44.5}\\
%w/ diff.&42.2&xx&xx&xx\\
\hline
\end{tabular}}
\end{center}
\vspace{-0.4cm}
\caption{Performance using other 3D estimators as backbone. MPJPE$\downarrow$ is reported. H=K=1.}
\label{tab:backbone}
\vspace{-0.2cm}
\end{table}

\begin{table}[t]
%\centering
\begin{center}
  %\centering
  %\renewcommand\tabcolsep{4.0pt}
%\vspace{-0.1cm}
%\vspace{-0.4cm}
\begin{tabular}{ccc}
\hline
Camera Model&Access&MPJPE$\downarrow$\\
\hline
w/ distortion&GT&39.74\\
w/o distortion&GT&39.80\\
w/ distortion&estimated&39.78\\
w/o distortion&estimated&39.82\\
\hline
\end{tabular}
\end{center}
\vspace{-0.4cm}
%\vspace{-0.6cm}
\caption{Impact of different camera models and different ways of accessing intrinsic camera parameters. $H$=5, $K$=5. GT: ground truth. J-Agg setting is used.}
\label{tab:cam_param}
\vspace{-0.2cm}
\end{table}

\subsection{Intrinsic Camera Parameters in JPMA}
\label{sec:ablation_cam}
As shown in Table~\ref{tab:cam_param}, we investigate the effect of two factors on the performance of JPMA: 1) the camera model, including pinhole camera (w/o distortion) or distorted pinhole camera (w/ distortion); 2) the way of accessing the intrinsic camera parameters, including using the ground truth (GT) or using a 2-layer MLPs to estimate the parameters (estimated). When the ground truth intrinsic camera parameters are used, the ideal pinhole camera model without distortion shows a performance degradation of 0.06mm compared with the case with distortion. When a neural network is utilized to estimate the parameters (the ground truth is not available), the performance drops by only 0.04mm and 0.02mm in distorted and distortion-free cases, respectively. These results indicate that the proposed JPMA method is robust to the noise caused by incorrect camera models or inaccurate estimations of intrinsic camera parameters.

\begin{figure}[t]
\begin{center}
\includegraphics[width=\linewidth]{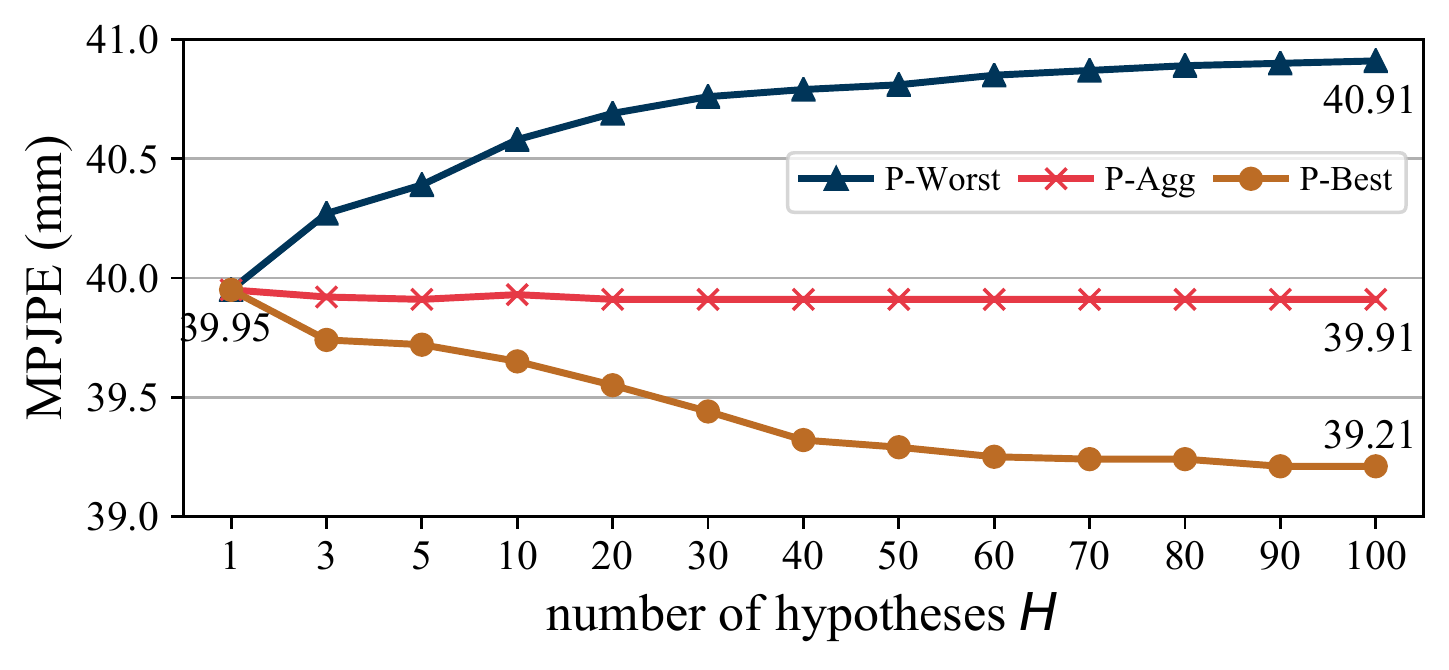}
\end{center}
\vspace*{-0.4cm}
\caption{Effect of different numbers of hypotheses $H$ on the performance of the best, worst, and average hypotheses. $K$=10.}
%\vspace*{-0.4cm}
\label{fig:h_var}
\end{figure}

\subsection{Variance of Hypotheses}
\label{variance}
Fig.~\ref{fig:h_var} shows how the performance of the best, worst, and average hypotheses changes with the number of hypotheses $H$. As $H$ increases, the performance of the average hypothesis remains essentially unchanged, with the worst hypothesis getting worse and the best hypothesis getting better. This result validates the statement in the main paper that the mean of all hypotheses remains roughly the same (the error stays around 39.9mm), while the variance rises.

\section{Additional Quantitative Results}
Table~\ref{tab:h36m_pmpjpe} provides quantitative comparisons between our D3DP with JPMA and the state-of-the-art approaches on Human3.6M when P-MPJPE is reported using 2D keypoints obtained from 2D detectors as inputs. Table~\ref{tab:h36m_gt} shows the results when MPJPE is reported using ground truth 2D keypoints as inputs. Without bells and whistles, D3DP transforms an existing deterministic 3D pose estimator into a probabilistic version with simple modifications and achieves considerable performance gains. Our method produces favorable results under conventional pose-level settings (P-Agg and P-Best), and the performance is further enhanced under the proposed joint-level settings (J-Agg and J-Best), which demonstrates the effectiveness of disentangling the hypothesis at the joint level. Experimental results show that the proposed method surpasses the others by a wide margin.

\begin{table*}[t]
\centering
\input{table/h36m_pmpjpe_new.tex}
\vspace{-0.4cm}
\caption{Results on Human3.6M in millimeters under P-MPJPE. $N,H,K$: the number of input frames, hypotheses, and iterations of the proposed D3DP. ($\ddagger$) - Our implementation. ($\sharp$) - Not feasible in real-world applications. (*) - Use CPN ~\cite{chen2018cascaded} as the 2D keypoint detector to generate the inputs. \textcolor{red}{Red}: Best. \textcolor{blue}{Blue}: Second best. \colorbox{bestcolor}{Gray}: our method.}
\label{tab:h36m_pmpjpe}
%\vspace{-0.4cm}
\end{table*}

\begin{table*}[t]
\centering
\input{table/h36m_gt_new.tex}
\vspace{-0.4cm}
\caption{Results on Human3.6M in millimeters under MPJPE. $N,H,K$: the number of input frames, hypotheses, and iterations of the proposed D3DP. ($\ddagger$) - Our implementation. ($\sharp$) - Not feasible in real-world applications. The ground truth 2D keypoints are used as inputs. \textcolor{red}{Red}: Best. \textcolor{blue}{Blue}: Second best. \colorbox{bestcolor}{Gray}: our method.}
\label{tab:h36m_gt}
%\vspace{-0.4cm}
\end{table*}

\section{Additional Qualitative Results}
\subsection{Different Numbers of Hypotheses and\\ Iterations}
\begin{figure*}
\begin{center}
\small
\setlength{\tabcolsep}{2pt}
\begin{tabular}{ccc}
%{\small H36M} & {\small 3DHP} & {\small 3DPW} \\
{\includegraphics[width=0.3\textwidth]{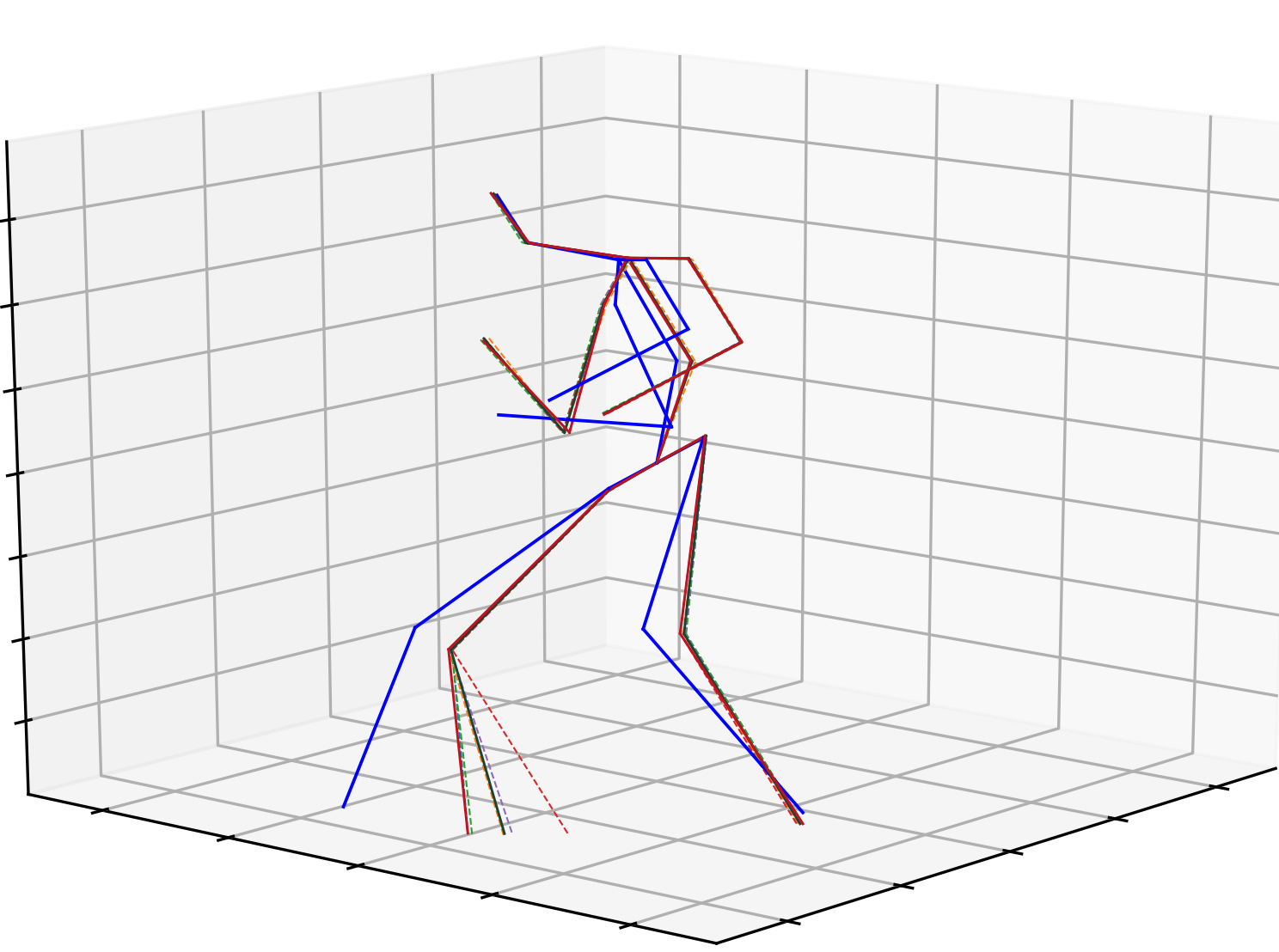}} &
{\includegraphics[width=0.3\textwidth]{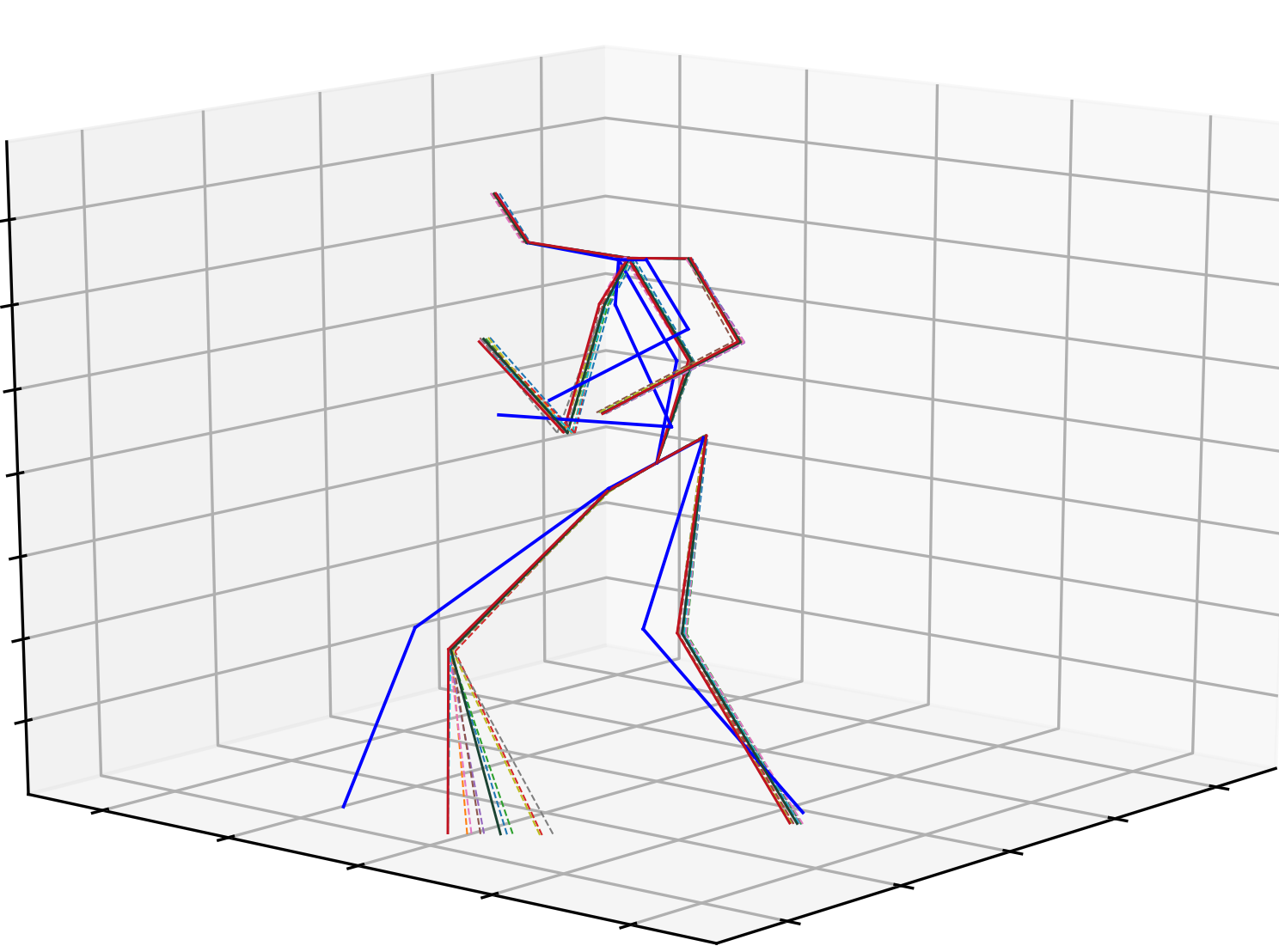}} &
{\includegraphics[width=0.3\textwidth]{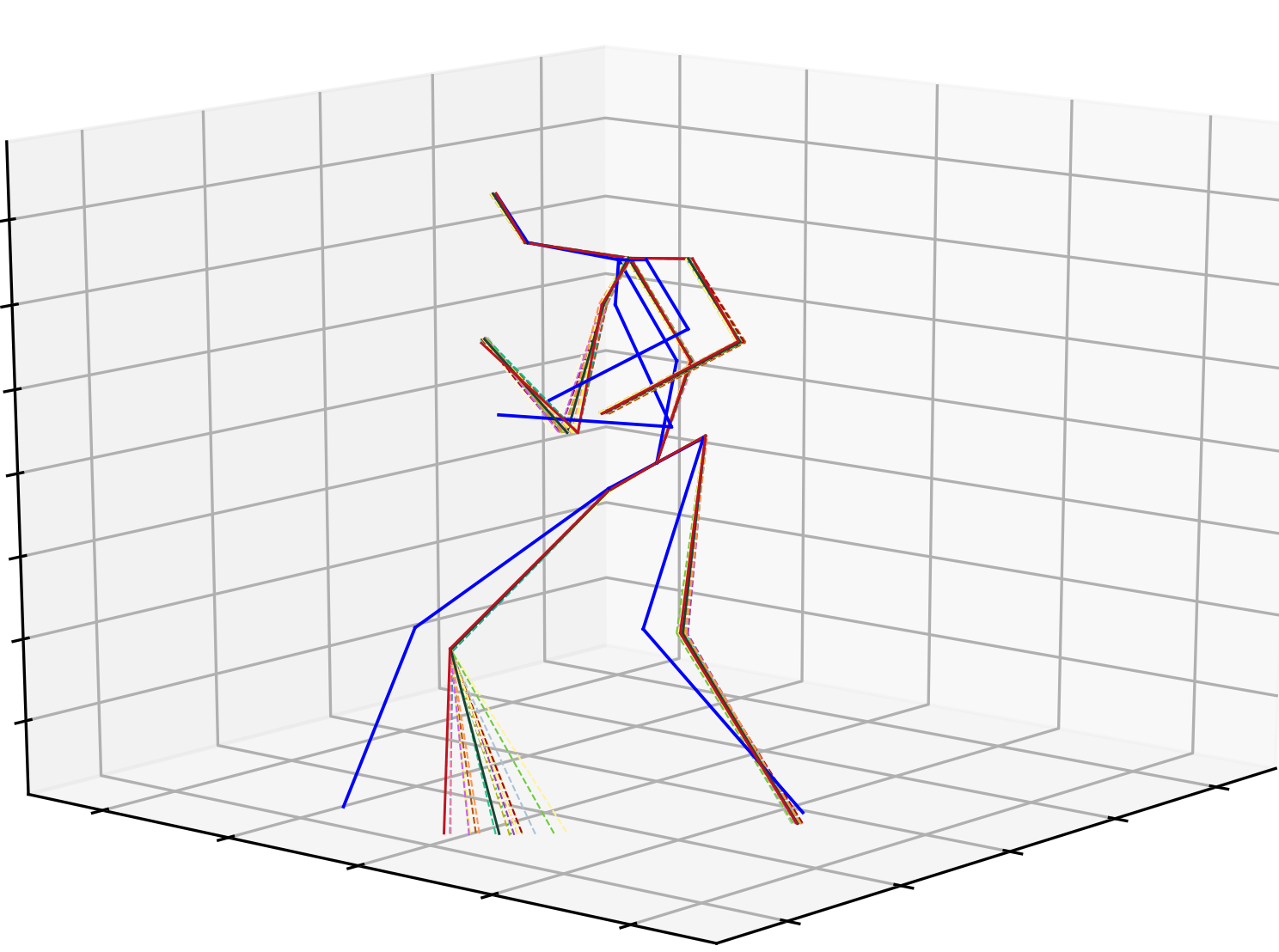}} 
\\
\multicolumn{1}{c}{\small $H$=5,$K$=5,$k$=4}&\multicolumn{1}{c}{\small $H$=10,$K$=5,$k$=4}&\multicolumn{1}{c}{\small $H$=20,$K$=5,$k$=4}\\

{\includegraphics[width=0.3\textwidth]{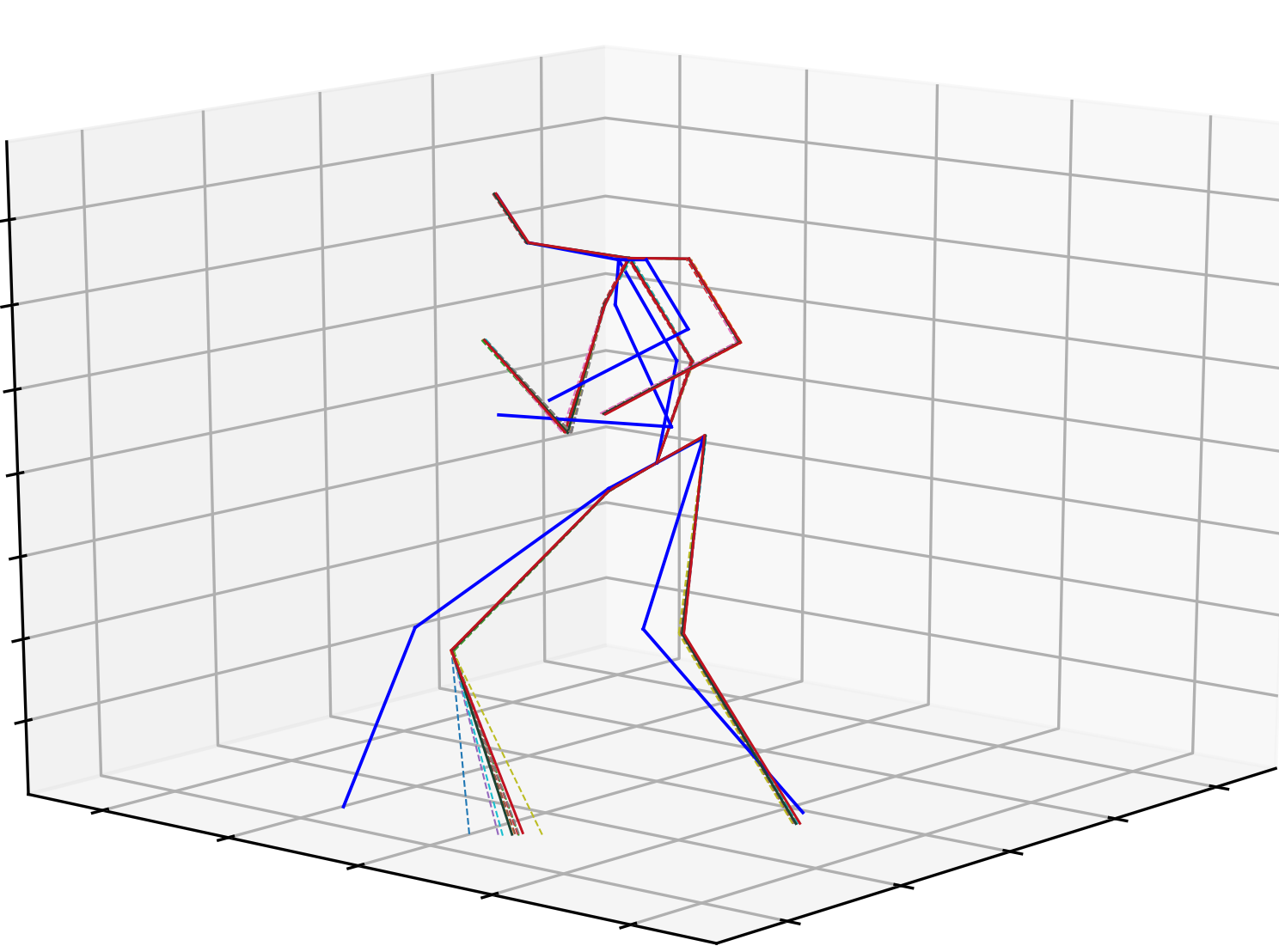}} &
{\includegraphics[width=0.3\textwidth]{figure/hk_vis/H10K5.png}} &
{\includegraphics[width=0.3\textwidth]{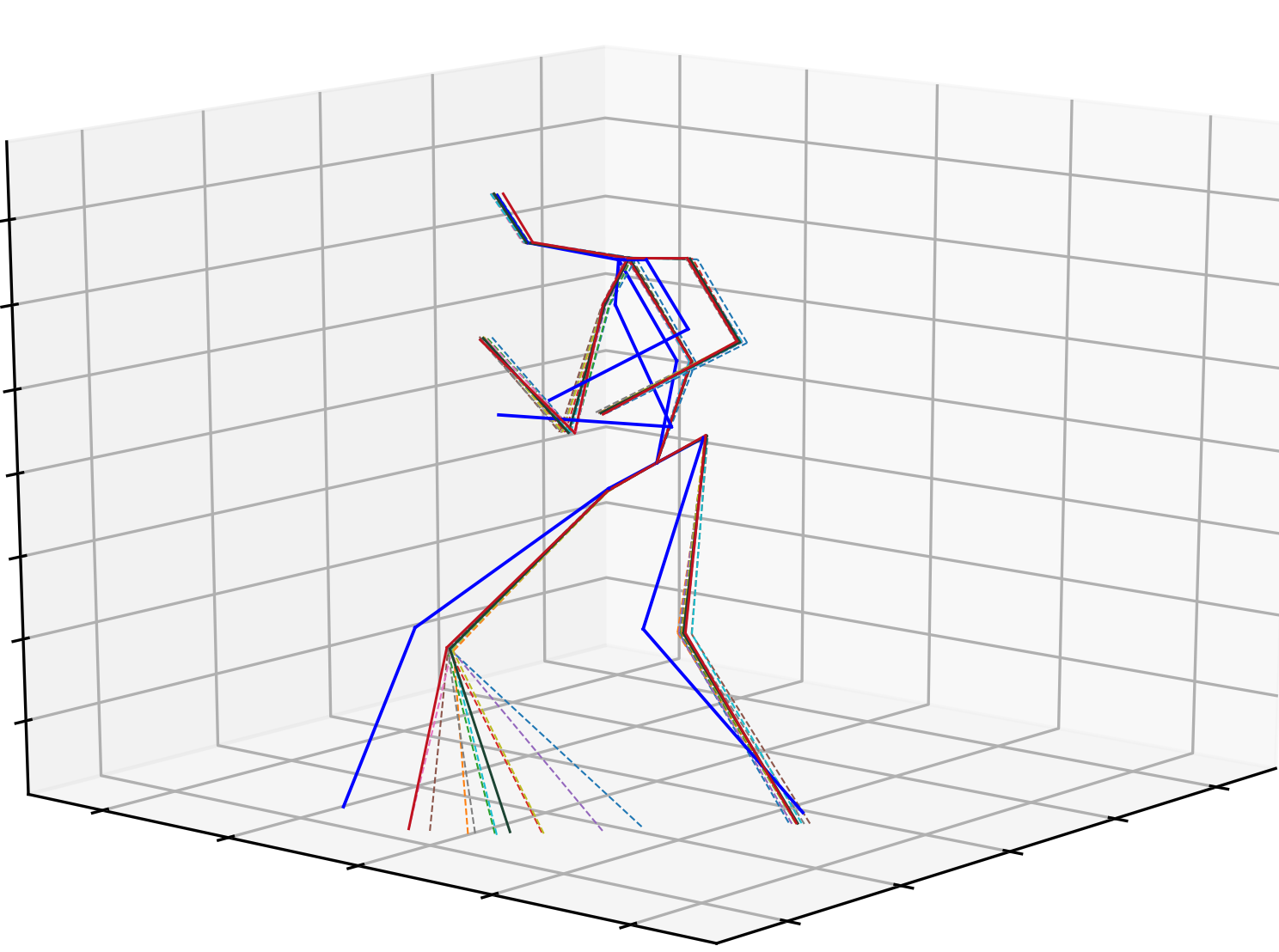}} 
\\
\multicolumn{1}{c}{\small $H$=10,$K$=2,$k$=1}&\multicolumn{1}{c}{\small $H$=10,$K$=5,$k$=4}&\multicolumn{1}{c}{\small $H$=10,$K$=10,$k$=9}\\

{\includegraphics[width=0.3\textwidth]{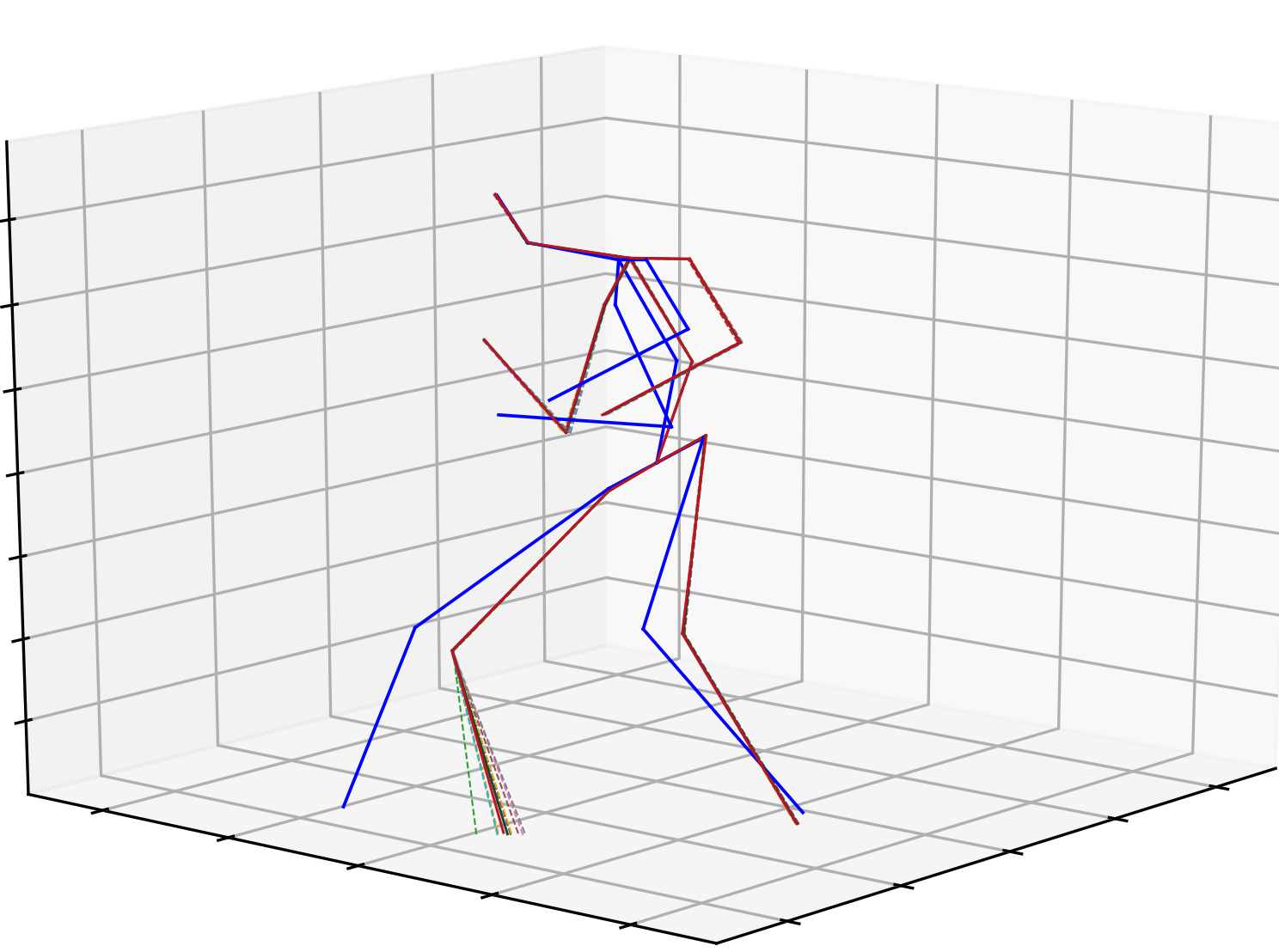}} &
{\includegraphics[width=0.3\textwidth]{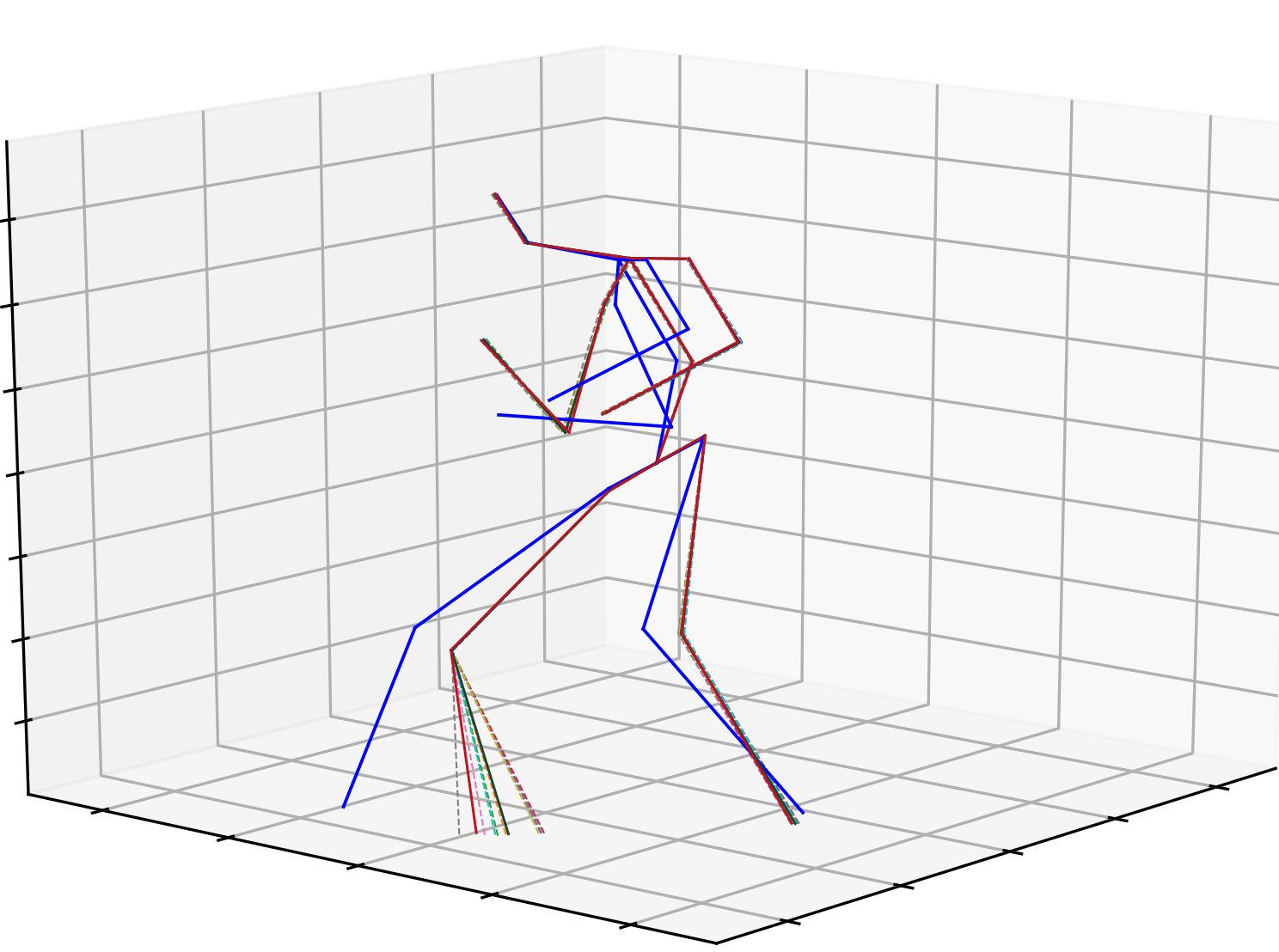}} &
{\includegraphics[width=0.3\textwidth]{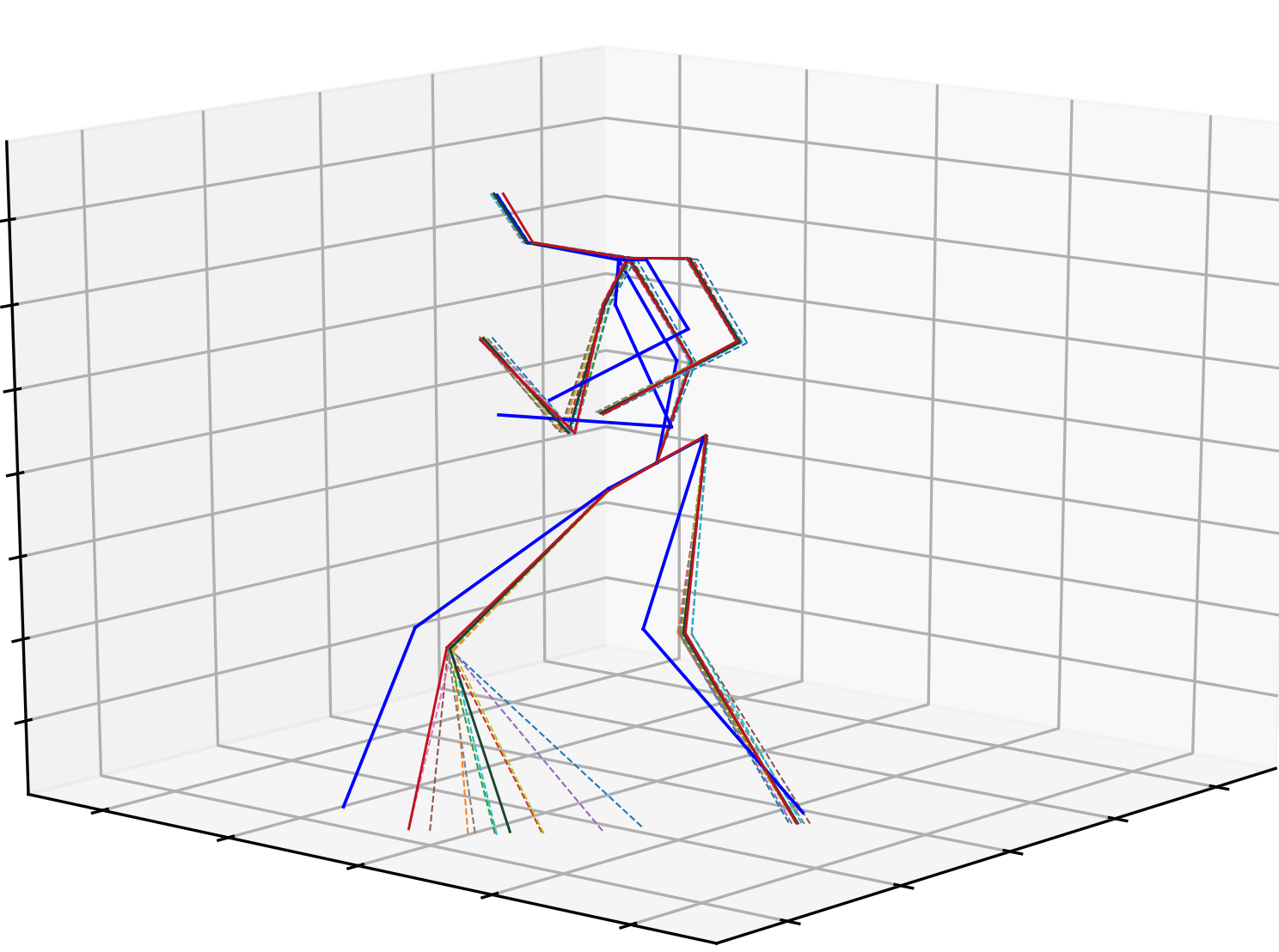}} 
\\
\multicolumn{1}{c}{\small $H$=10,$K$=10,$k$=0}&\multicolumn{1}{c}{\small $H$=10,$K$=10,$k$=5}&\multicolumn{1}{c}{\small $H$=10,$K$=10,$k$=9}\\

% {\includegraphics[width=0.3\textwidth]{figure/hk_vis/H5K5.png}} &
% {\includegraphics[width=0.3\textwidth]{figure/hk_vis/H10K5.png}} &
% {\includegraphics[width=0.3\textwidth]{figure/hk_vis/H20K5.png}} 
% \\
% \multicolumn{1}{c}{\small $H$=5}&\multicolumn{1}{c}{\small $H$=10}&\multicolumn{1}{c}{\small $H$=20}\\

\end{tabular}
%\vspace*{-0.2cm}
\end{center}
\vspace*{-0.2cm}
\caption{Qualitative results under different numbers of hypotheses $H$ and iterations $K$. Top: $H$ varies, when $K$=5. The results are the outputs of the last iteration, \ie, $k$=4 for all three subfigures. Middle: $K$ varies, when $H$=10. The results are the outputs of the last iteration, \ie, $k$=1, 4, 9 for three subfigures respectively. Bottom: $H$=10, $K$=10. The results are the outputs of the first, intermediate, and last iterations, \ie, $k$=0, 5, 9 for three subfigures respectively. Dashed line: predicted 3D pose hypotheses. Each color represents an individual hypothesis. Solid blue line: ground truth 3D poses. Solid red line: the final prediction obtained by using JPMA as the aggregation method. Solid green line: the final prediction obtained by using average as the aggregation method. }
\vspace*{-0.2cm}
\label{fig:qualitative_hk}
\end{figure*}

We show the qualitative results under different numbers of hypotheses $H$ and iterations $K$ in Fig.~\ref{fig:qualitative_hk}. When $H$ increases (first row), the mean of hypotheses is basically unchanged while the variance increases, which is consistent with the conclusion in Section~\ref{variance}. When $K$ increases (second row), the variance also raises gradually, meaning that the diversity of hypotheses is improved. This is because DDIM re-adds different noise to the predicted 3D poses to generate inputs for the next iteration, resulting in a progressive increase in the gap between different hypotheses as $K$ grows. When the variance is small, the 2D reprojections of all hypotheses are close to each other, which may affect the performance of JPMA (middle left: the solid red line is not better than the green one in the left foot region). When $H,K$ are fixed (third row), we show the results of the 3D hypotheses estimated in each iteration. As $k$ increases, the diversity of hypotheses is improved and the advantage of JPMA over average becomes apparent.

\subsection{In-the-Wild Videos}
We train our method on Human3.6M dataset and evaluate on in-the-wild videos from 3DPW~\cite{vonMarcard2018}, Penn Action~\cite{zhang2013actemes}, JHMDB~\cite{Jhuang:ICCV:2013}, and youtube. We use AlphaPose~\cite{fang2017rmpe} as the 2D keypoint detector to generate 2D poses. As shown in Fig.~\ref{fig:in_the_wild}, our method achieves satisfactory performance in most of the frames. For the cases of severe occlusion ($\text{3}^\text{rd}$ row), fast motion ($\text{5}^\text{th}$ row), low illumination ($\text{7}^\text{th}$ row), and rare poses ($\text{8}^\text{th}$ row), the proposed method generates highly uncertain predictions to represent the possible locations of 3D poses. Besides, our method can also generalize to animations ($\text{8}^\text{th}$ row) and monkey poses ($\text{9}^\text{th}$ row) because 2D keypoints are used as inputs, excluding the interference of complex textures. Theoretically, our method can be applied to any type of humanoid 3D pose estimation.

\subsection{Analysis of Failure Cases}
\begin{figure*}
\begin{center}
\small
\setlength{\tabcolsep}{2pt}
\begin{tabular}{cccccc}
%{\small H36M} & {\small 3DHP} & {\small 3DPW} \\
{\includegraphics[width=0.12\textwidth]{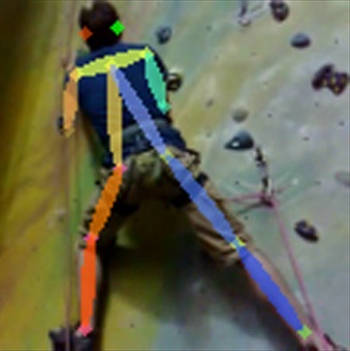}} &
{\includegraphics[width=0.18\textwidth]{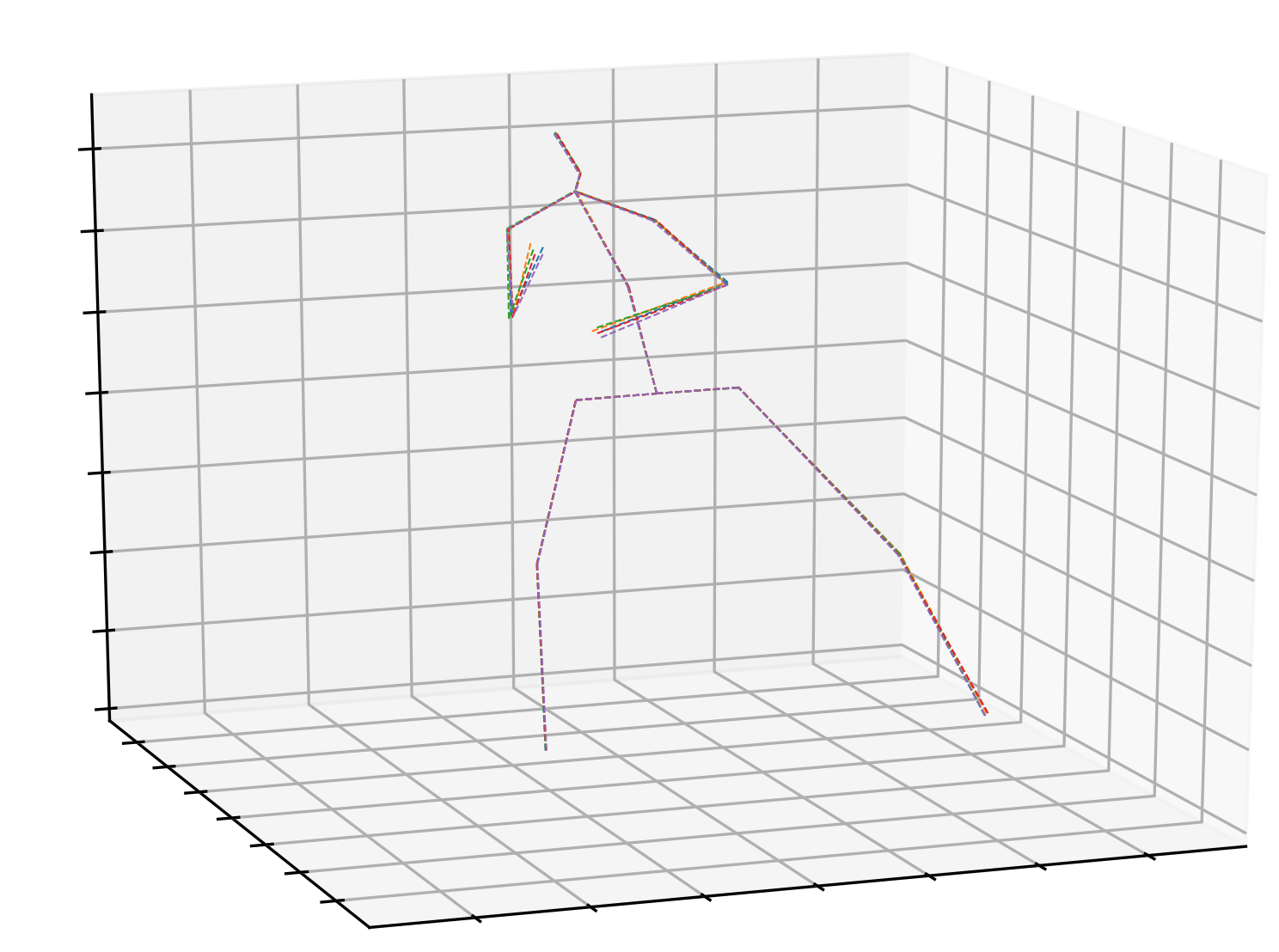}} &
{\includegraphics[width=0.12\textwidth]{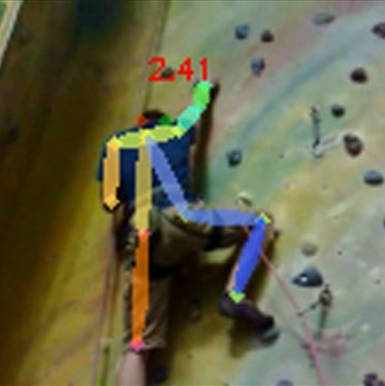}} &
{\includegraphics[width=0.18\textwidth]{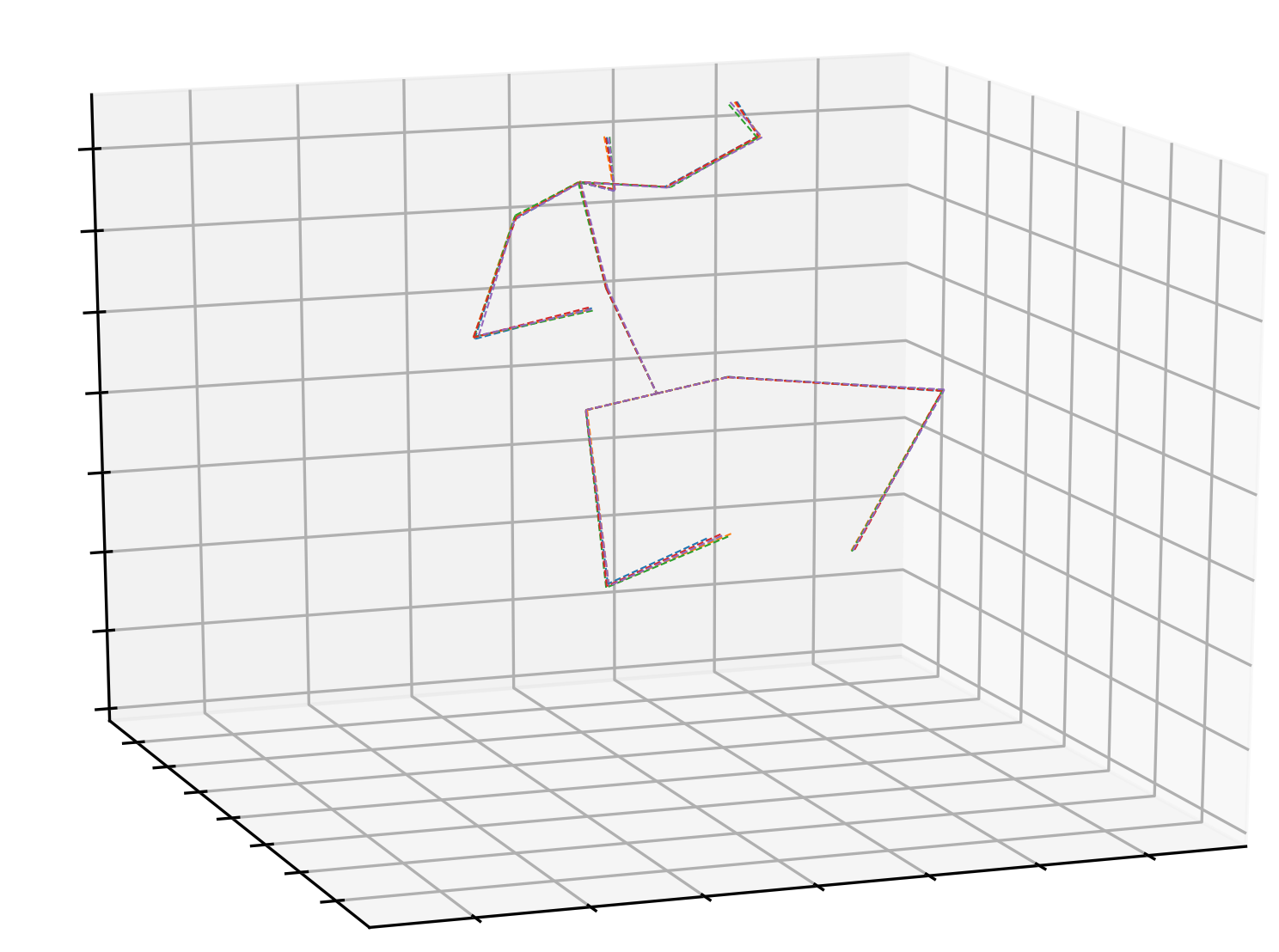}} &
{\includegraphics[width=0.12\textwidth]{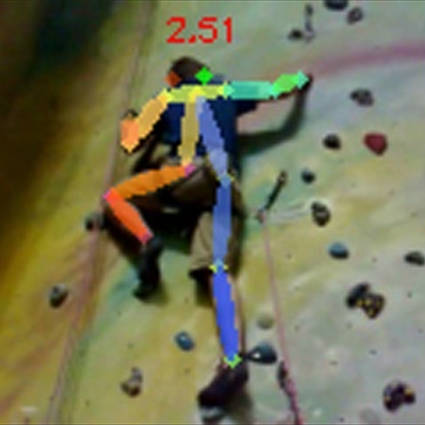}} &
{\includegraphics[width=0.18\textwidth]{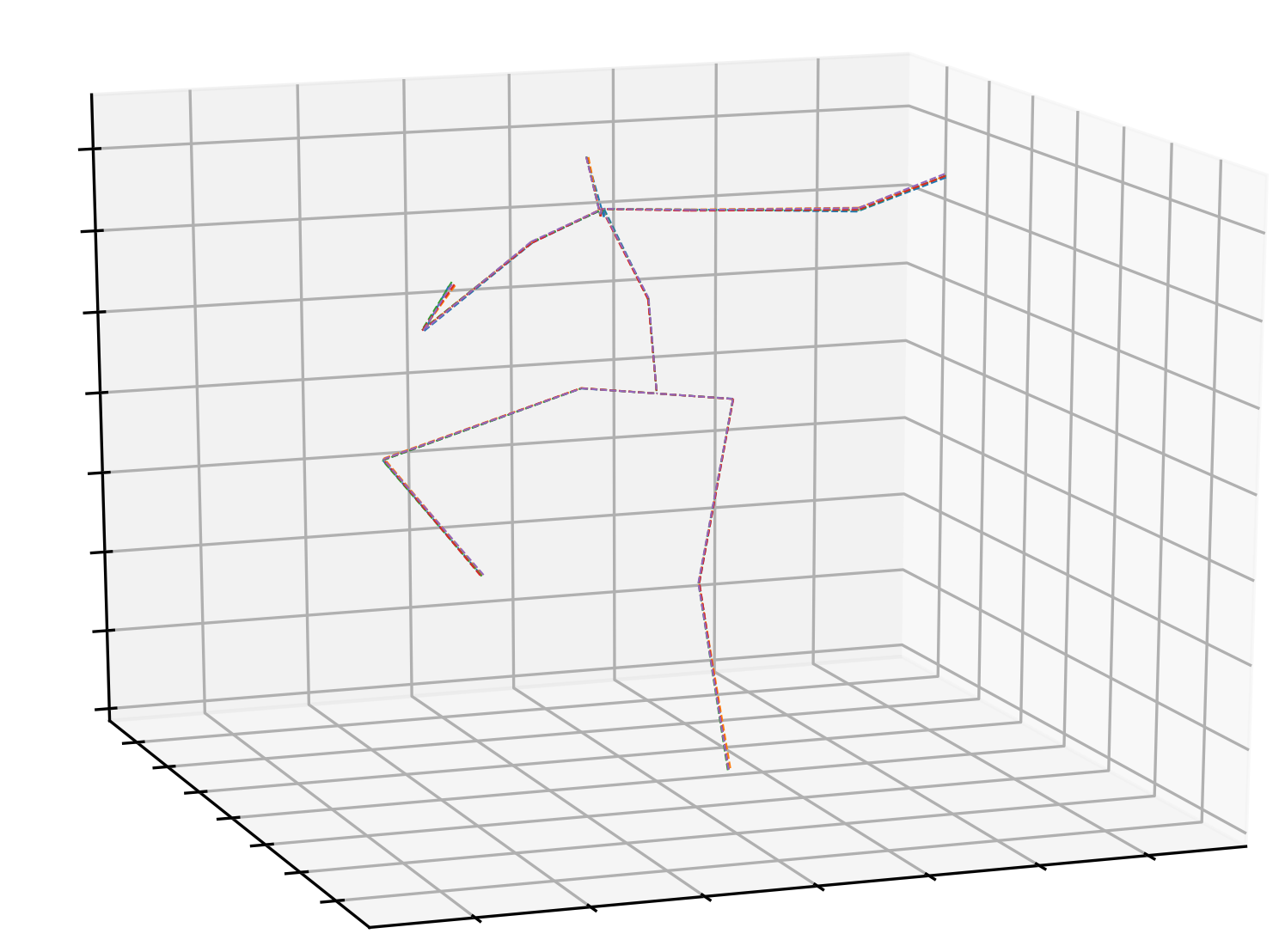}} \\

{\includegraphics[width=0.12\textwidth]{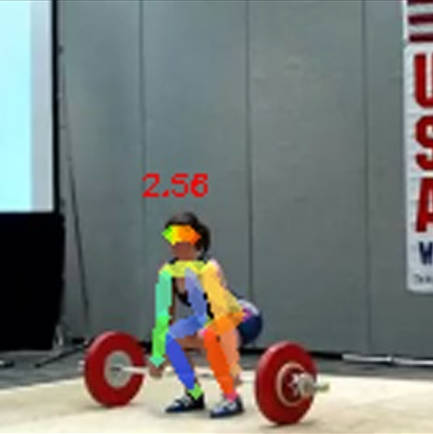}} &
{\includegraphics[width=0.18\textwidth]{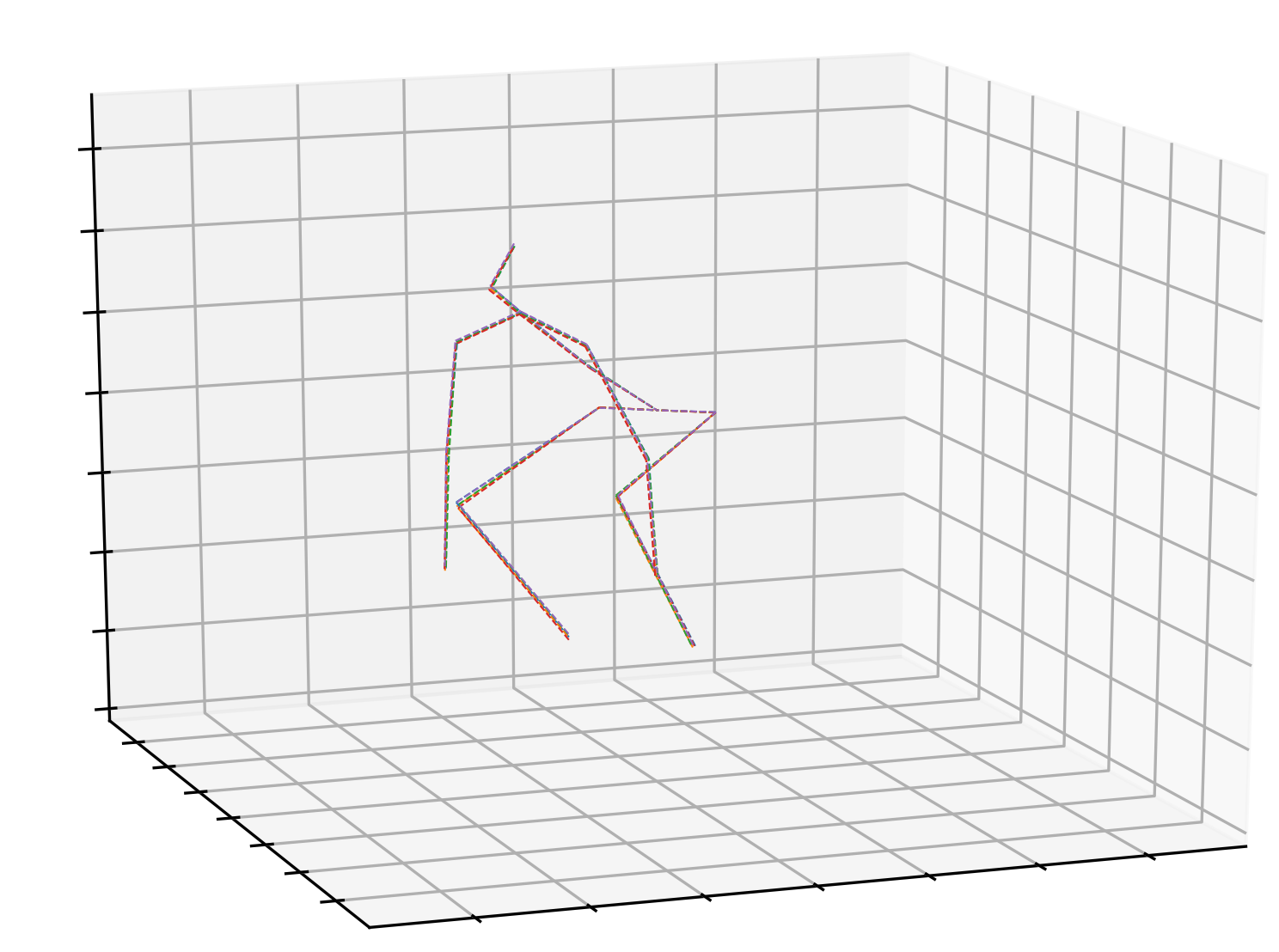}} &
{\includegraphics[width=0.12\textwidth]{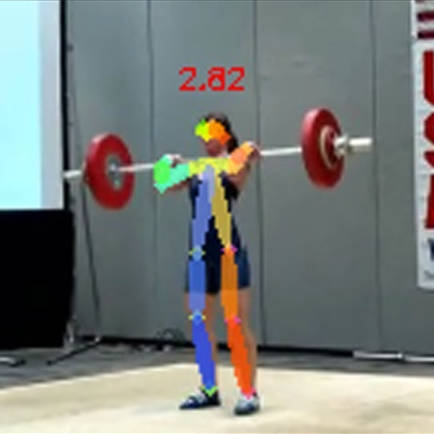}} &
{\includegraphics[width=0.18\textwidth]{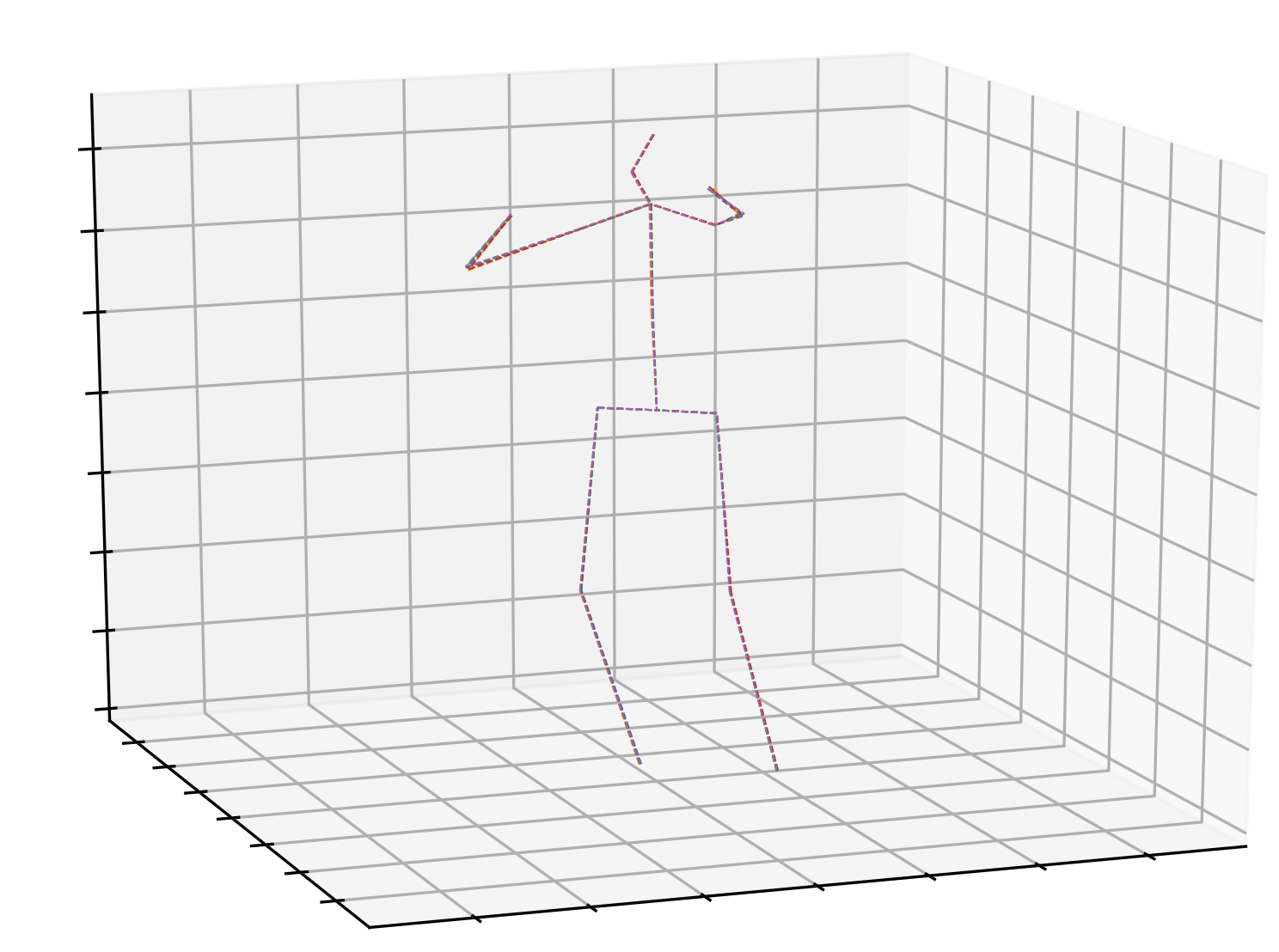}} &
{\includegraphics[width=0.12\textwidth]{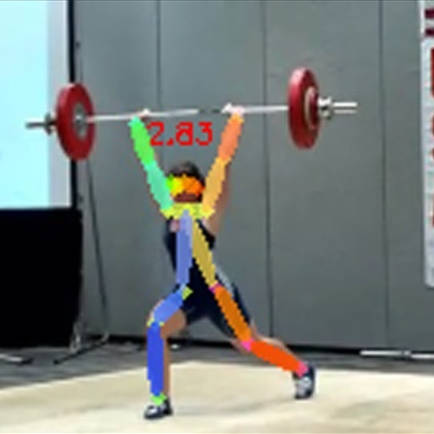}} &
{\includegraphics[width=0.18\textwidth]{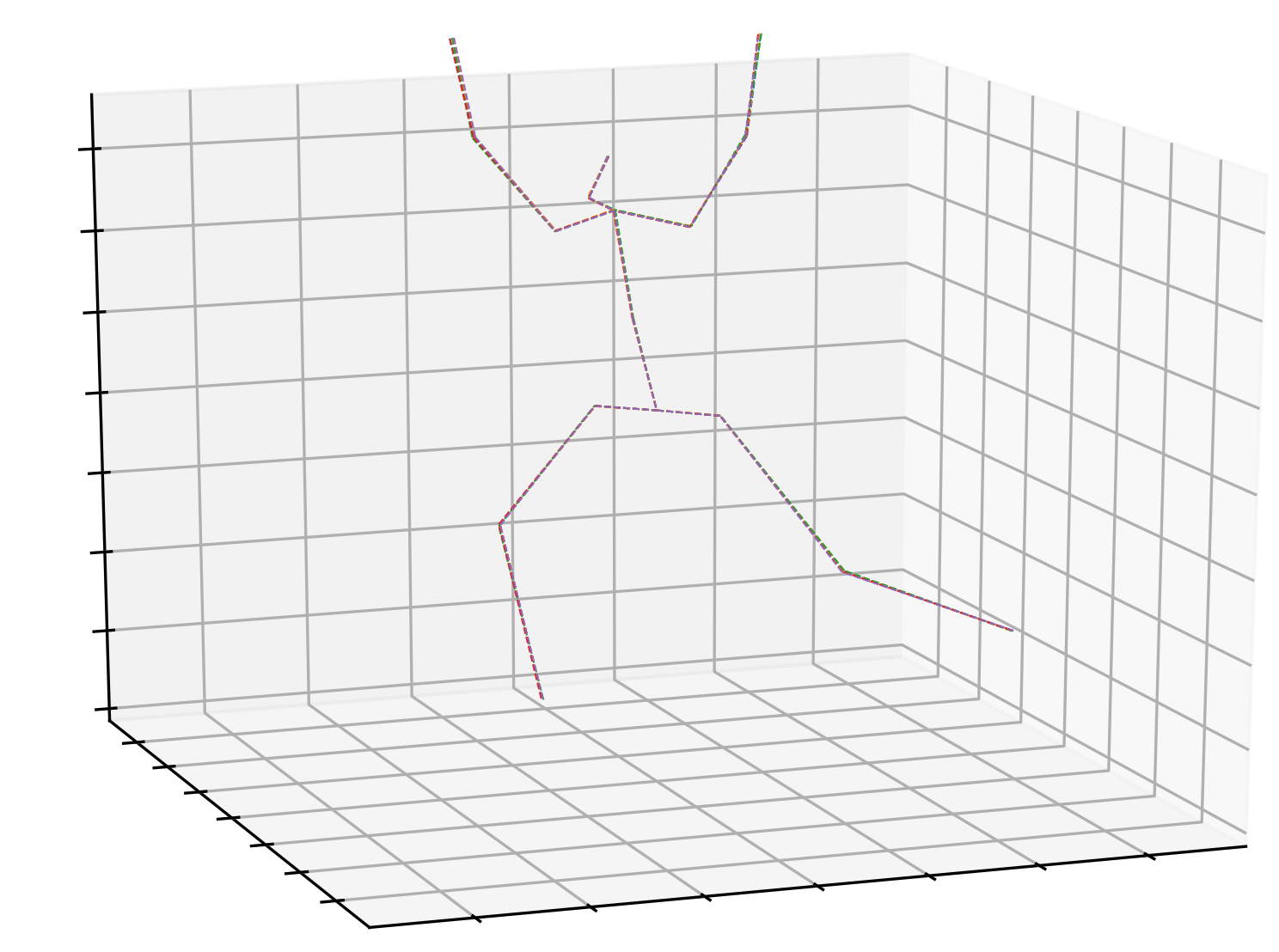}} \\

{\includegraphics[width=0.12\textwidth]{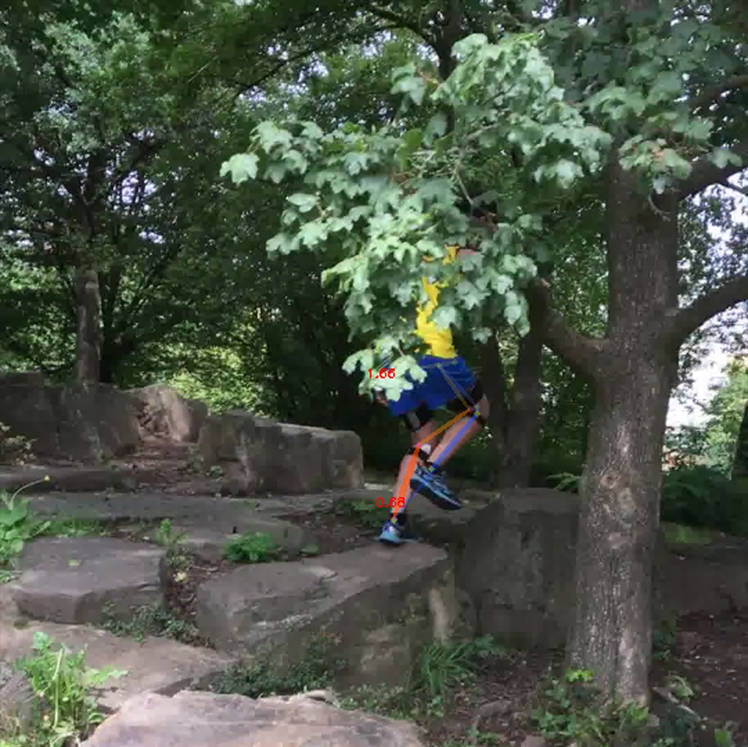}} &
{\includegraphics[width=0.18\textwidth]{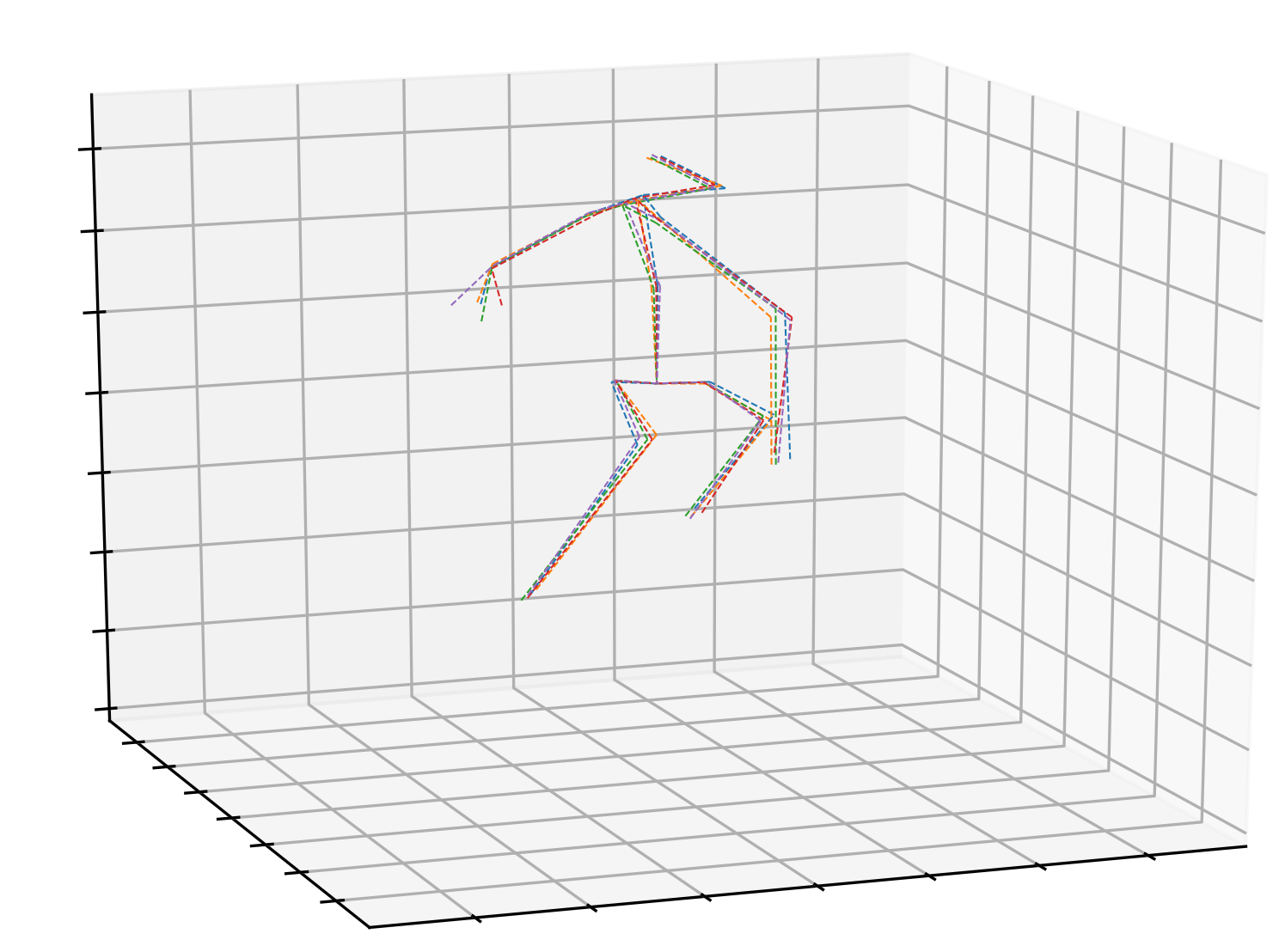}} &
{\includegraphics[width=0.12\textwidth]{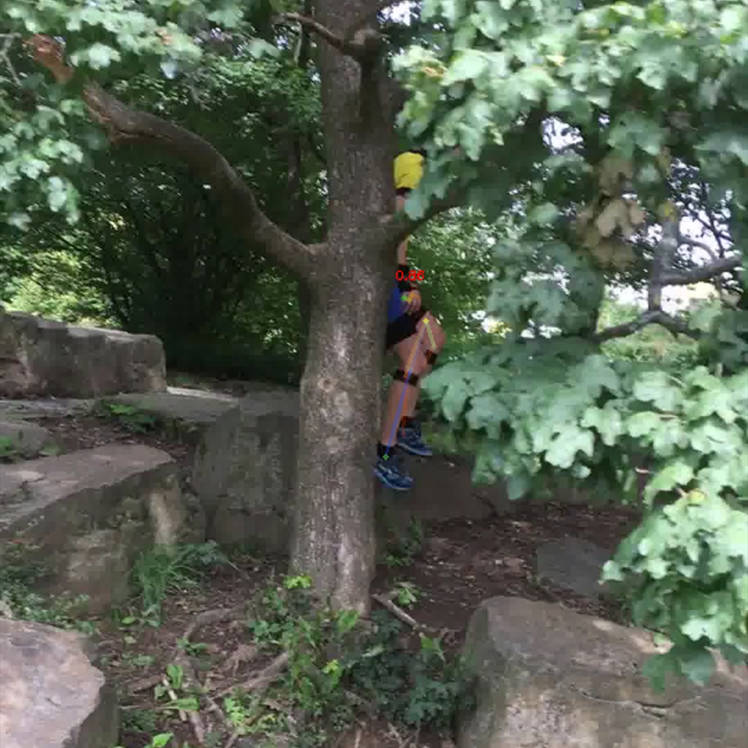}} &
{\includegraphics[width=0.18\textwidth]{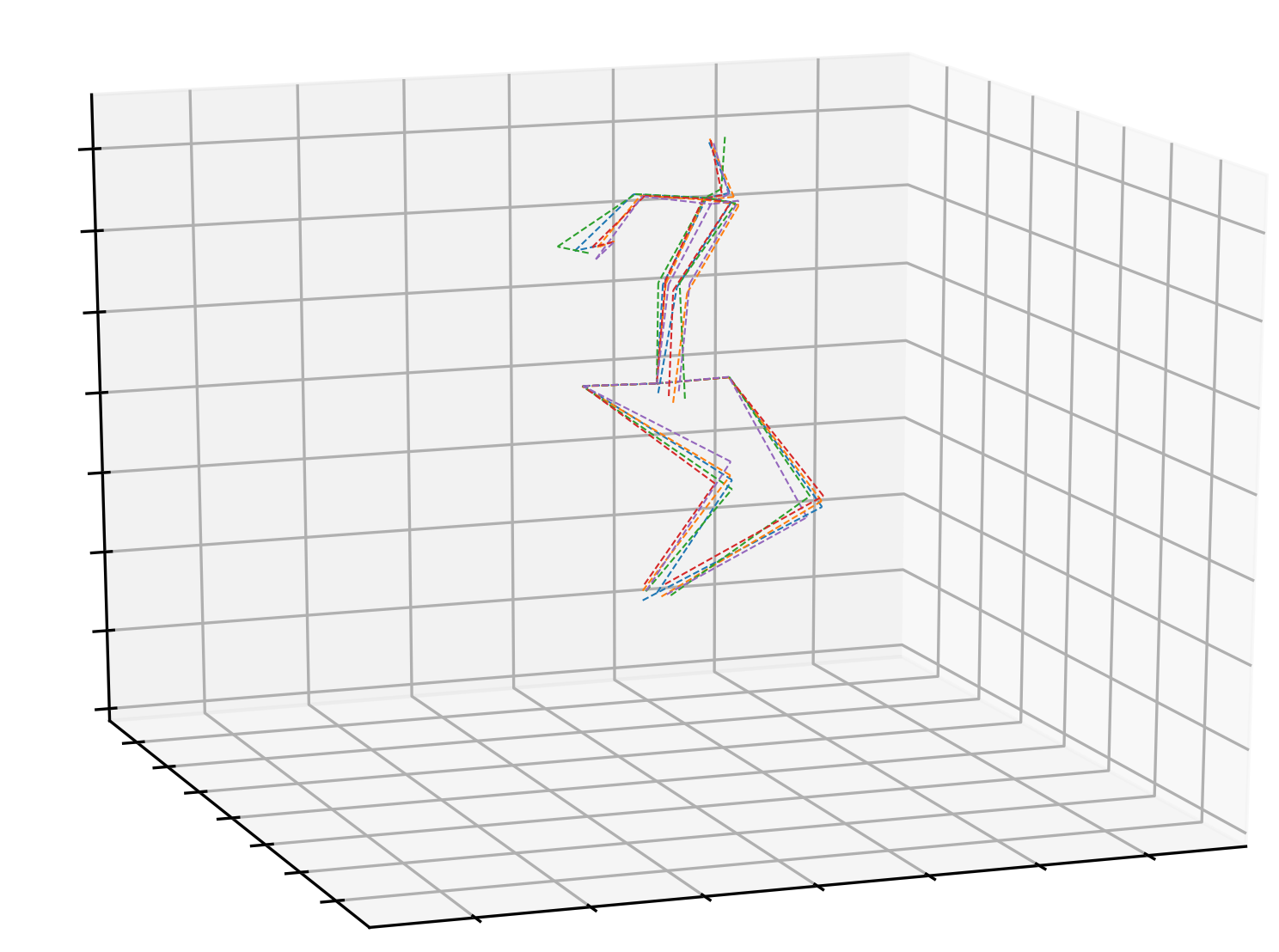}} &
{\includegraphics[width=0.12\textwidth]{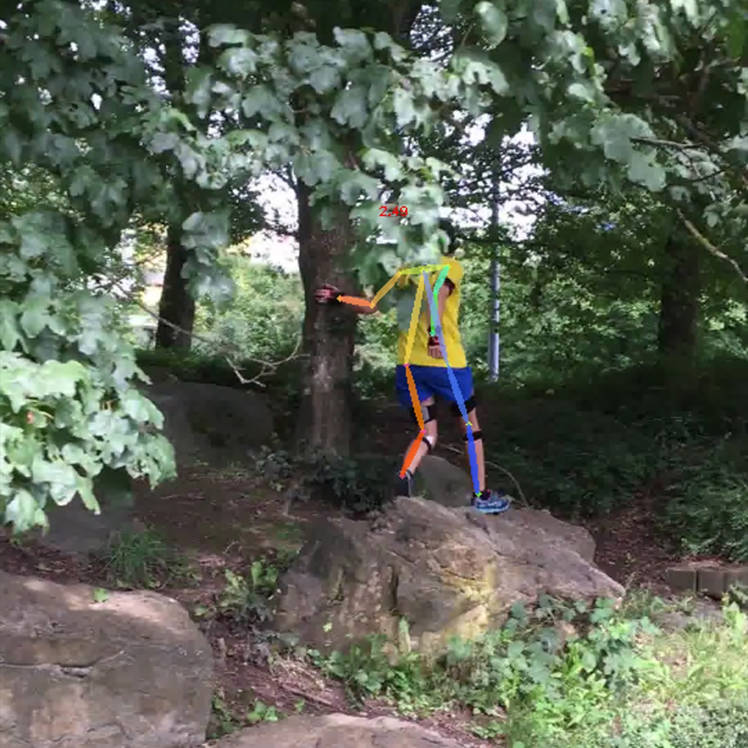}} &
{\includegraphics[width=0.18\textwidth]{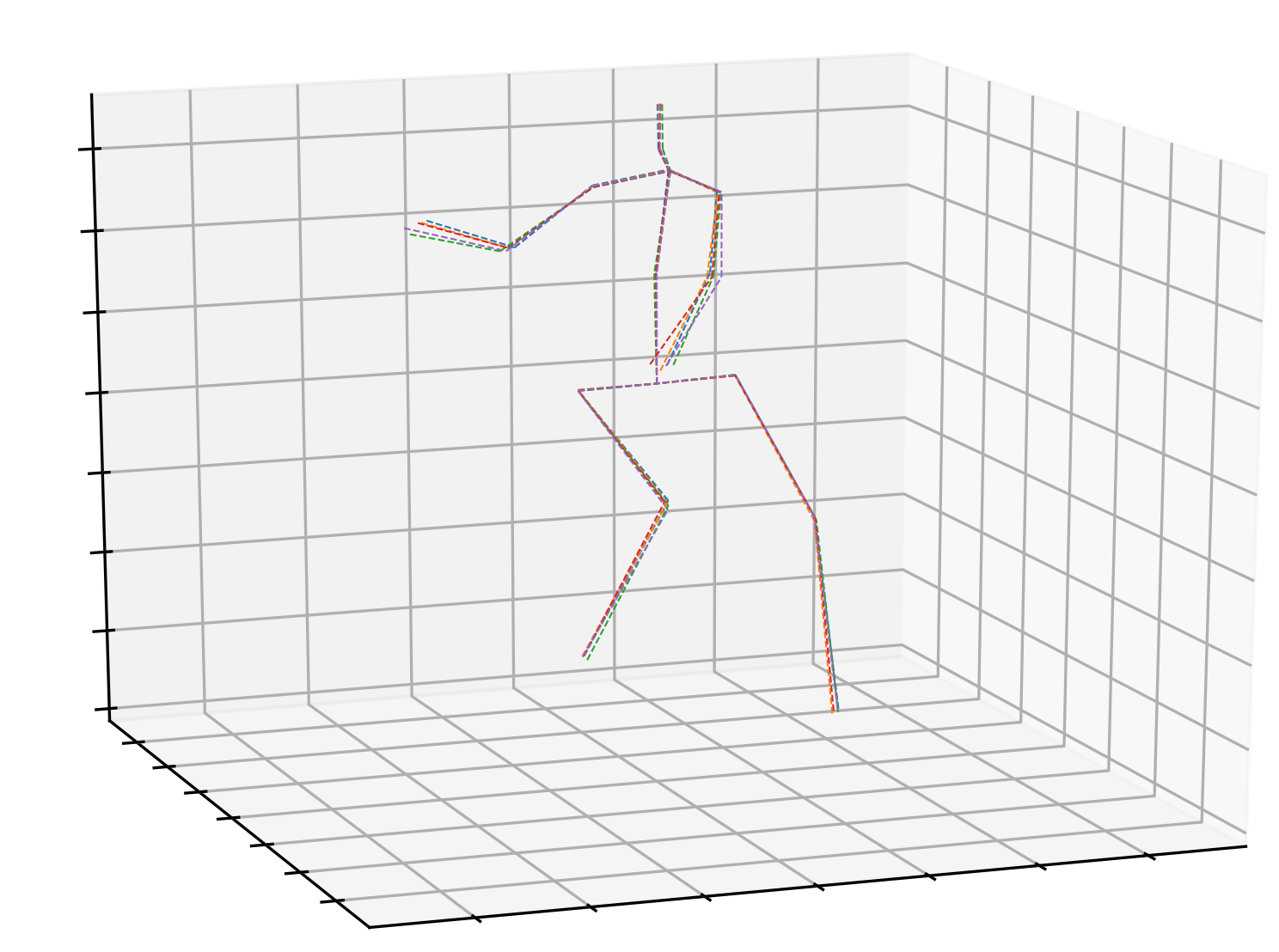}} \\

{\includegraphics[width=0.12\textwidth]{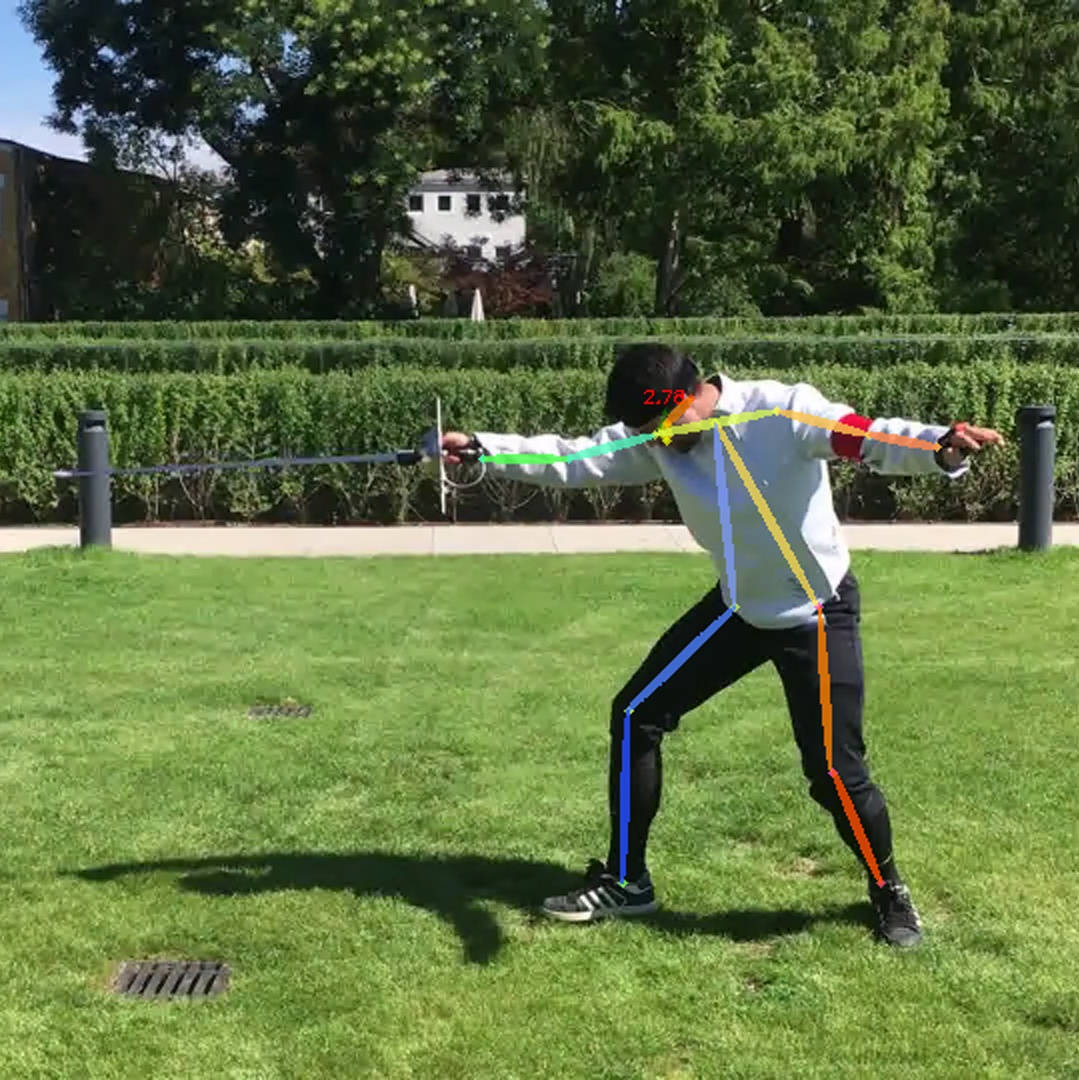}} &
{\includegraphics[width=0.18\textwidth]{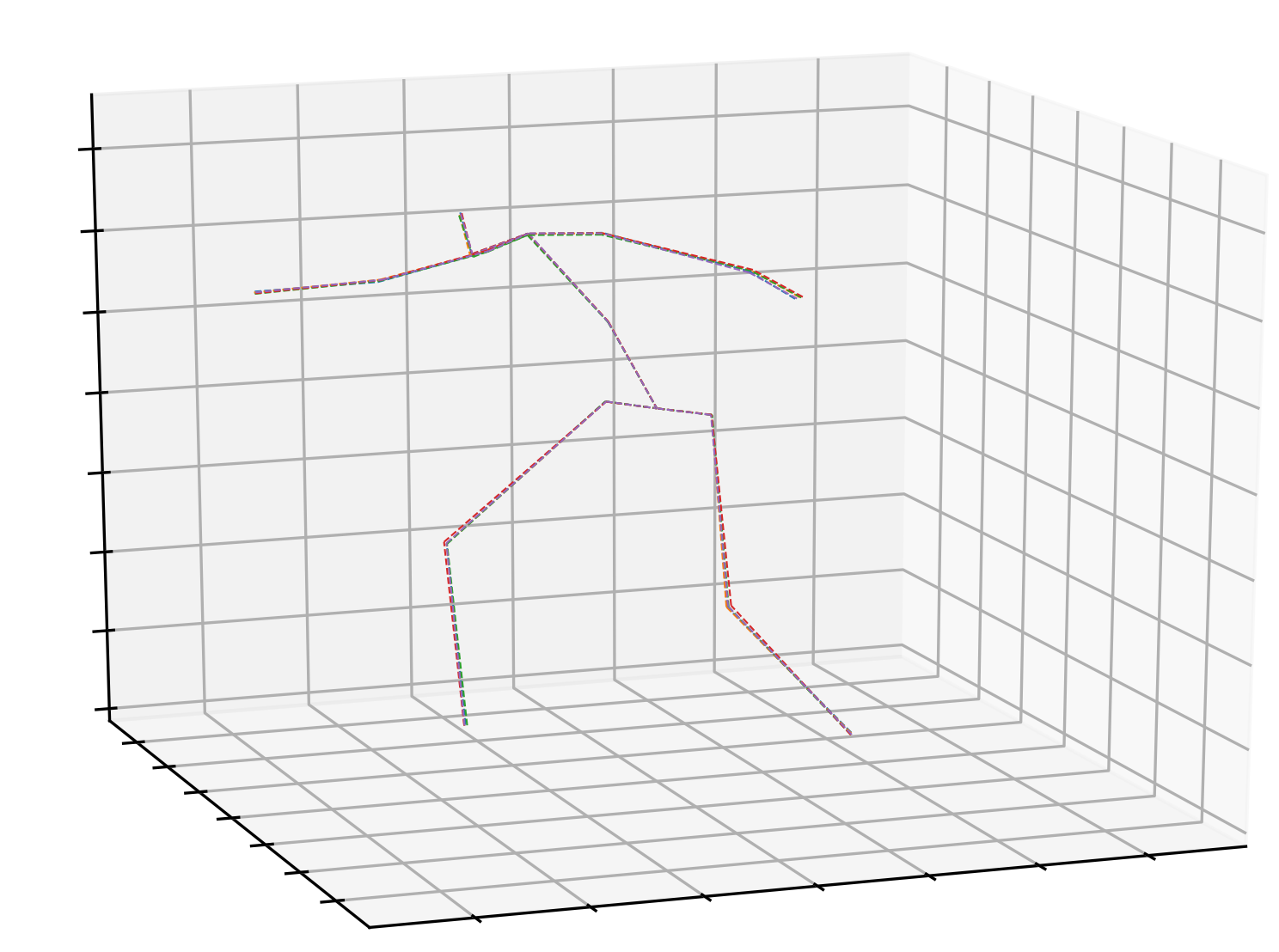}} &
{\includegraphics[width=0.12\textwidth]{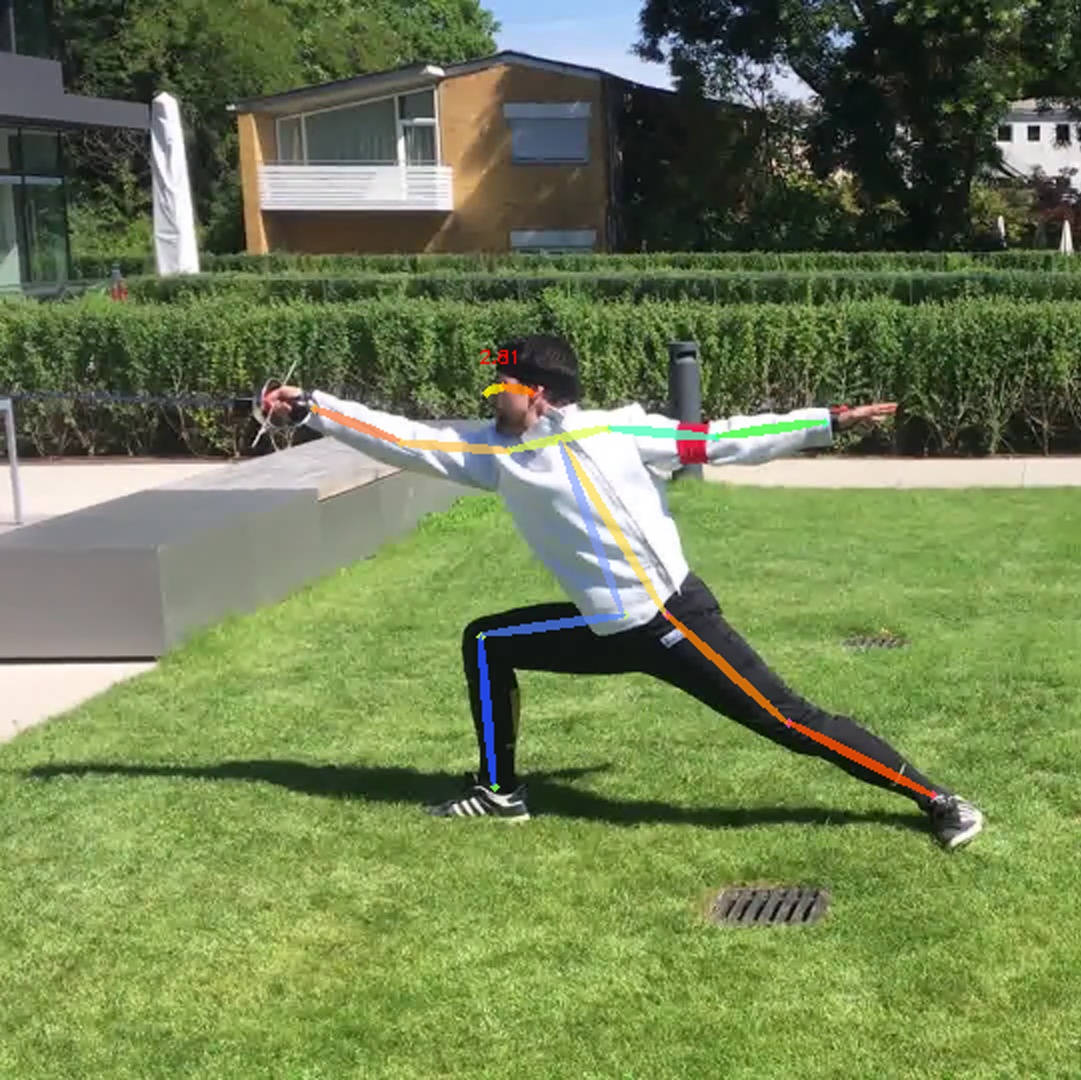}} &
{\includegraphics[width=0.18\textwidth]{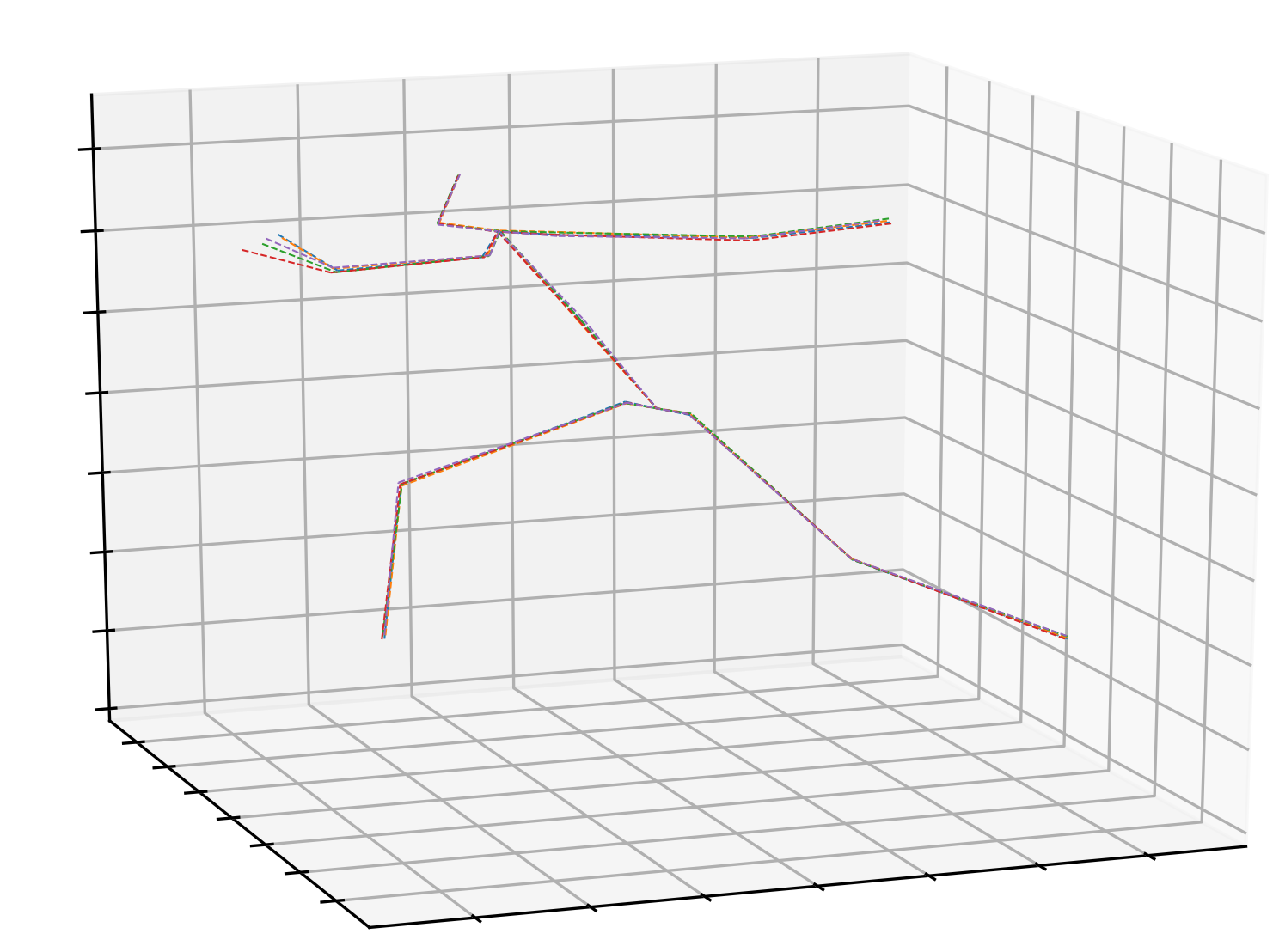}} &
{\includegraphics[width=0.12\textwidth]{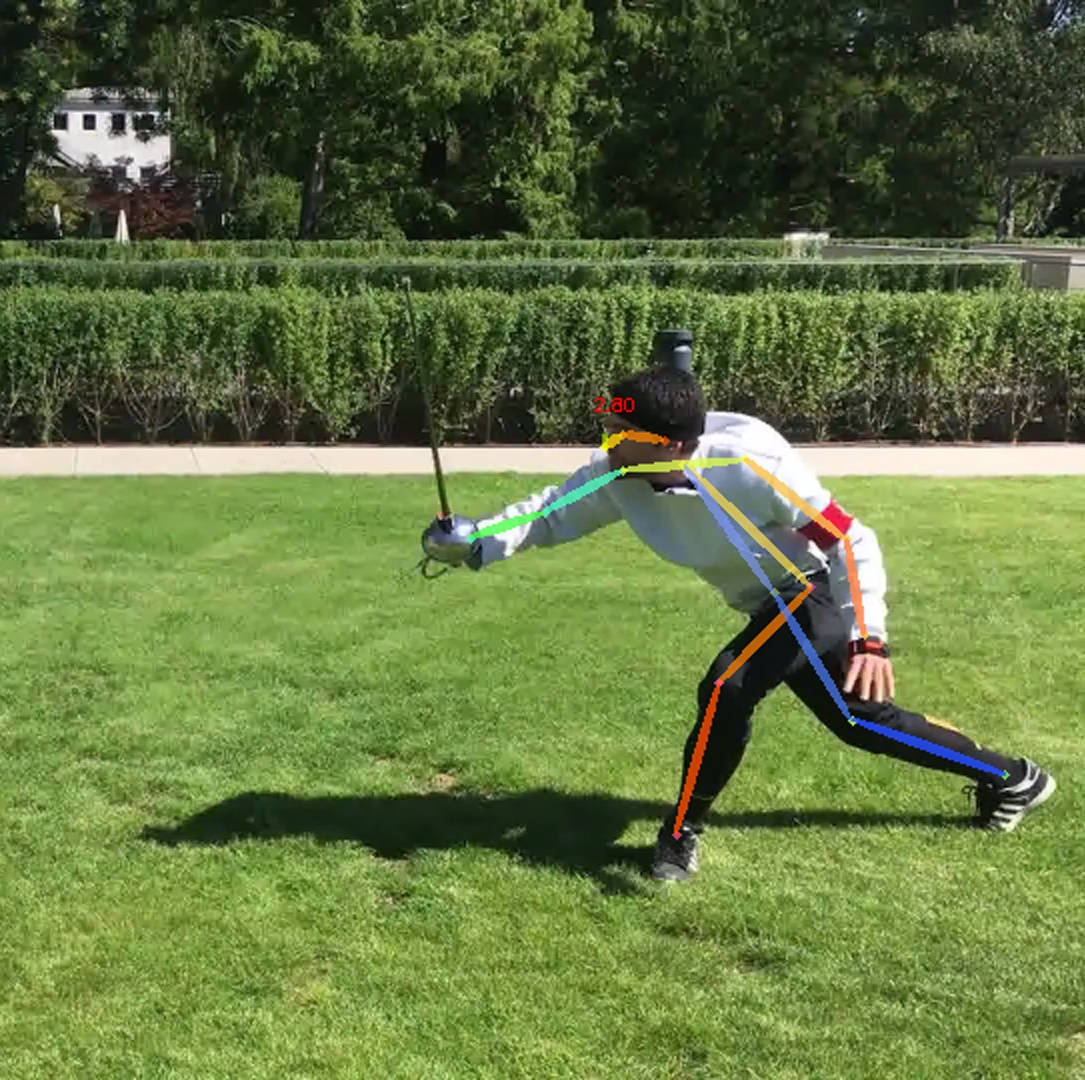}} &
{\includegraphics[width=0.18\textwidth]{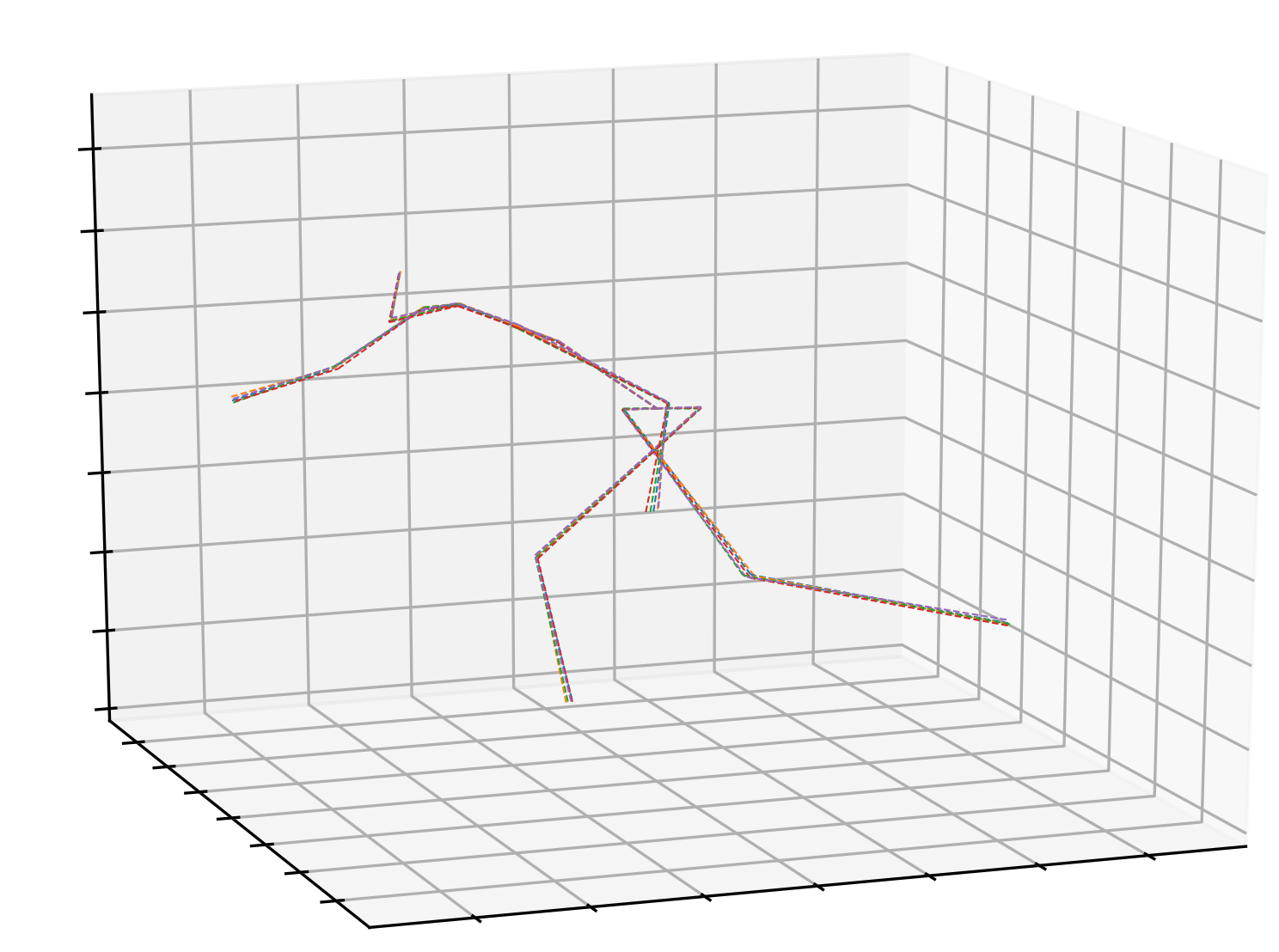}} \\

{\includegraphics[width=0.12\textwidth]{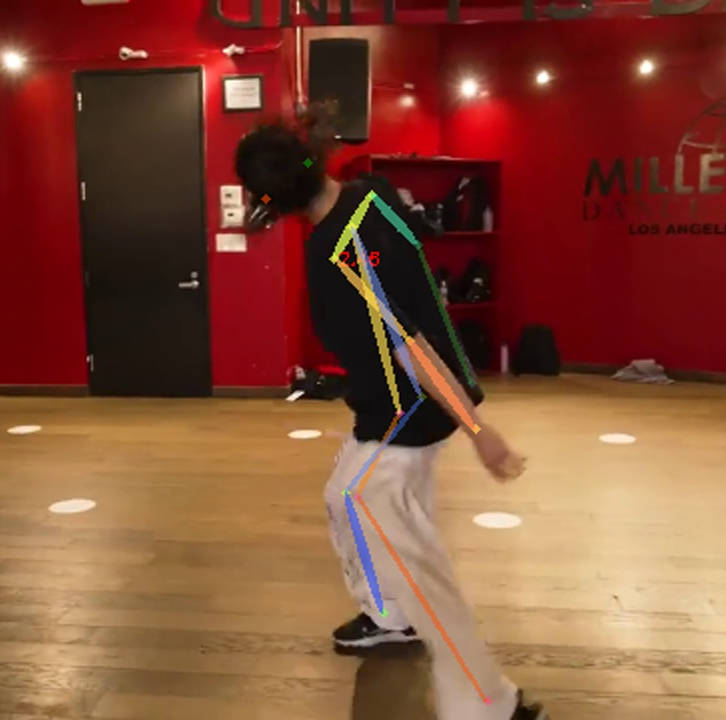}} &
{\includegraphics[width=0.18\textwidth]{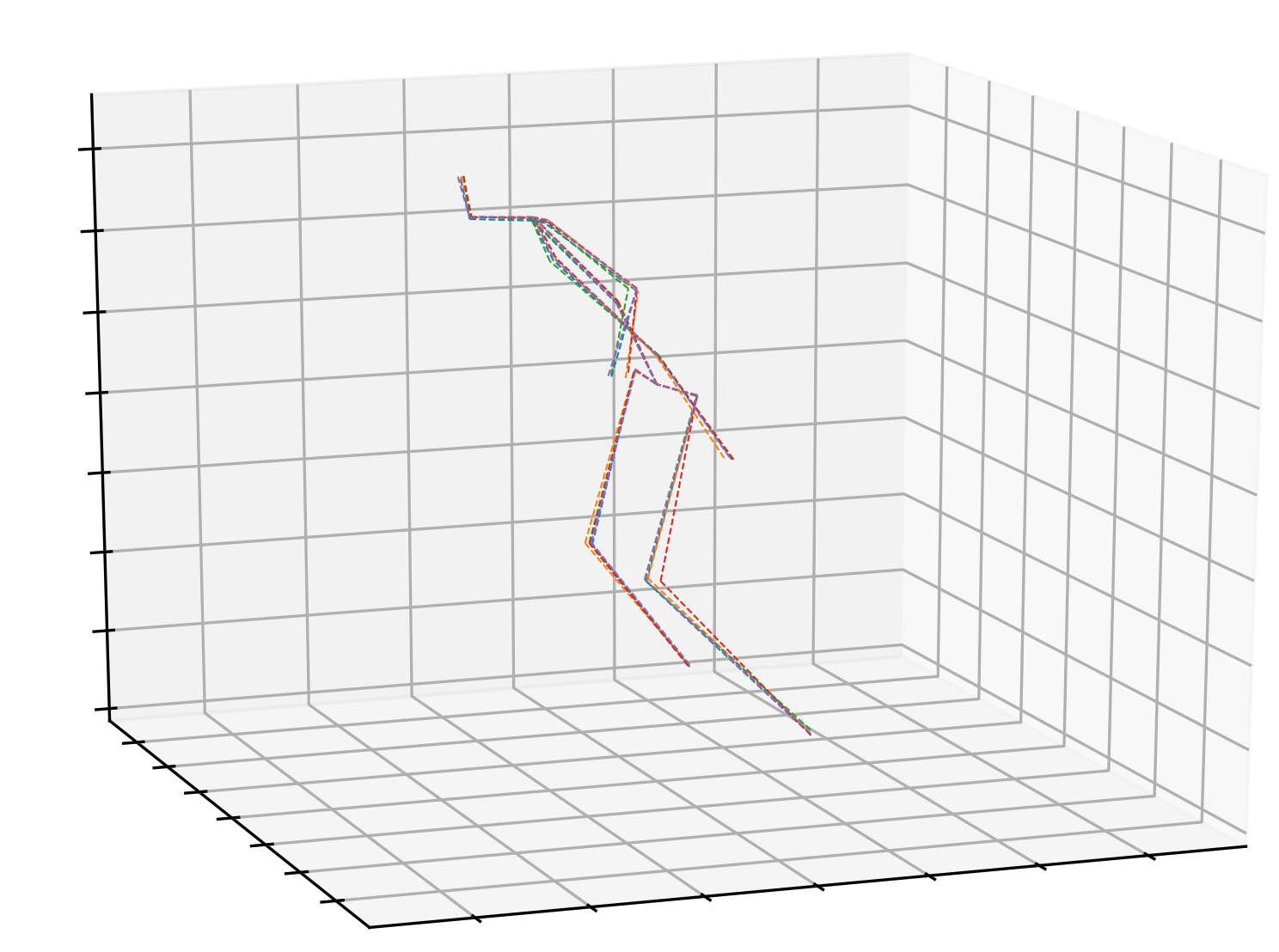}} &
{\includegraphics[width=0.12\textwidth]{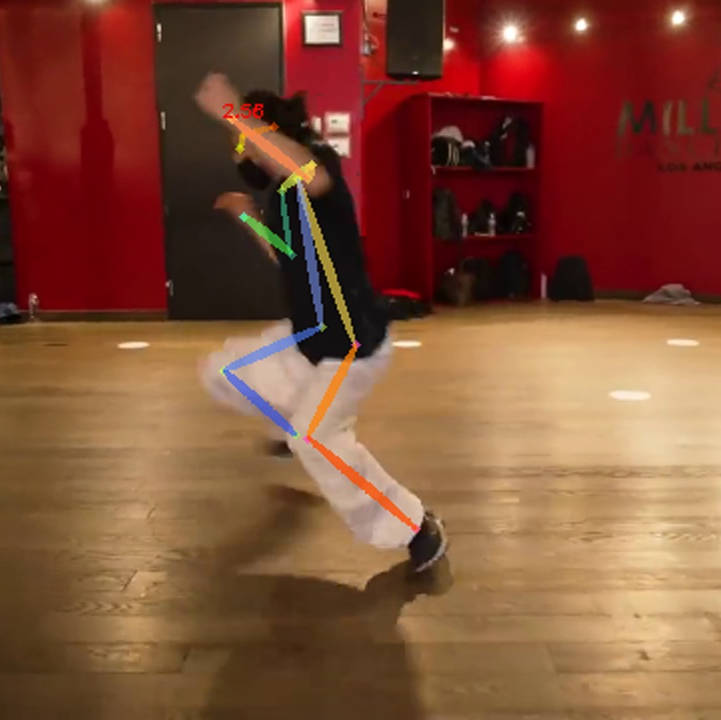}} &
{\includegraphics[width=0.18\textwidth]{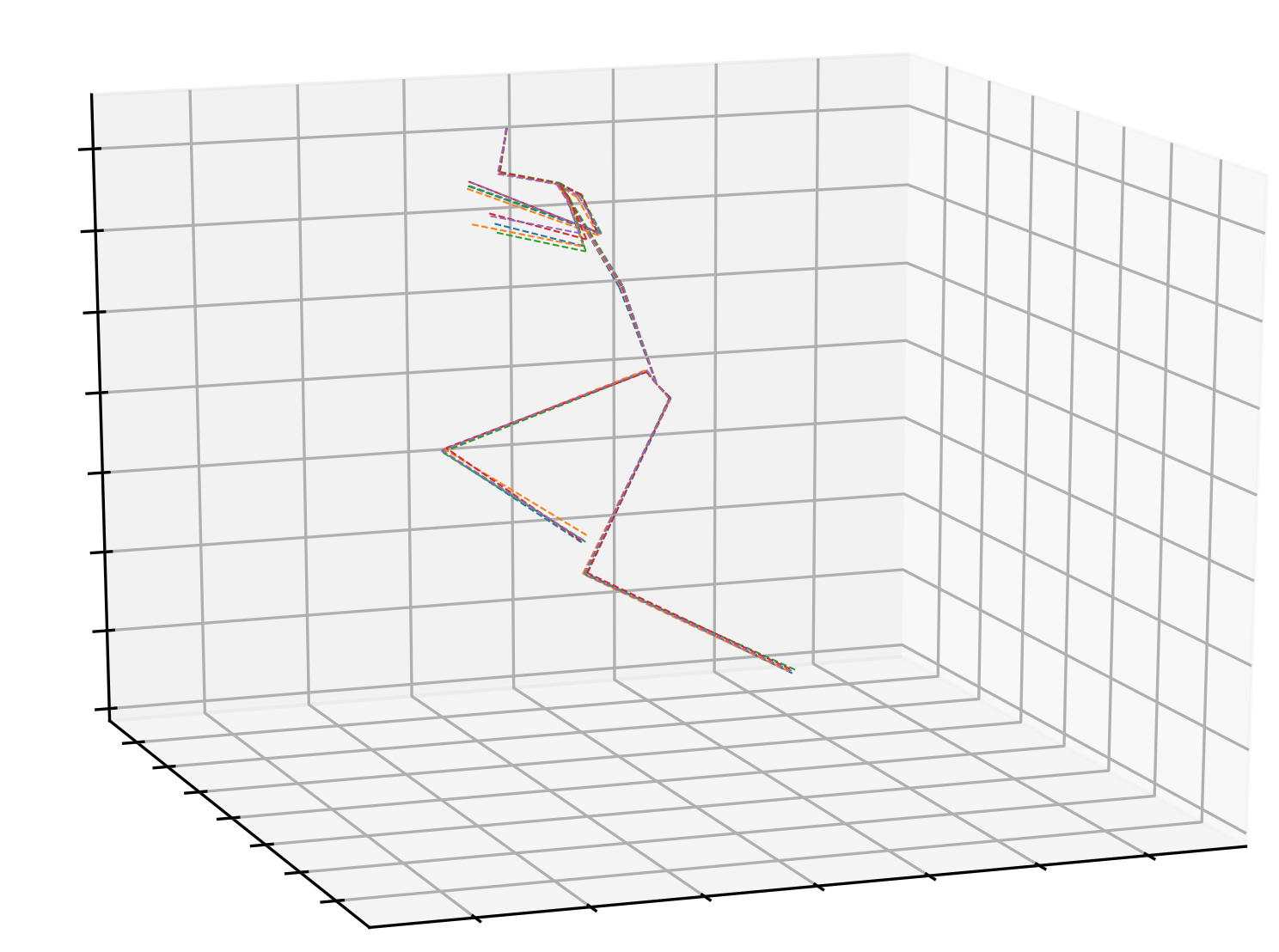}} &
{\includegraphics[width=0.12\textwidth]{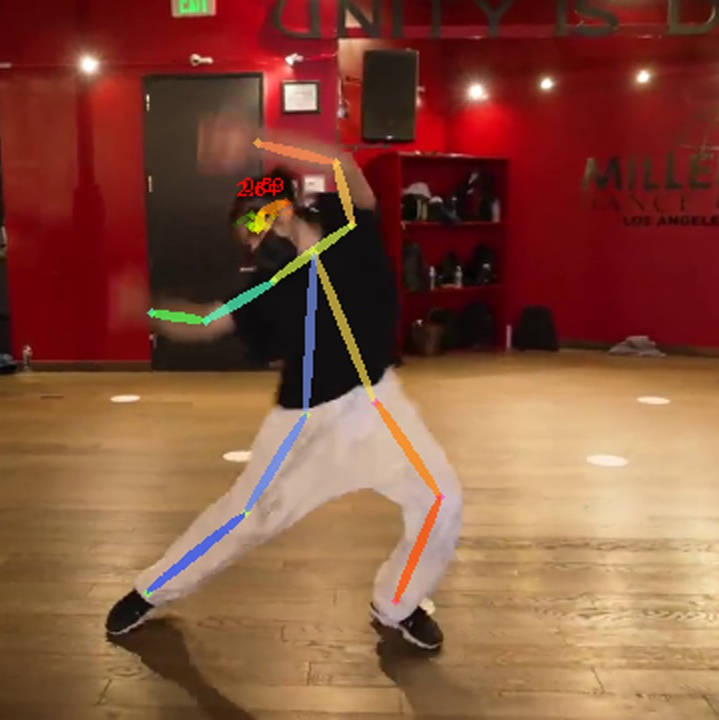}} &
{\includegraphics[width=0.18\textwidth]{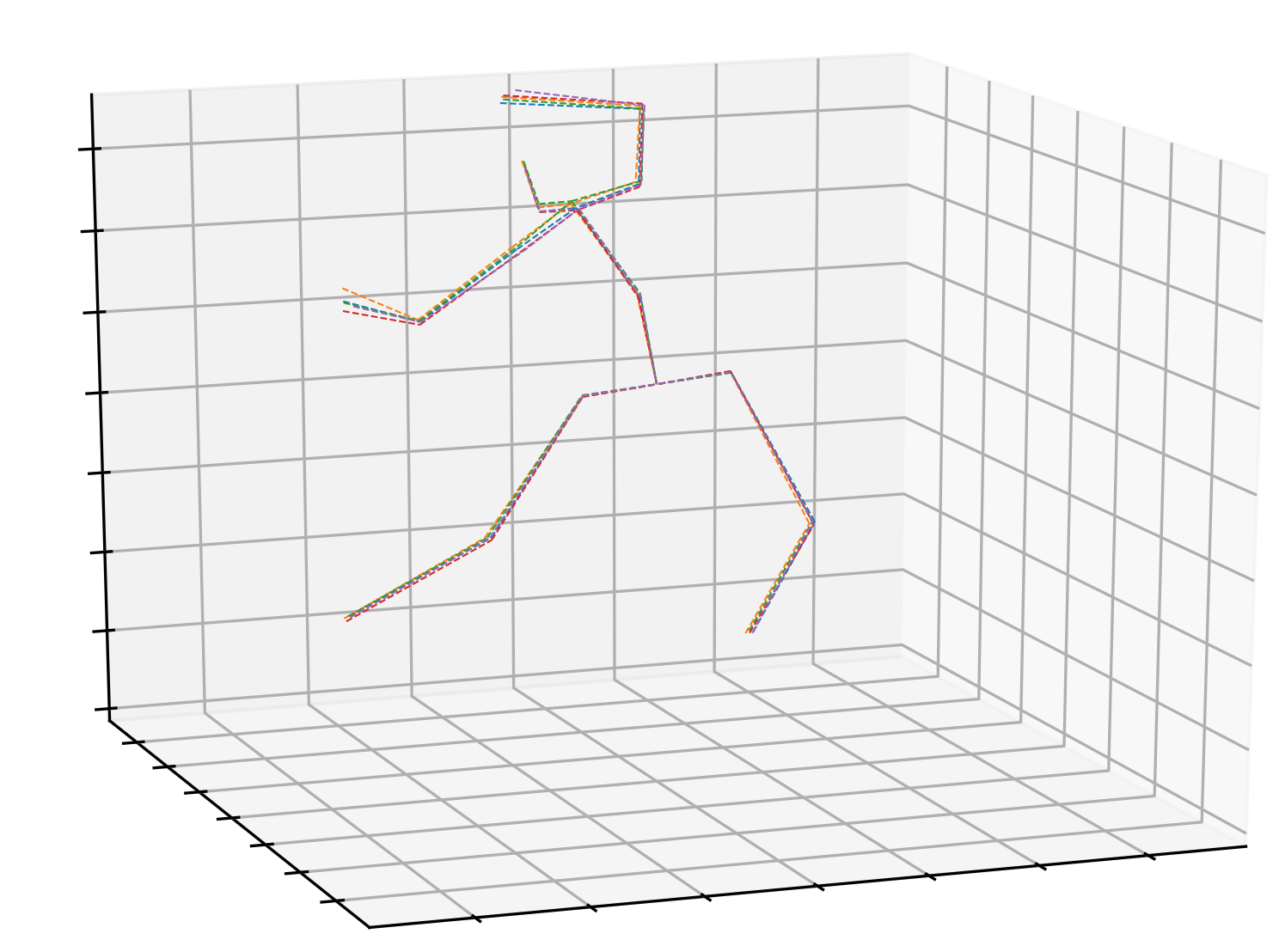}} \\

{\includegraphics[width=0.12\textwidth]{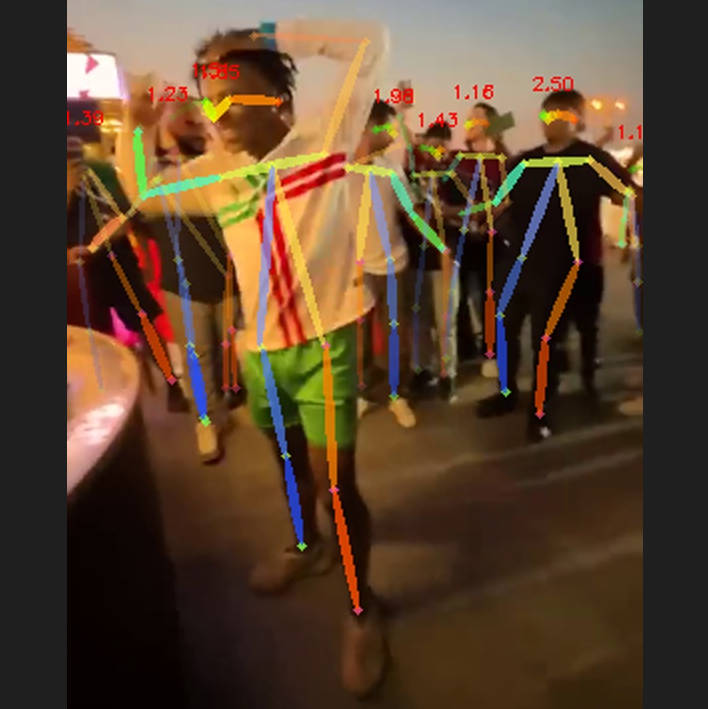}} &
{\includegraphics[width=0.18\textwidth]{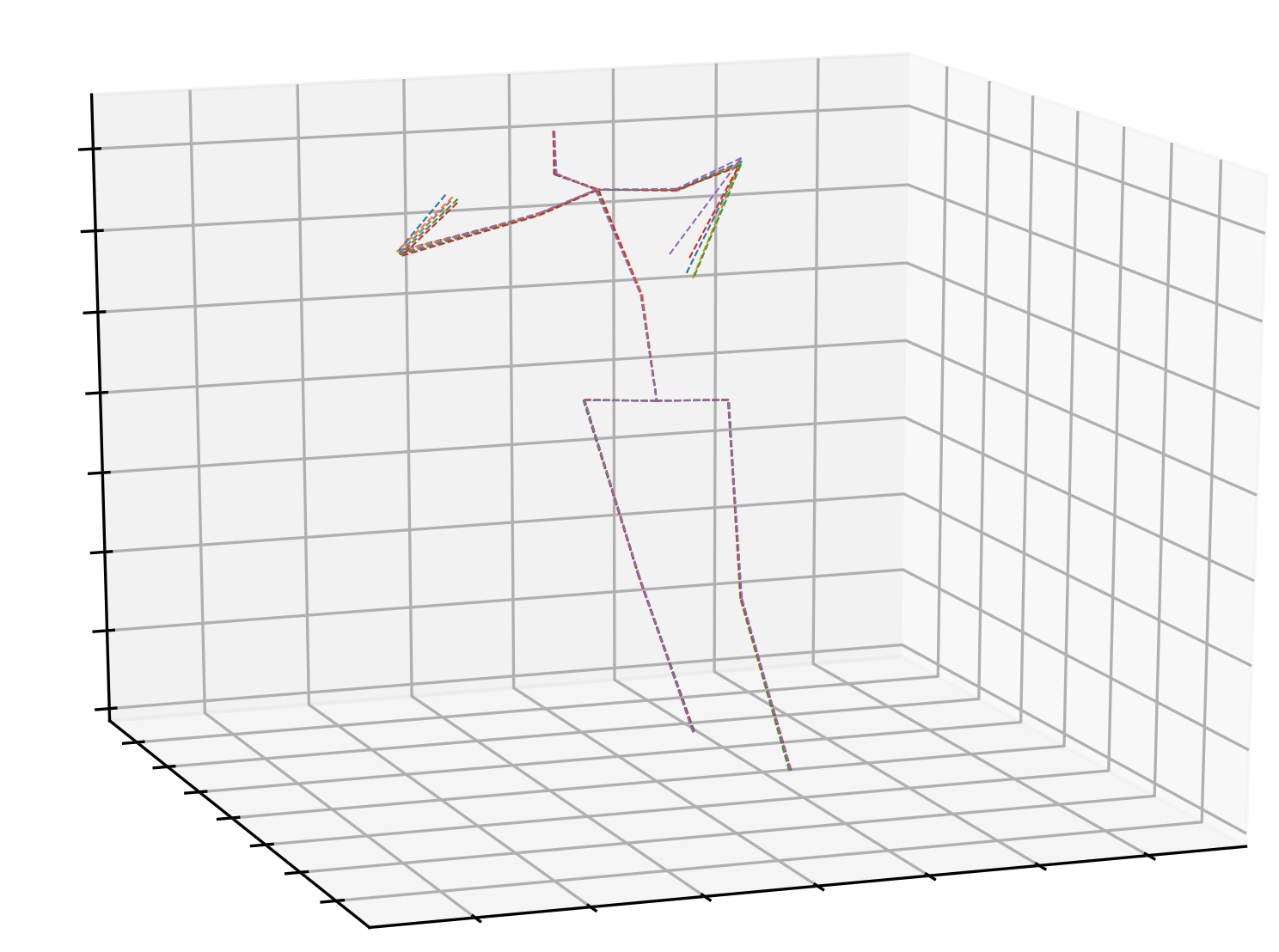}} &
{\includegraphics[width=0.12\textwidth]{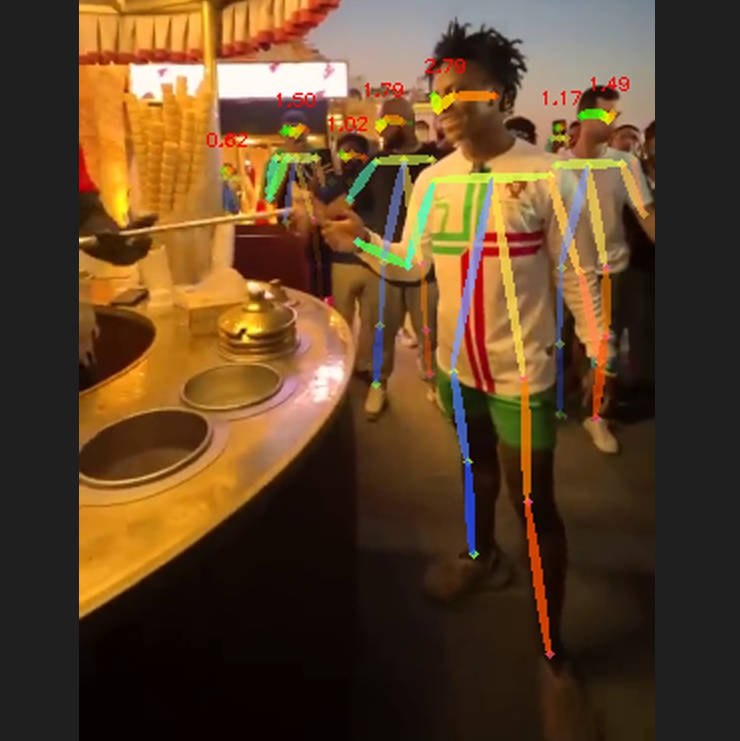}} &
{\includegraphics[width=0.18\textwidth]{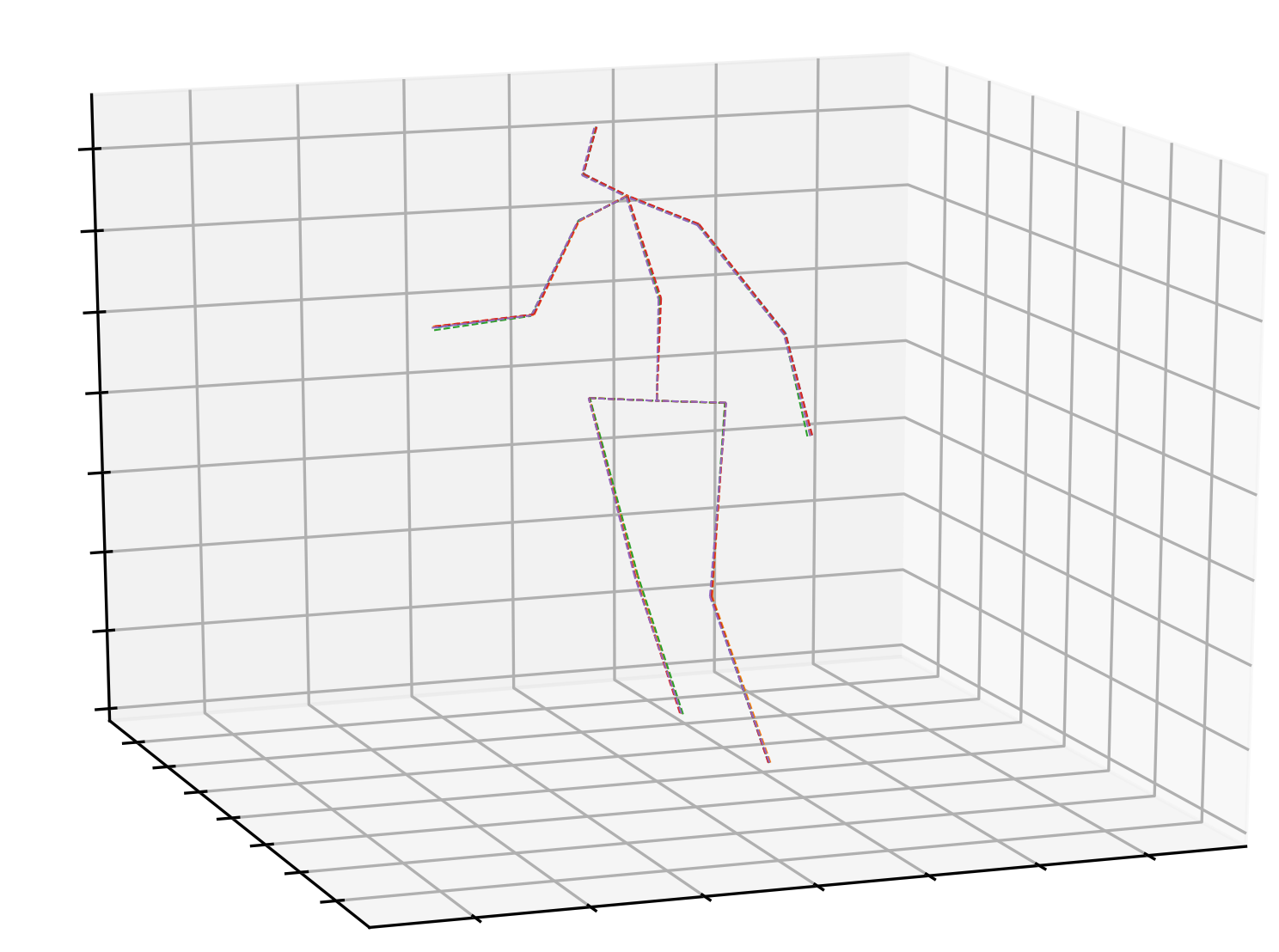}} &
{\includegraphics[width=0.12\textwidth]{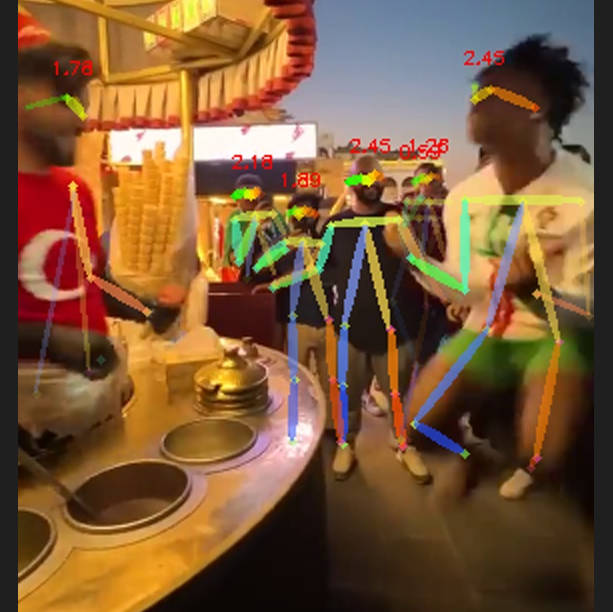}} &
{\includegraphics[width=0.18\textwidth]{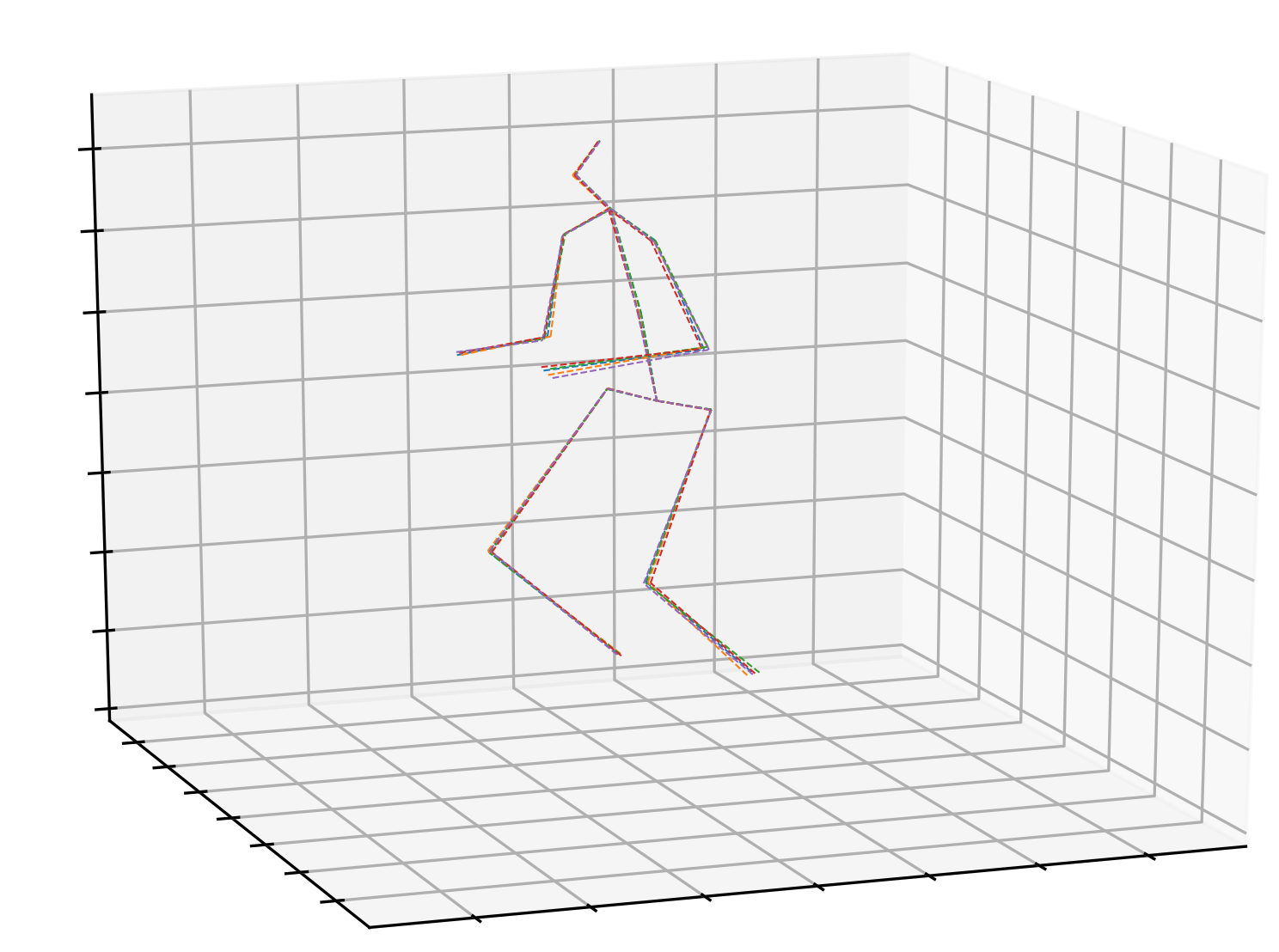}} \\

{\includegraphics[width=0.12\textwidth]{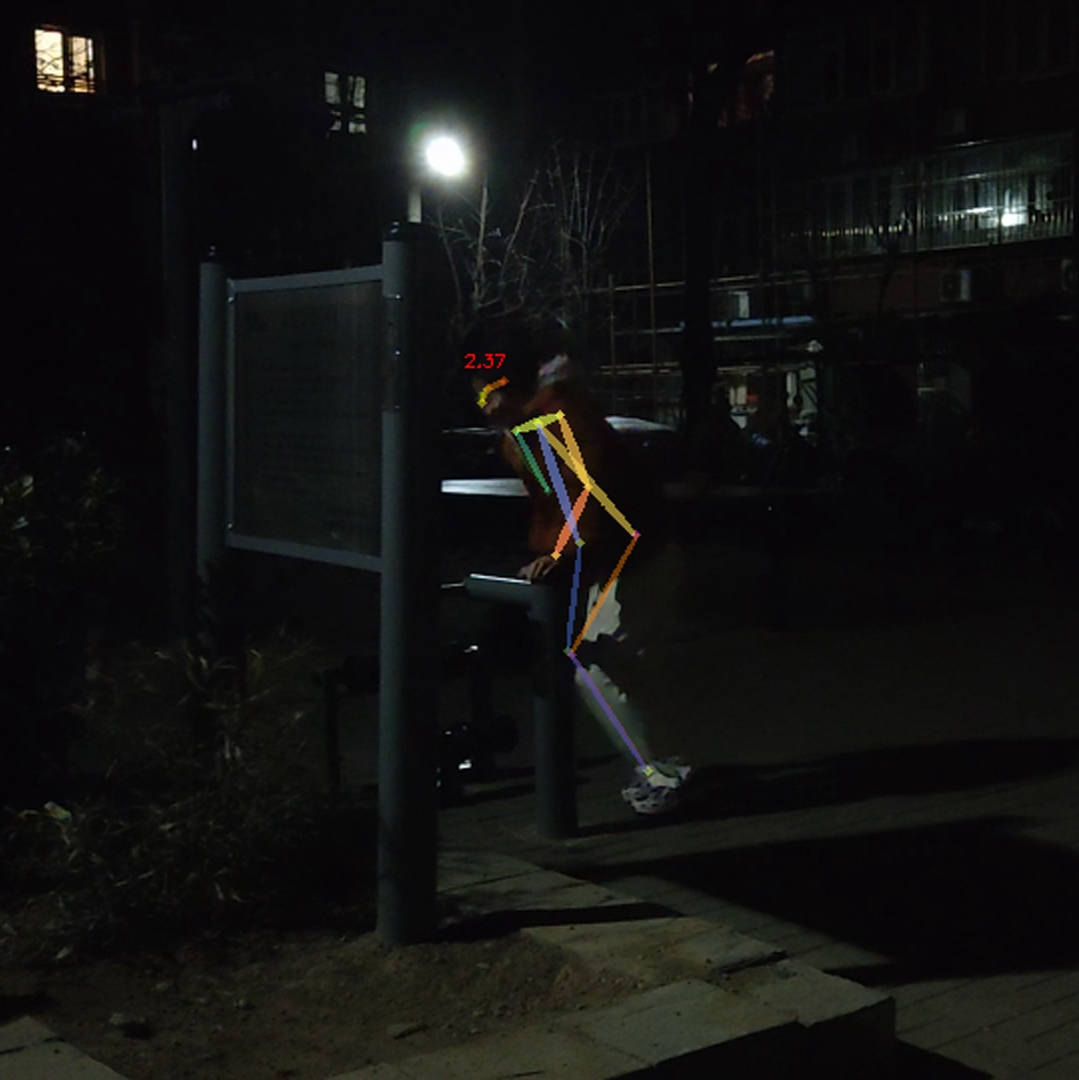}} &
{\includegraphics[width=0.18\textwidth]{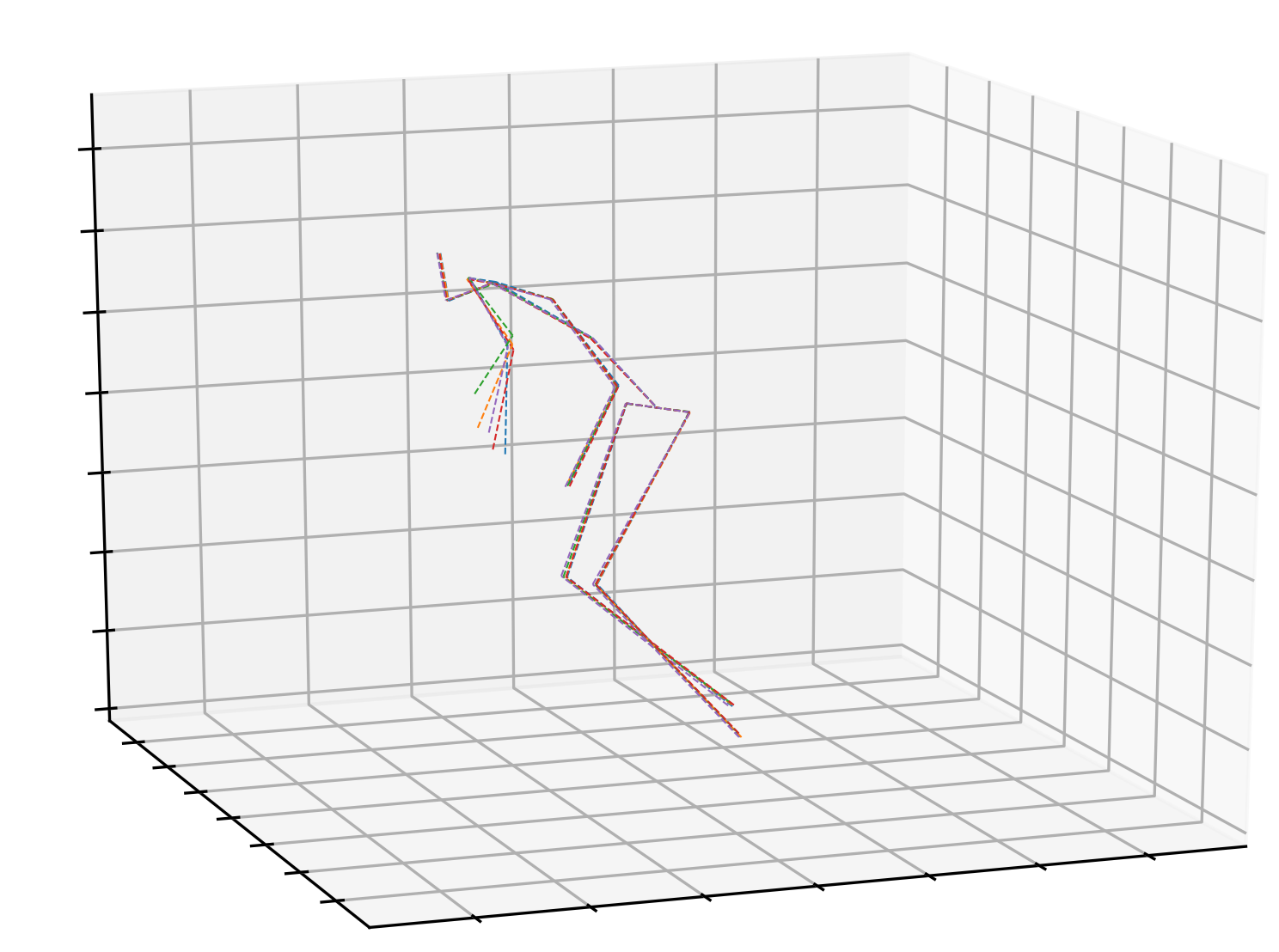}} &
{\includegraphics[width=0.12\textwidth]{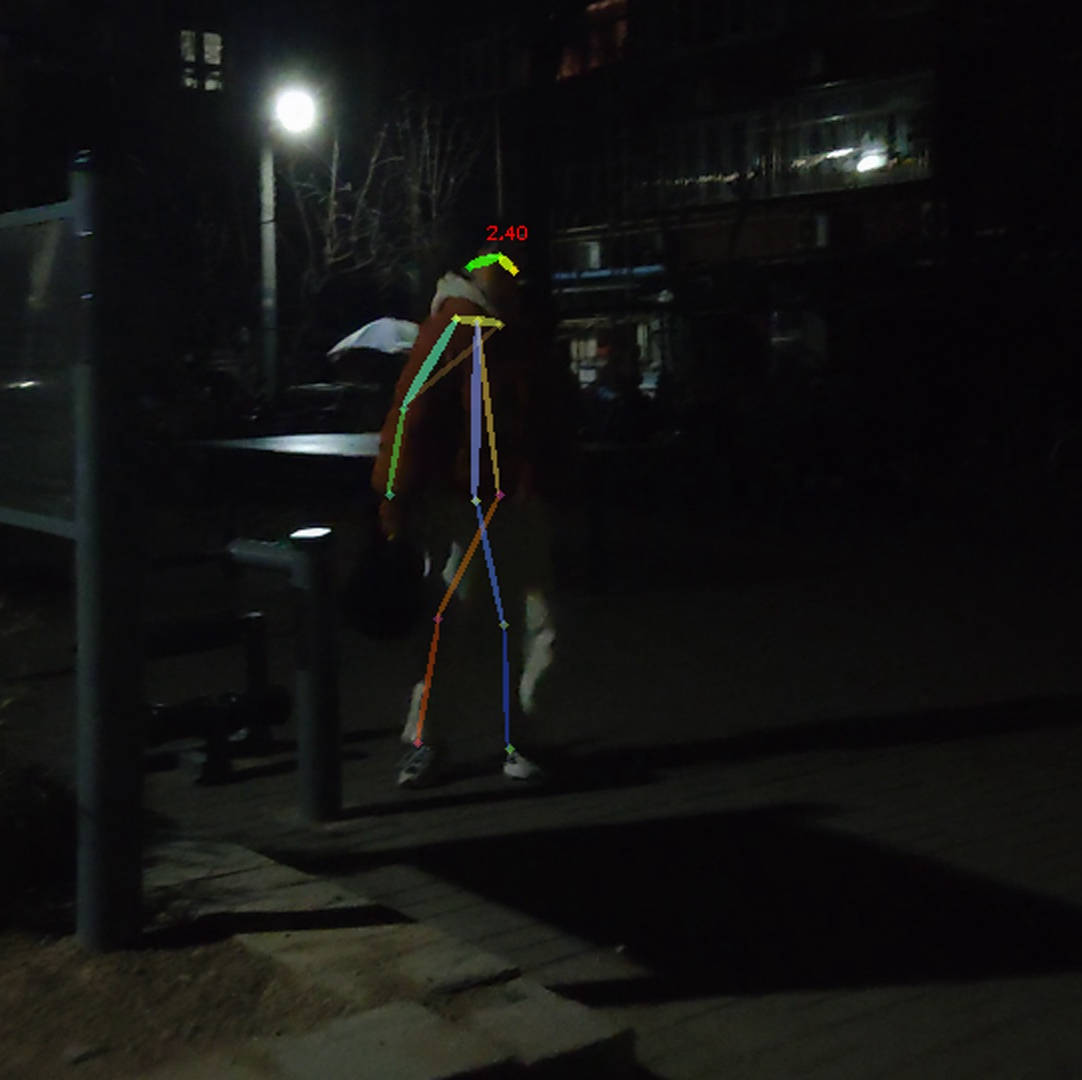}} &
{\includegraphics[width=0.18\textwidth]{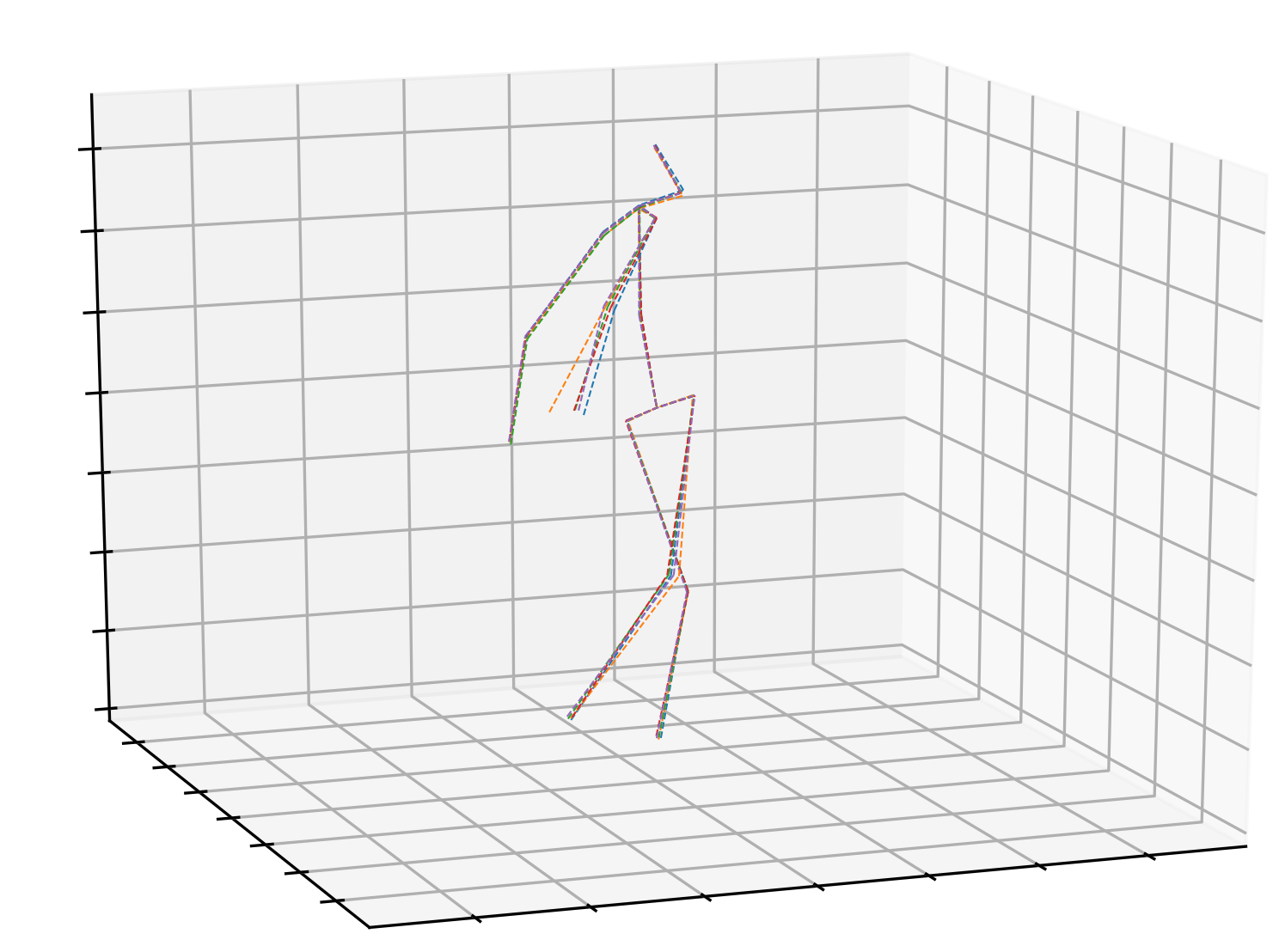}} &
{\includegraphics[width=0.12\textwidth]{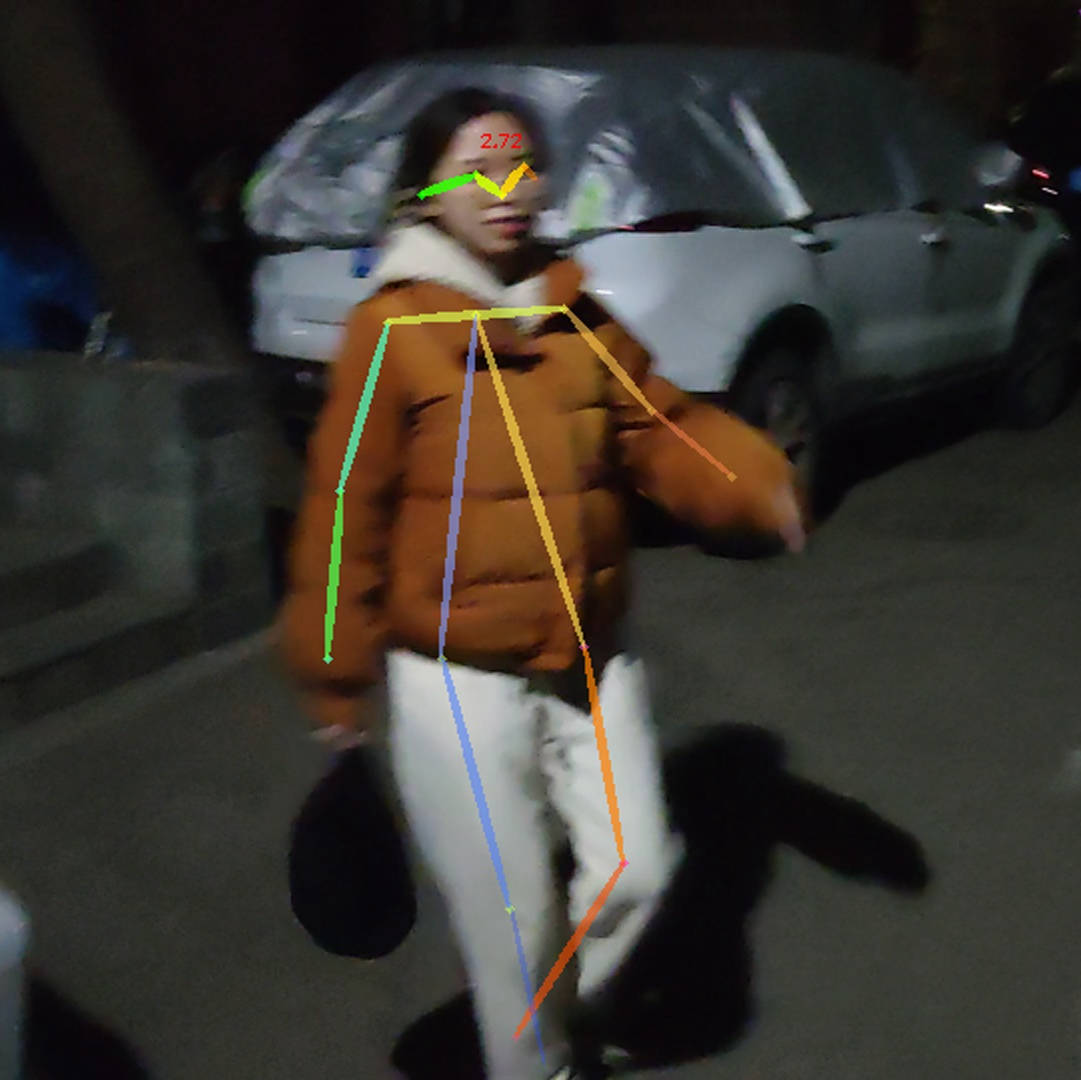}} &
{\includegraphics[width=0.18\textwidth]{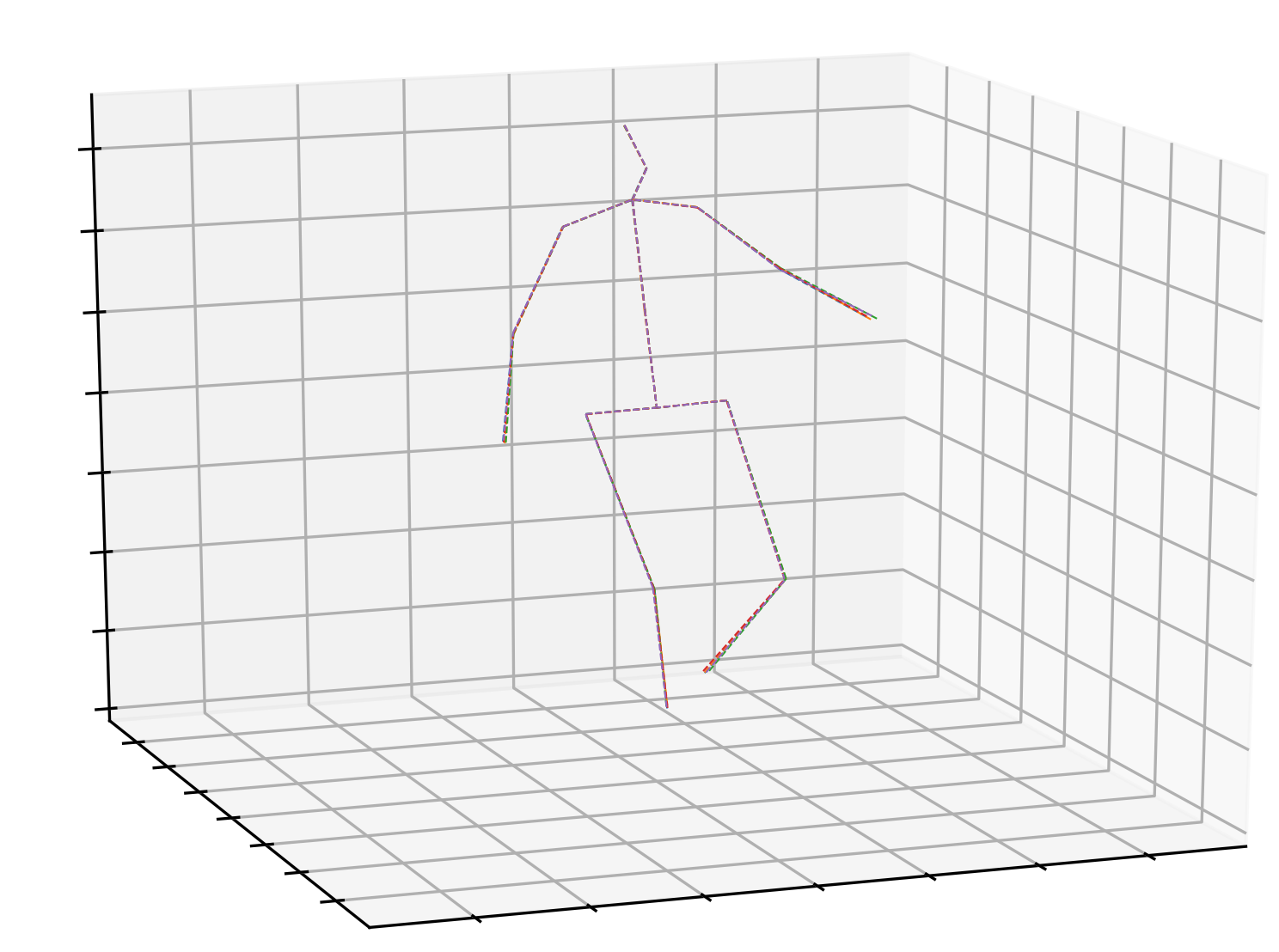}} \\

{\includegraphics[width=0.12\textwidth]{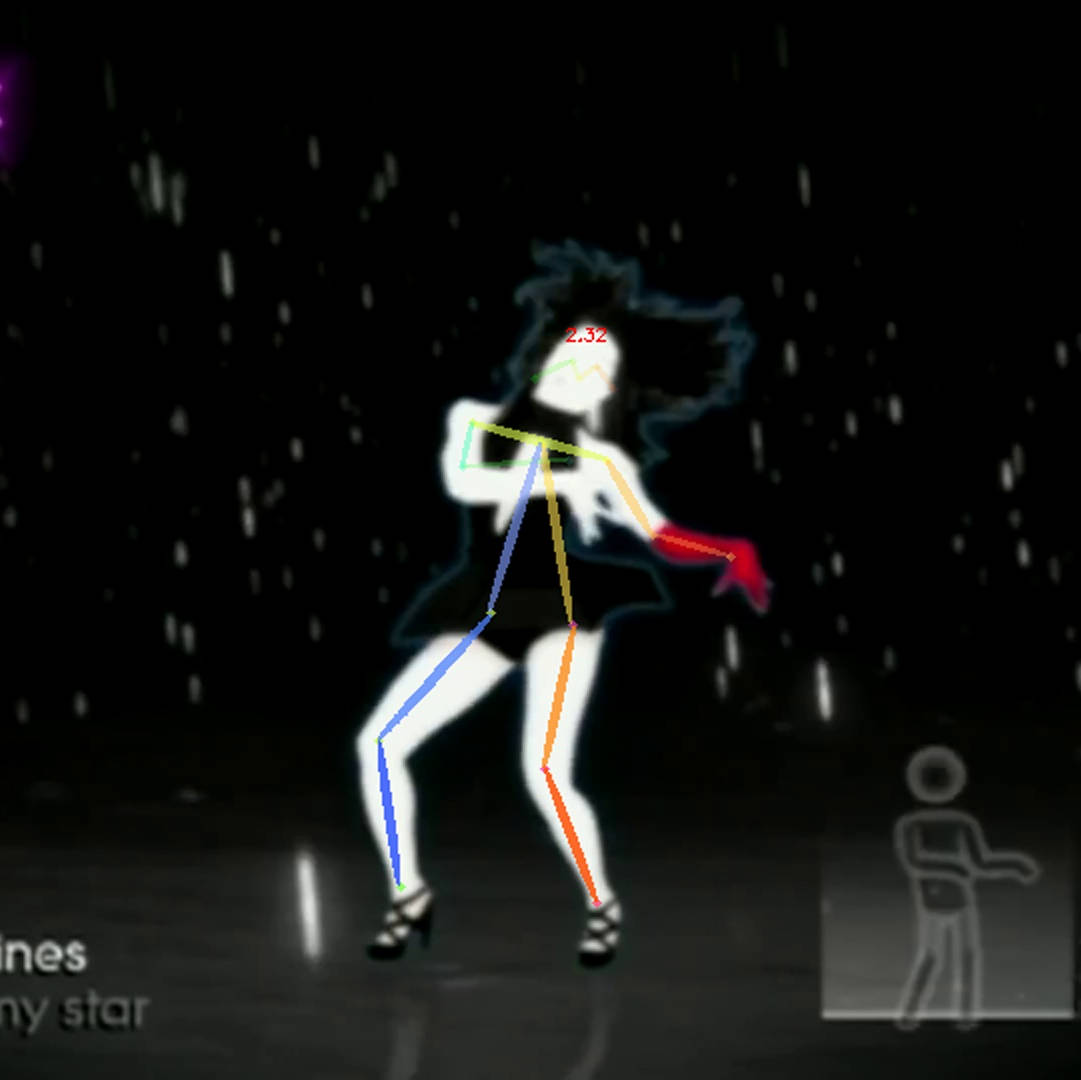}} &
{\includegraphics[width=0.18\textwidth]{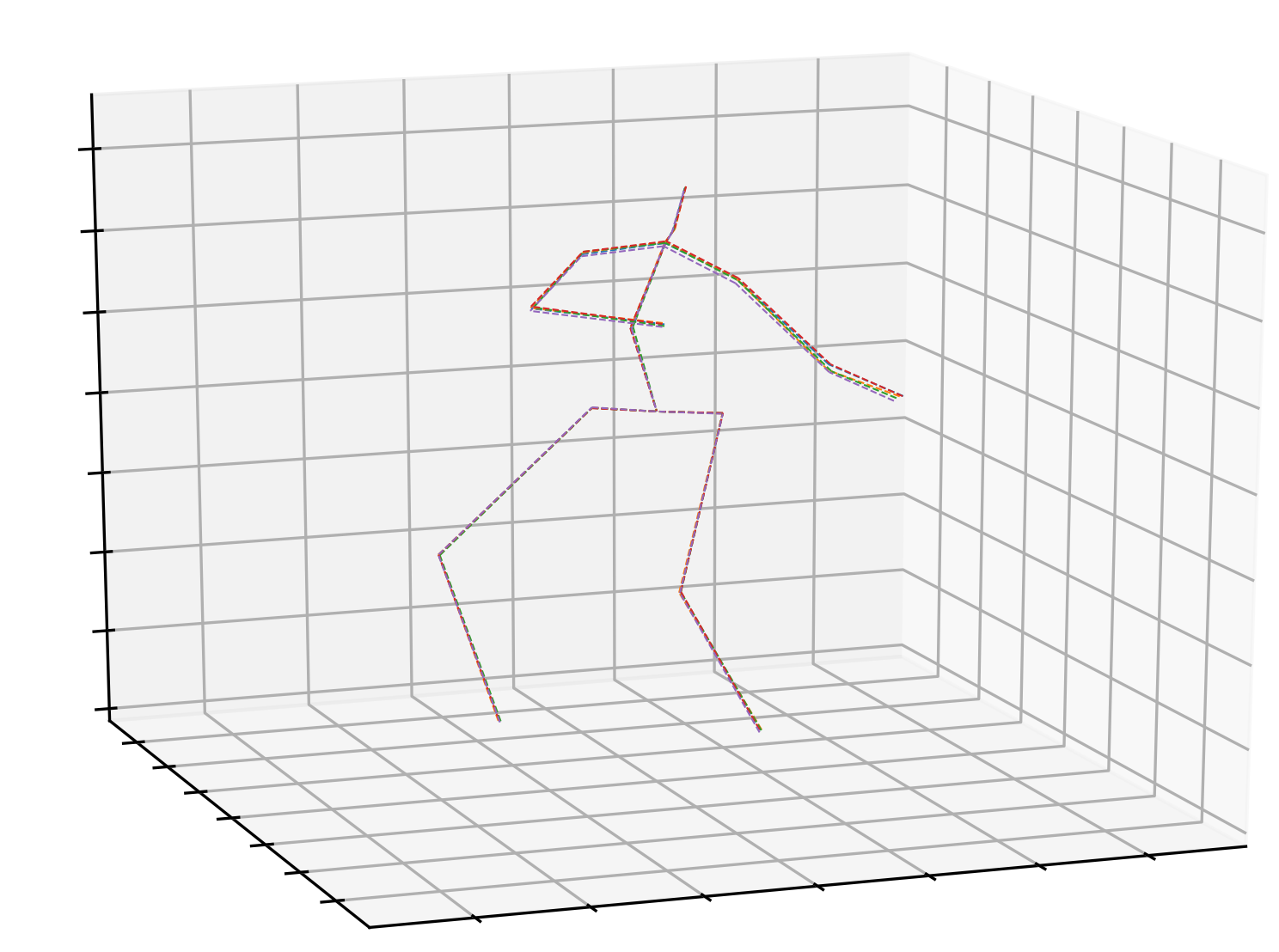}} &
{\includegraphics[width=0.12\textwidth]{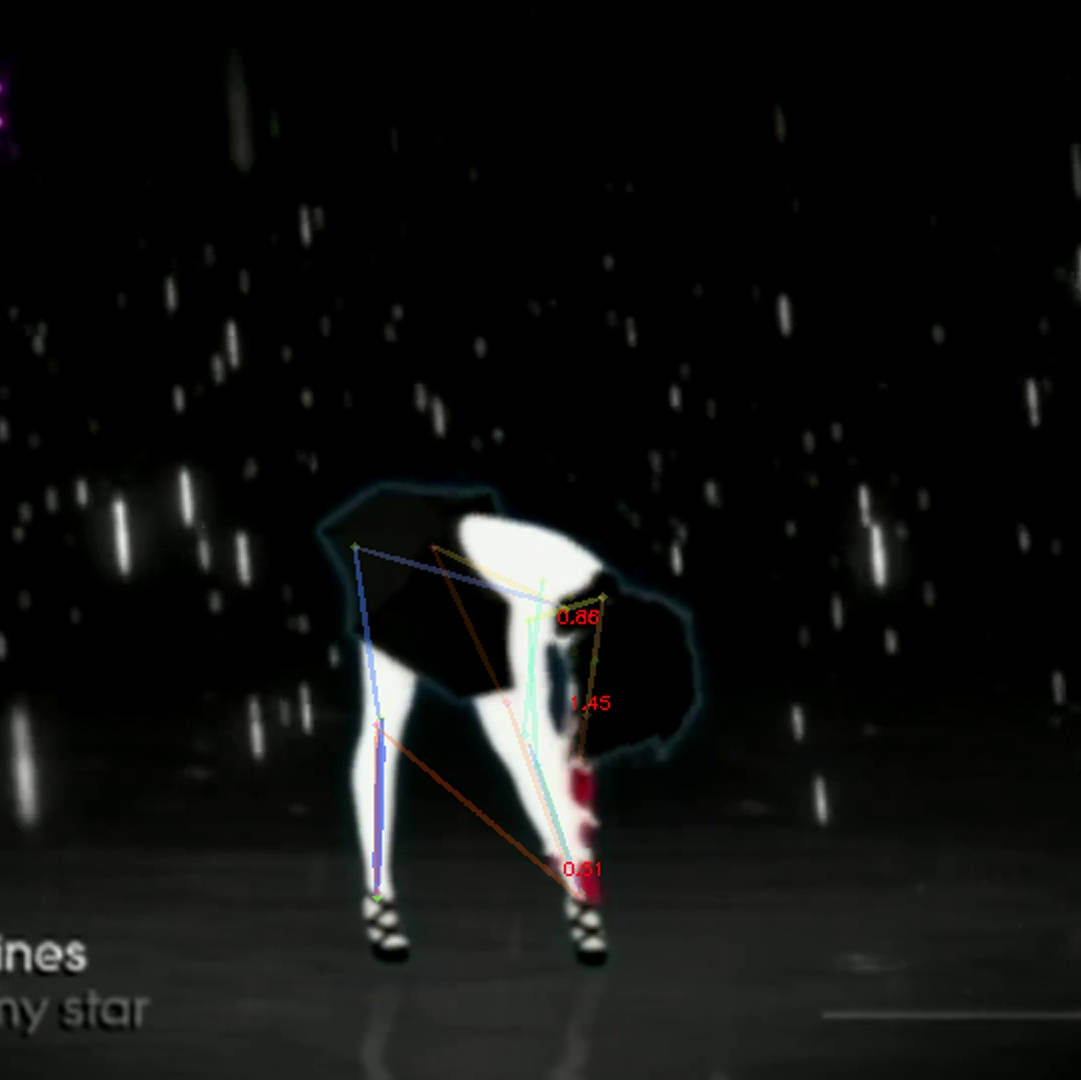}} &
{\includegraphics[width=0.18\textwidth]{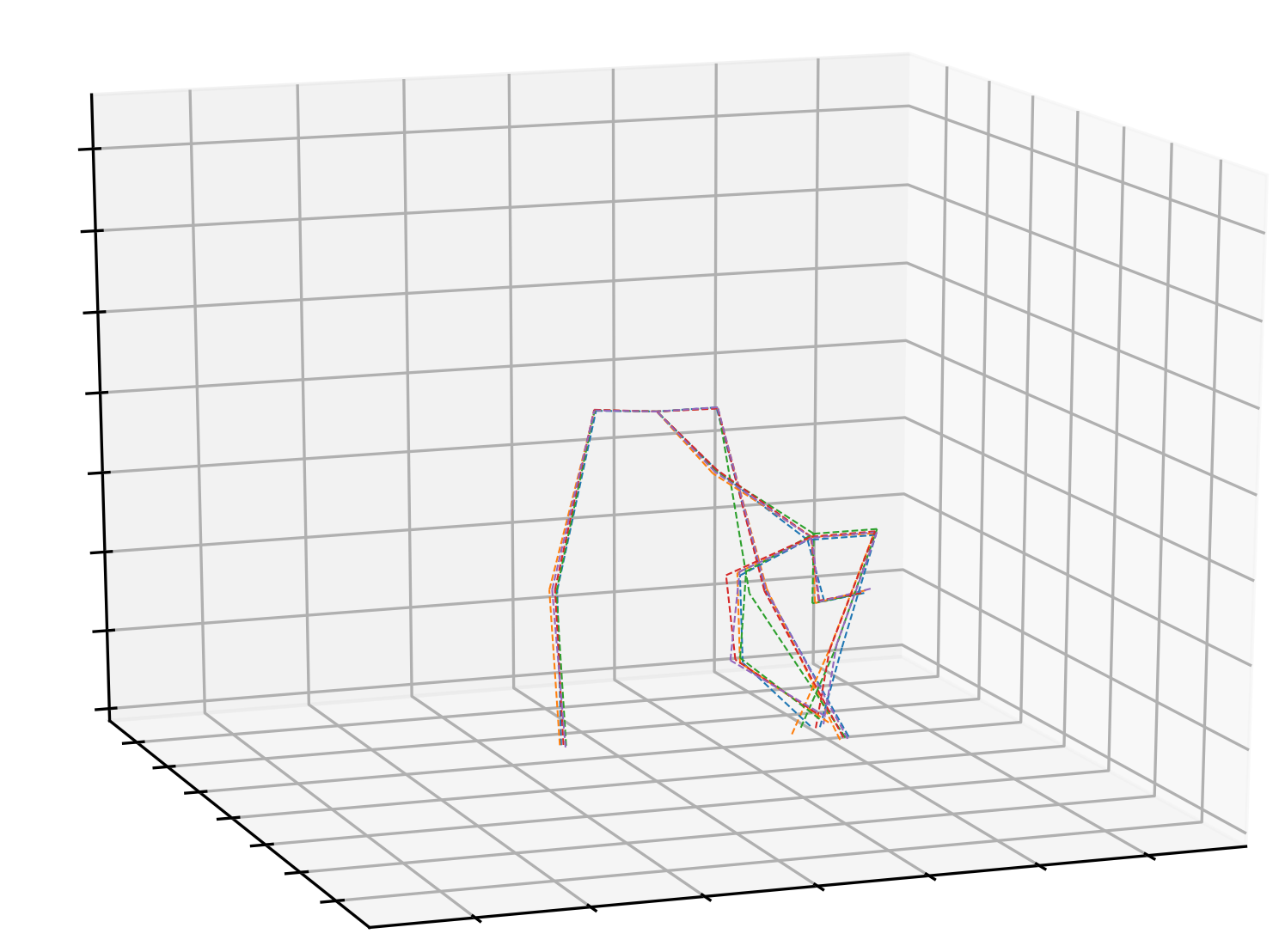}} &
{\includegraphics[width=0.12\textwidth]{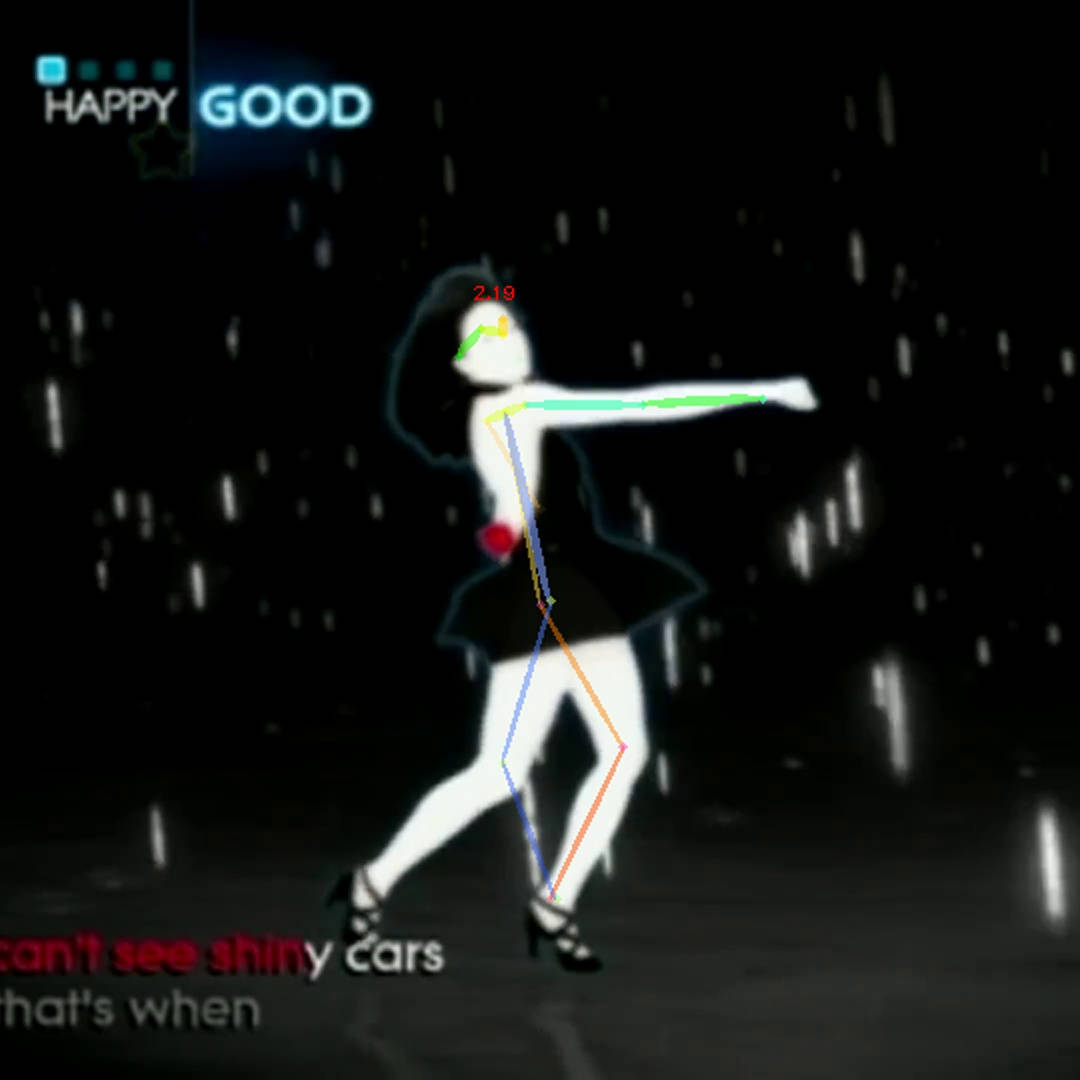}} &
{\includegraphics[width=0.18\textwidth]{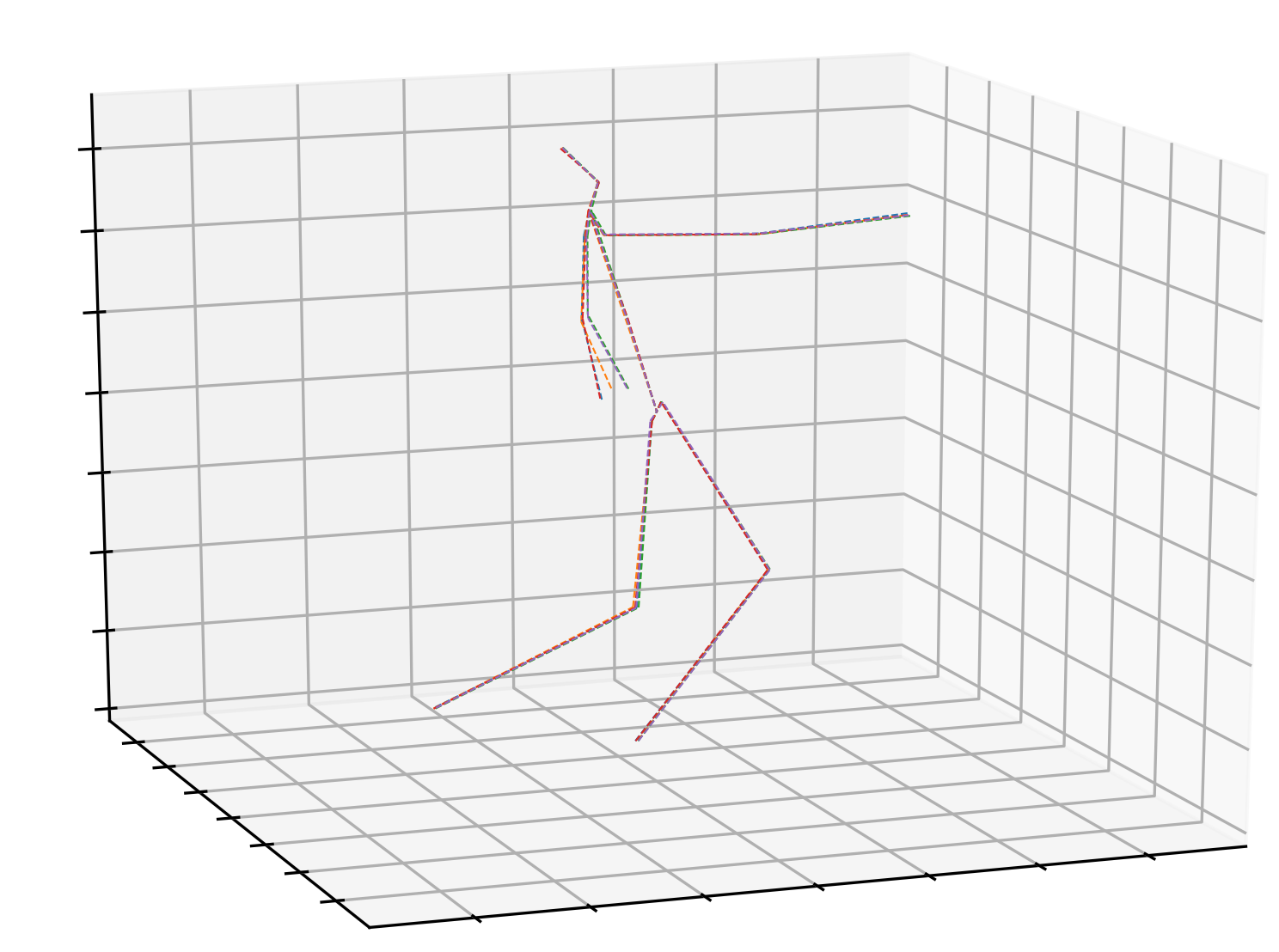}} \\

{\includegraphics[width=0.12\textwidth]{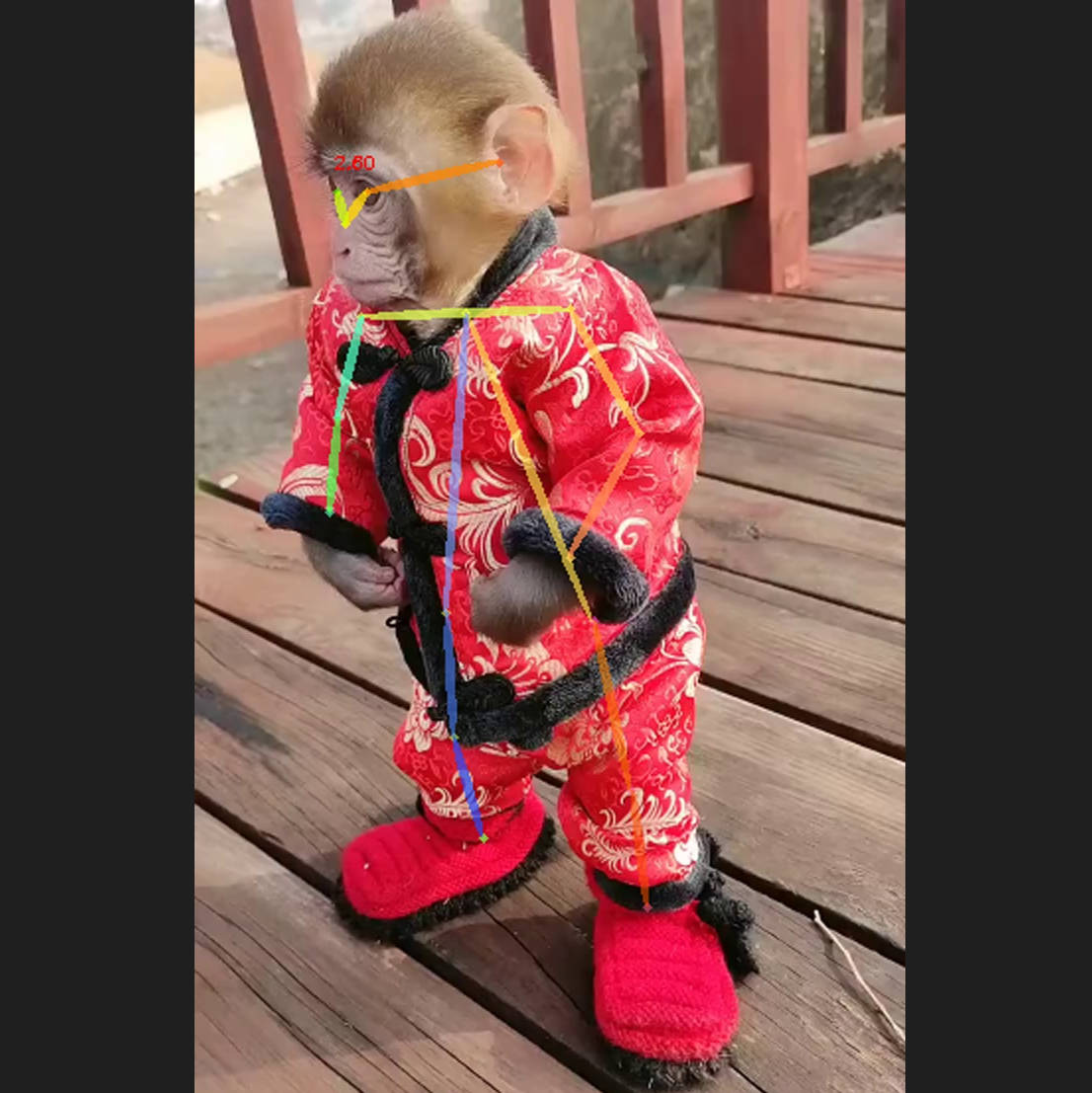}} &
{\includegraphics[width=0.18\textwidth]{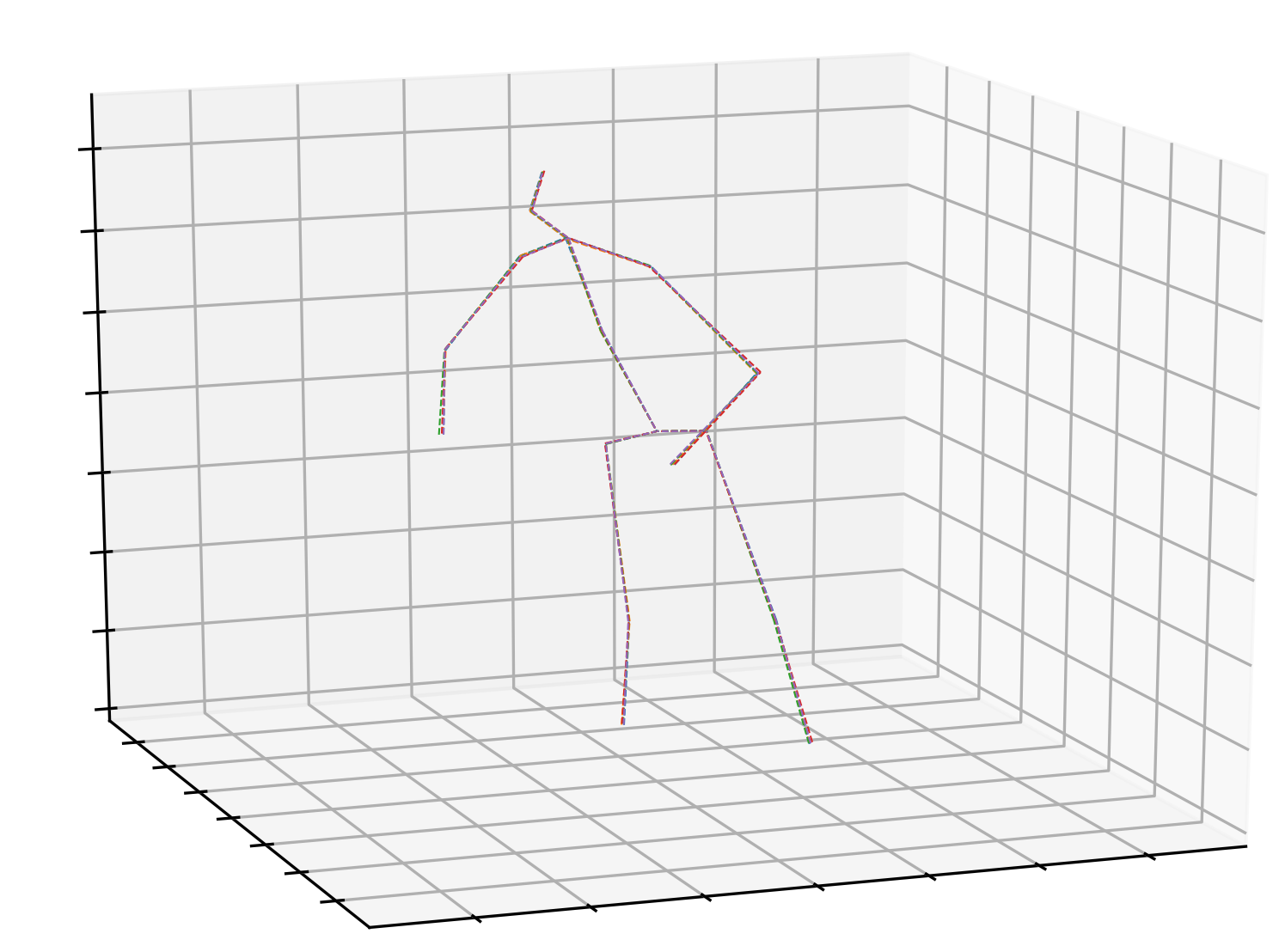}} &
{\includegraphics[width=0.12\textwidth]{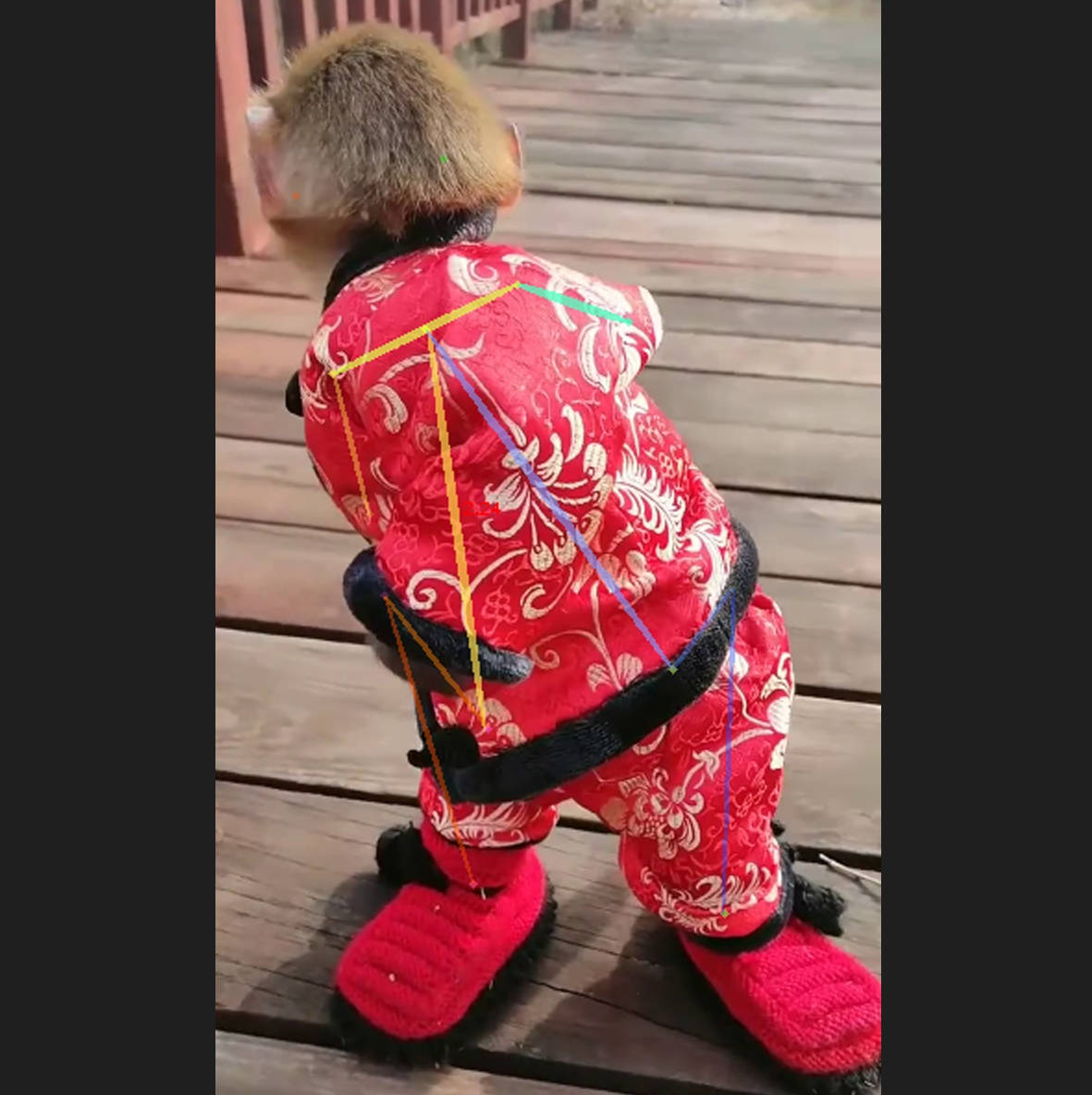}} &
{\includegraphics[width=0.18\textwidth]{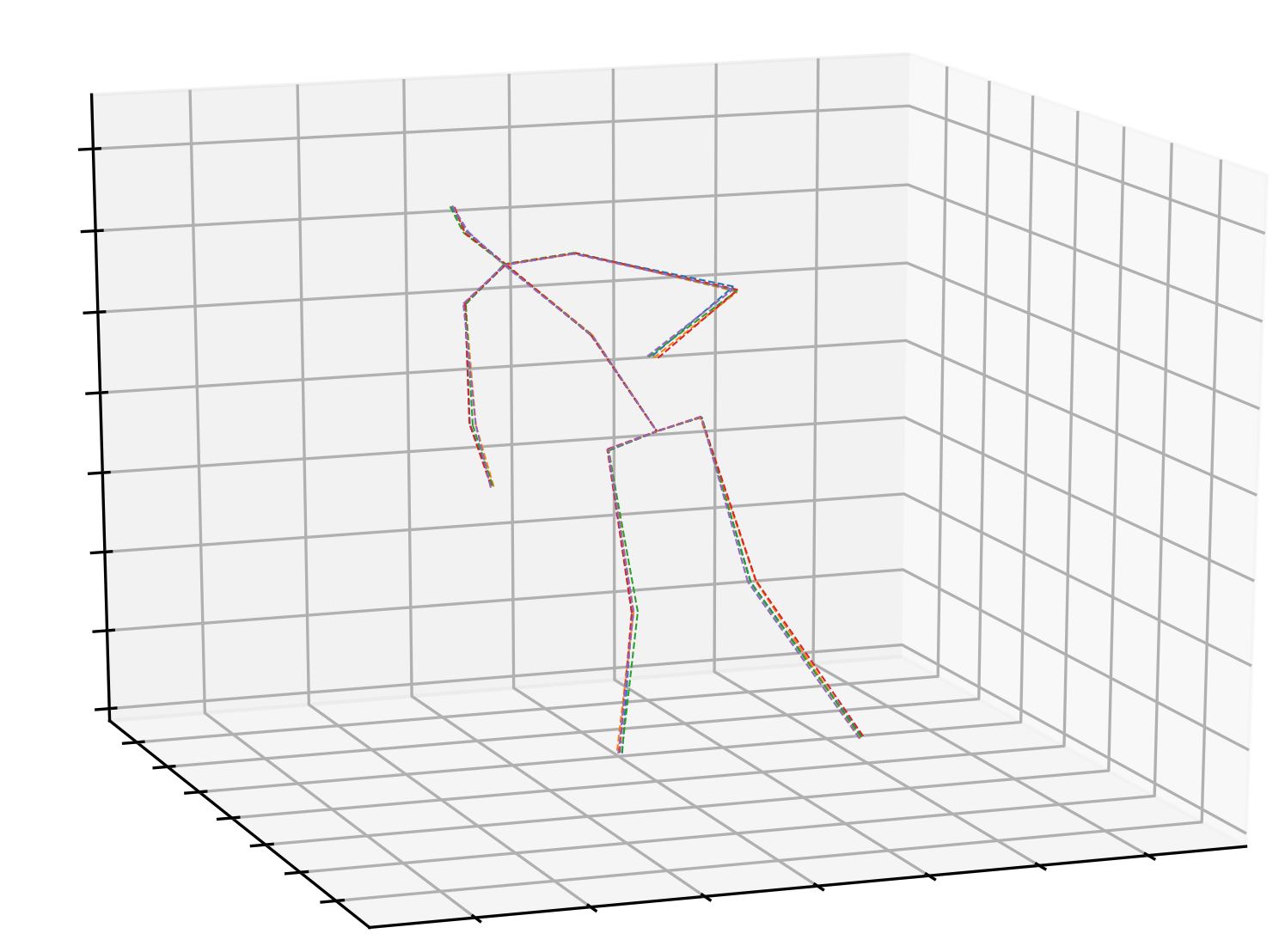}} &
{\includegraphics[width=0.12\textwidth]{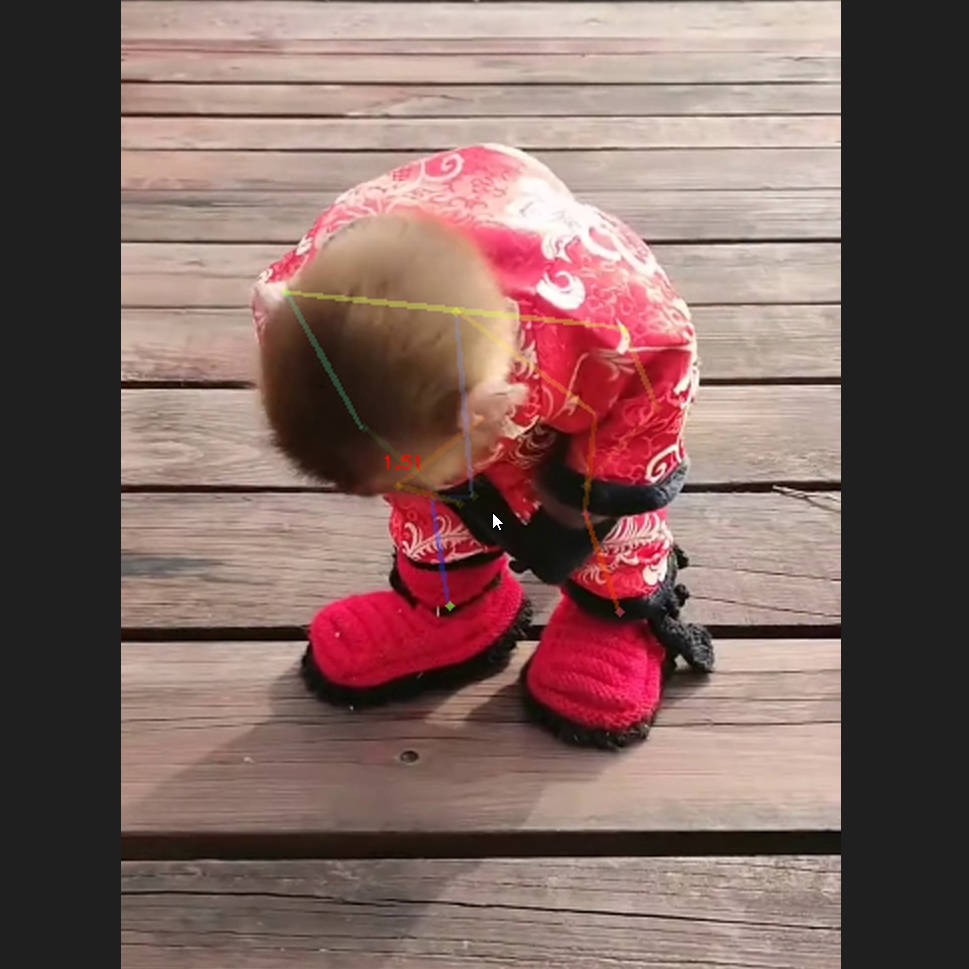}} &
{\includegraphics[width=0.18\textwidth]{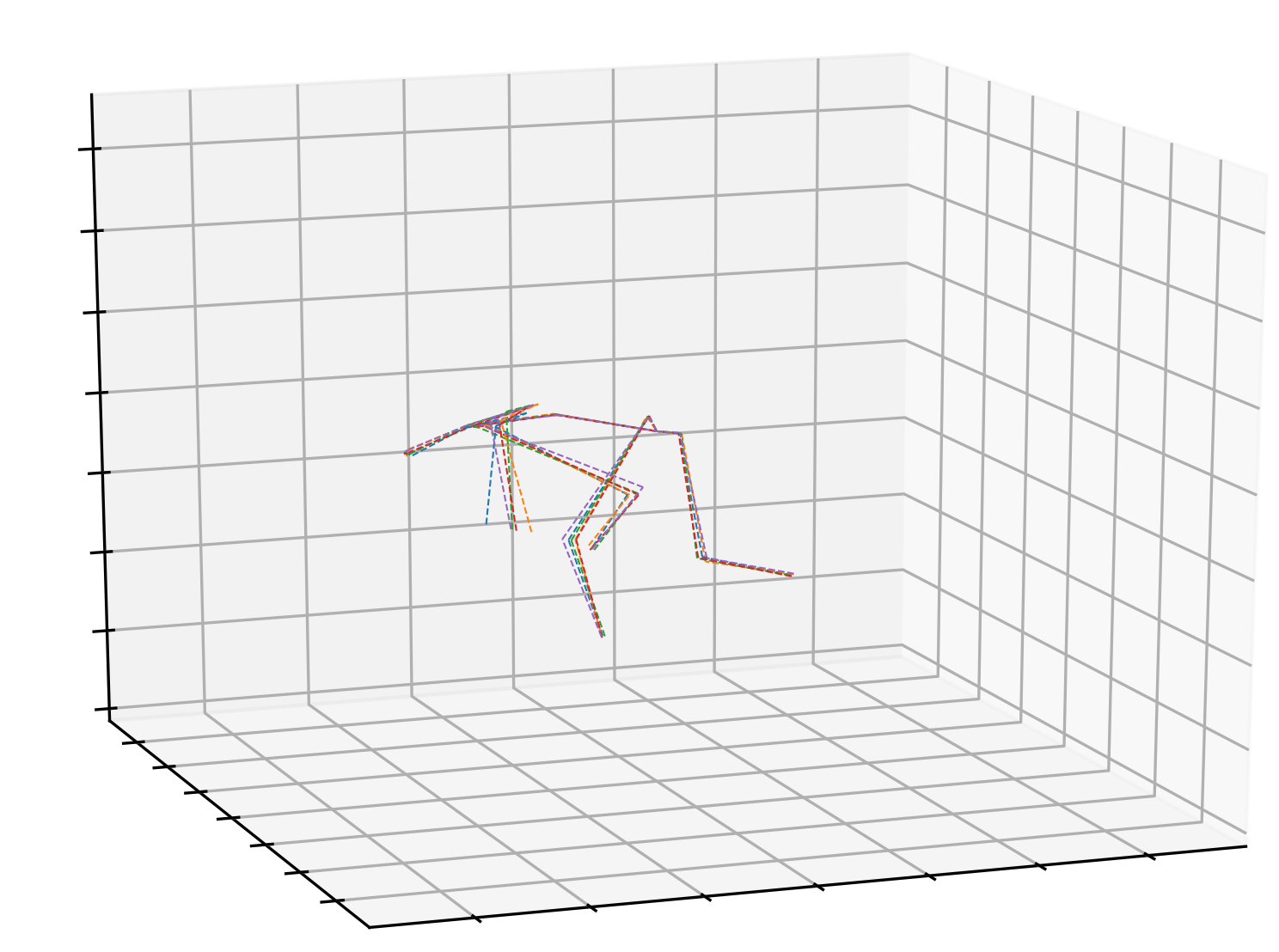}} \\

\end{tabular}
%\vspace*{-0.2cm}
\end{center}
\vspace*{-0.2cm}
\caption{Qualitative results of the proposed method on in-the-wild videos. Dashed line: predicted 3D pose hypotheses. Each color represents an individual hypothesis. $H$=5, $K$=5}
%\vspace*{-0.4cm}
\label{fig:in_the_wild}
\end{figure*}

Our method may fail under the previously mentioned cases such as severe occlusions, fast motion, rare poses, \etc. The $\text{3}^\text{rd}$ row of Fig.~\ref{fig:in_the_wild} shows a person walking through the woods. Most of her body is occluded and our approach cannot obtain a very accurate result in this case. Besides, the person in the $\text{6}^\text{th}$ row of Fig.~\ref{fig:in_the_wild} raises his hand above his head, a movement that is rare in the training set. Our method fails to generalize to this pose and therefore produces incorrect predictions.

\clearpage

\clearpage
%\end{appendices}

{\small
\bibliographystyle{ieee_fullname}
\bibliography{ref}
}

\end{document}

%% file: table/h36m_new.tex
\begin{center}
  %\centering
  %\renewcommand\tabcolsep{4.0pt}
%\vspace{-0.5cm}
\renewcommand{\arraystretch}{0.7}
\resizebox{\textwidth}{!}{\begin{tabular}{cx{45}|ccccccccccccccc|c}
\hline
\noalign{\smallskip}
%\rowcolor[HTML]{DADADA}
\multicolumn{18}{c}{\textbf{\large{Deterministic Methods}}}\\
\noalign{\smallskip}
\hline
\noalign{\smallskip}
\multicolumn{2}{c|}{MPJPE} & Dir. & Disc. & Eat & Greet & Phone & Photo & Pose & Pur. & Sit & SitD. & Smoke & Wait & WalkD. & Walk & WalkT. & Avg \\
\noalign{\smallskip}
\hline
\noalign{\smallskip}
TCN~\cite{pavllo20193d} ($N$=243)*& CVPR'19&45.2&46.7&43.3&45.6&48.1&55.1&44.6&44.3&57.3&65.8&47.1&44.0&49.0&32.8&33.9&46.8 \\
% Lin \textit{et al.}~\cite{lin2019trajectory} BMVC'19 ($N$=50)& 42.5&44.8&42.6&44.2&48.5&57.1&42.6&41.4&56.5&64.5&47.4&43.0&48.1&33.0&35.1&46.6 \\
% Xu \textit{et al.}~\cite{xu2020deep} CVPR'20 ($N$=9)
% &\textbf{37.4}&43.5&42.7&42.7&46.6&59.7&41.3&45.1&\textbf{52.7}&60.2&45.8&43.1&47.7&33.7&37.1&45.6\\
% Wang \textit{et al.}~\cite{wang2020motion} ECCV'20 ($N$=96)
% &41.3&43.9&44.0&42.2&48.0&57.1&42.2&43.2&57.3&61.3&47.0&43.5&47.0&32.6&31.8&45.6\\
% Attn-TCN~\cite{liu2020attention} CVPR'20 ($N$=243)& 41.8&44.8&41.1&44.9&47.4&54.1&43.4&42.2&56.2&63.6&45.3&43.5&45.3&31.3&32.2&45.1 \\
SRNet~\cite{zeng2020srnet} ($N$=243)*& ECCV'20& 46.6&47.1&43.9&41.6&45.8&49.6&46.5&40.0&53.4&61.1&46.1&42.6&43.1&31.5&32.6&44.8 \\
% Zeng \textit{et al.}~\cite{zeng2021learning} ICCV'21 ($N$=9)
% &-&-&-&-&-&-&-&-&-&-&-&-&-&-&-&45.7\\
PoseFormer~\cite{zheng20213d} ($N$=81)*& ICCV'21 &
41.5&44.8&39.8&42.5&46.5&51.6&42.1&42.0&53.3&60.7&45.5&43.3&46.1&31.8&32.2&44.3\\
RIE~\cite{shan2021improving} ($N$=243)*&MM'21 &
40.8&44.5&41.4&42.7&46.3&55.6&41.8&41.9&53.7&60.8&45.0&41.5&44.8&30.8&31.9&44.3\\
Anatomy~\cite{chen2021anatomy} ($N$=243)*&TCSVT'21 &
41.4&43.5&40.1&42.9&46.6&51.9&41.7&42.3&53.9&60.2&45.4&41.7&46.0&31.5&32.7&44.1\\
U-CDGCN~\cite{hu2021conditional} ($N$=96)*&MM'21 &
38.0&43.3&39.1&\textcolor{blue}{39.4}&45.8&53.6&41.4&41.4&55.5&61.9&44.6&41.9&44.5&31.6&29.4&43.4\\
Ray3D~\cite{zhan2022ray3d} ($N$=9)*&CVPR'22 &
44.7&48.7&48.7&48.4&51.0&59.9&46.8&46.9&58.7&61.7&50.2&46.4&51.5&38.6&41.8&49.7\\
STE~\cite{li2022exploiting} ($N$=351)*&TMM'22 &
39.9&43.4&40.0&40.9&46.4&50.6&42.1&\textcolor{blue}{39.8}&55.8&61.6&44.9&43.3&44.9&29.9&30.3&43.6\\
P-STMO~\cite{shan2022p} ($N$=243)*&ECCV'22 &
38.9&42.7&40.4&41.1&45.6&49.7&40.9&39.9&55.5&59.4&44.9&42.2&42.7&29.4&29.4&42.8\\
MixSTE~\cite{zhang2022mixste} ($N$=243)*$\ddagger$&CVPR'22 &
37.9&\textcolor{blue}{40.1}&37.5&\textcolor{blue}{39.4}&43.3&50.0&\textcolor{blue}{39.8}&39.9&52.5&56.6&42.4&40.1&40.5&\textcolor{blue}{27.6}&\textcolor{blue}{27.7}&41.0\\
MixSTE~\cite{zhang2022mixste} ($N$=243)*&CVPR'22 &
\textcolor{red}{37.6}&40.9&37.3&39.7&42.3&49.9&40.1&\textcolor{blue}{39.8}&51.7&55.0&42.1&\textcolor{blue}{39.8}&41.0&27.9&27.9&40.9\\
DUE~\cite{zhang2022uncertainty} ($N$=300)*&MM'22 &
37.9&41.9&\textcolor{blue}{36.8}&39.5&\textcolor{red}{40.8}&\textcolor{blue}{49.2}&40.1&40.7&\textcolor{red}{47.9}&\textcolor{red}{53.3}&\textcolor{red}{40.2}&41.1&\textcolor{blue}{40.3}&30.8&28.6&\textcolor{blue}{40.6}\\
\bestcell{D3DP ($N$=243, $H$=1, $K$=1)*}&\bestcell{}&
\bestcell{\textcolor{blue}{37.7}}&\bestcell{\textcolor{red}{39.9}}&\bestcell{\textcolor{red}{35.7}}&\bestcell{\textcolor{red}{38.2}}&\bestcell{\textcolor{blue}{41.9}}&\bestcell{\textcolor{red}{48.8}}&\bestcell{\textcolor{red}{39.5}}&\bestcell{\textcolor{red}{38.3}}&\bestcell{\textcolor{blue}{50.5}}&\bestcell{\textcolor{blue}{53.9}}&\bestcell{\textcolor{blue}{41.6}}&\bestcell{\textcolor{red}{39.4}}&\bestcell{\textcolor{red}{39.8}}&\bestcell{\textcolor{red}{27.4}}&\bestcell{\textcolor{red}{27.5}}&\bestcell{\textcolor{red}{40.0}}\\

\noalign{\smallskip}
\hline
\hline
\noalign{\smallskip}
%\rowcolor[HTML]{DADADA}
\multicolumn{18}{c}{\textbf{\large{Probabilistic Methods}}}\\
\noalign{\smallskip}
\hline
\noalign{\smallskip}
\multicolumn{2}{c|}{MPJPE}& Dir. & Disc. & Eat & Greet & Phone & Photo & Pose & Pur. & Sit & SitD. & Smoke & Wait & WalkD. & Walk & WalkT. & Avg \\
\noalign{\smallskip}
\hline
\noalign{\smallskip}

CVAE~\cite{sharma2019monocular} ($N$=1, $H$=200, P-Agg)&ICCV'19 &
48.6&54.5&54.2&55.7&62.6&72.0&50.5&54.3&70.0&78.3&58.1&55.4&61.4&45.2&49.7&58.0\\
GAN~\cite{li2020weakly} ($N$=1, $H$=10, P-Agg)&BMVC'20 &
67.9&75.5&71.8&81.8&81.4&93.7&75.2&81.3&88.8&114.1&75.9&79.1&83.3&74.3&79.0&81.1\\
GraphMDN~\cite{oikarinen2021graphmdn} ($N$=1, $H$=5, P-Agg)&IJCNN'21& 
51.9&56.1&55.3&58.0&63.5&75.1&53.3&56.5&69.4&92.7&60.1&58.0&65.5&49.8&53.6&61.3\\
NF~\cite{wehrbein2021probabilistic} ($N$=1, $H$=1, P-Agg)&ICCV'21 &
52.4&60.2&57.8&57.4&65.7&74.1&56.2&59.1&69.3&78.0&61.2&63.7&67.0&50.0&54.9&61.8\\
MHFormer~\cite{li2022mhformer} ($N$=351, $H$=3, P-Agg)*&CVPR'22 &
39.2&43.1&40.1&40.9&44.9&51.2&40.6&41.3&53.5&60.3&43.7&41.1&43.8&29.8&30.6&43.0
\\
\bestcell{D3DP ($N$=243, $H$=1, $K$=1, P-Agg)*}&\bestcell{}&
\bestcell{37.7}&\bestcell{39.9}&\bestcell{\textcolor{blue}{35.7}}&\bestcell{38.2}&\bestcell{41.9}&\bestcell{48.8}&\bestcell{39.5}&\bestcell{38.3}&\bestcell{50.5}&\bestcell{53.9}&\bestcell{41.6}&\bestcell{\textcolor{blue}{39.4}}&\bestcell{39.8}&\bestcell{\textcolor{blue}{27.4}}&\bestcell{27.5}&\bestcell{40.0}
\\
\bestcell{D3DP ($N$=243, $H$=20, $K$=10, P-Agg)*}&\bestcell{}&
\bestcell{\textcolor{blue}{37.6}}&\bestcell{\textcolor{blue}{39.7}}&\bestcell{35.8}&\bestcell{\textcolor{blue}{38.0}}&\bestcell{\textcolor{blue}{41.7}}&\bestcell{\textcolor{blue}{48.7}}&\bestcell{\textcolor{blue}{39.4}}&\bestcell{\textcolor{blue}{38.2}}&\bestcell{\textcolor{blue}{50.3}}&\bestcell{\textcolor{blue}{53.3}}&\bestcell{\textcolor{blue}{41.4}}&\bestcell{\textcolor{blue}{39.4}}&\bestcell{\textcolor{blue}{39.7}}&\bestcell{\textcolor{blue}{27.4}}&\bestcell{\textcolor{blue}{27.3}}&\bestcell{\textcolor{blue}{39.9}}
\\
\bestcell{D3DP ($N$=243, $H$=20, $K$=10, J-Agg)*}&\bestcell{}&
\bestcell{\textcolor{red}{37.3}}&\bestcell{\textcolor{red}{39.5}}&\bestcell{\textcolor{red}{35.6}}&\bestcell{\textcolor{red}{37.8}}&\bestcell{\textcolor{red}{41.3}}&\bestcell{\textcolor{red}{48.2}}&\bestcell{\textcolor{red}{39.1}}&\bestcell{\textcolor{red}{37.6}}&\bestcell{\textcolor{red}{49.9}}&\bestcell{\textcolor{red}{52.8}}&\bestcell{\textcolor{red}{41.2}}&\bestcell{\textcolor{red}{39.2}}&\bestcell{\textcolor{red}{39.4}}&\bestcell{\textcolor{red}{27.2}}&\bestcell{\textcolor{red}{27.1}}&\bestcell{\textcolor{red}{39.5}}
\\

\noalign{\smallskip}
\hline
\noalign{\smallskip}

MDN~\cite{li2019generating} ($N$=1, $H$=5, P-Best$\sharp$)&CVPR'19 &
43.8&48.6&49.1&49.8&57.6&64.5&45.9&48.3&62.0&73.4&54.8&50.6&56.0&43.4&45.5&52.7\\
CVAE~\cite{sharma2019monocular} ($N$=1, $H$=200, P-Best$\sharp$)&ICCV'19 &
37.8&43.2&43.0&44.3&51.1&57.0&39.7&43.0&56.3&64.0&48.1&45.4&50.4&37.9&39.9&46.8\\
GAN~\cite{li2020weakly} ($N$=1, $H$=10, P-Best$\sharp$)&BMVC'20& 
62.0&69.7&64.3&73.6&75.1&84.8&68.7&75.0&81.2&104.3&70.2&72.0&75.0&67.0&69.0&73.9\\
GraphMDN~\cite{oikarinen2021graphmdn} ($N$=1, $H$=200, P-Best$\sharp$)&IJCNN'21& 
40.0&43.2&41.0&43.4&50.0&53.6&40.1&41.4&52.6&67.3&48.1&44.2&49.0&39.5&40.2&46.2\\
NF~\cite{wehrbein2021probabilistic} ($N$=1, $H$=200, P-Best$\sharp$)&ICCV'21 &
38.5&42.5&39.9&41.7&46.5&51.6&39.9&40.8&\textcolor{blue}{49.5}&56.8&45.3&46.4&46.8&37.8&40.4&44.3\\

\bestcell{D3DP ($N$=243, $H$=1, $K$=1, P-Best$\sharp$)*}&\bestcell{}&
\bestcell{37.7}&\bestcell{39.9}&\bestcell{35.7}&\bestcell{38.2}&\bestcell{41.9}&\bestcell{48.8}&\bestcell{39.5}&\bestcell{38.3}&\bestcell{50.5}&\bestcell{53.9}&\bestcell{41.6}&\bestcell{39.4}&\bestcell{39.8}&\bestcell{27.4}&\bestcell{27.5}&\bestcell{40.0}
\\

\bestcell{D3DP ($N$=243, $H$=20, $K$=10, P-Best$\sharp$)*}&\bestcell{}&
\bestcell{\textcolor{blue}{37.3}}&\bestcell{\textcolor{blue}{39.4}}&\bestcell{\textcolor{blue}{35.4}}&\bestcell{\textcolor{blue}{37.8}}&\bestcell{\textcolor{blue}{41.3}}&\bestcell{\textcolor{blue}{48.1}}&\bestcell{\textcolor{blue}{39.0}}&\bestcell{\textcolor{blue}{37.9}}&\bestcell{49.8}&\bestcell{\textcolor{blue}{52.8}}&\bestcell{\textcolor{blue}{41.1}}&\bestcell{\textcolor{blue}{39.0}}&\bestcell{\textcolor{blue}{39.4}}&\bestcell{\textcolor{blue}{27.3}}&\bestcell{\textcolor{blue}{27.2}}&\bestcell{\textcolor{blue}{39.5}}
\\
\bestcell{D3DP ($N$=243, $H$=20, $K$=10, J-Best$\sharp$)*}&\bestcell{}&
\bestcell{\textcolor{red}{33.0}}&\bestcell{\textcolor{red}{34.8}}&\bestcell{\textcolor{red}{31.7}}&\bestcell{\textcolor{red}{33.1}}&\bestcell{\textcolor{red}{37.5}}&\bestcell{\textcolor{red}{43.7}}&\bestcell{\textcolor{red}{34.8}}&\bestcell{\textcolor{red}{33.6}}&\bestcell{\textcolor{red}{45.7}}&\bestcell{\textcolor{red}{47.8}}&\bestcell{\textcolor{red}{37.0}}&\bestcell{\textcolor{red}{35.0}}&\bestcell{\textcolor{red}{35.0}}&\bestcell{\textcolor{red}{24.3}}&\bestcell{\textcolor{red}{24.1}}&\bestcell{\textcolor{red}{35.4}}
\\
\noalign{\smallskip}

\hline
\end{tabular}}
\end{center}

%% file: table/3dhp_new.tex
\begin{center}
  %\centering
  %\renewcommand\tabcolsep{4.0pt}
%\vspace{-0.1cm}
%\vspace{-0.4cm}
\renewcommand{\arraystretch}{0.7}
\resizebox{\linewidth}{!}{\begin{tabular}{cx{45}|ccc}
\hline\noalign{\smallskip}
\multicolumn{2}{c|}{Method}&PCK$\uparrow$&AUC$\uparrow$&MPJPE$\downarrow$\\
\noalign{\smallskip}
\hline
\noalign{\smallskip}
U-CDGCN~\cite{hu2021conditional} ($N$=96)$\triangledown$& MM'21&97.9&69.5&42.5\\
P-STMO~\cite{shan2022p} ($N$=81)$\triangledown$ & ECCV'22&97.9&75.8&32.2\\
\noalign{\smallskip}
\hline
\noalign{\smallskip}
% Mehta \textit{et al.}~\cite{mehta2017monocular} 3DV'17 ($N$=1)&75.7&39.3&117.6\\
TCN~\cite{pavllo20193d} ($N$=81)& CVPR'19&86.0&51.9&84.0\\
% Lin \textit{et al.}~\cite{lin2019trajectory} BMVC'19 ($N$=25)&83.6&51.4&79.8\\
% Zeng \textit{et al.}~\cite{zeng2020srnet} ECCV'20 ($N$=1)&77.6&43.8&-\\
U-GCN~\cite{wang2020motion} ($N$=96)& ECCV'20&86.9&62.1&68.1\\
Anatomy~\cite{chen2021anatomy} ($N$=81)& TCSVT'21&87.9&54.0&78.8\\
PoseFormer~\cite{zheng20213d} ($N$=9)& ICCV'21&88.6&56.4&77.1\\
MHFormer~\cite{li2022mhformer} ($N$=9, $H$=3, P-Agg)& CVPR'22&93.8&63.3&58.0\\
MixSTE~\cite{zhang2022mixste} ($N$=27)& CVPR'22&94.4&66.5&54.9\\

MixSTE~\cite{zhang2022mixste} ($N$=243)$\ddagger$& CVPR'22&\textcolor{blue}{96.9}&75.8&35.4\\
\bestcell{D3DP ($N$=243, $H$=1, $K$=1)}&\bestcell{}&\bestcell{\textcolor{red}{97.7}}&\bestcell{77.8}&\bestcell{30.2}\\
\bestcell{D3DP ($N$=243, $H$=20, $K$=20, P-Agg)}&\bestcell{}&\bestcell{\textcolor{red}{97.7}}&\bestcell{\textcolor{blue}{78.0}}&\bestcell{\textcolor{blue}{30.0}}\\
\bestcell{D3DP ($N$=243, $H$=20, $K$=20, J-Agg)}&\bestcell{}&\bestcell{\textcolor{red}{97.7}}&\bestcell{\textcolor{red}{78.2}}&\bestcell{\textcolor{red}{29.7}}\\

\noalign{\smallskip}
\hline
\noalign{\smallskip}

\bestcell{D3DP ($N$=243, $H$=20, $K$=20, P-Best$\sharp$)}&\bestcell{}&\bestcell{\textcolor{blue}{97.8}}&\bestcell{\textcolor{blue}{78.2}}&\bestcell{\textcolor{blue}{29.8}}\\
\bestcell{D3DP ($N$=243, $H$=20, $K$=20, J-Best$\sharp$)}&\bestcell{}&\bestcell{\textcolor{red}{98.0}}&\bestcell{\textcolor{red}{79.1}}&\bestcell{\textcolor{red}{28.1}}\\
\noalign{\smallskip}
\hline
\end{tabular}}
\end{center}
%\vspace{-0.6cm}

%% file: table/ablation_short.tex
\newcommand\tr{\rule{0pt}{10pt}}
\newcommand\br{\rule[-3.3pt]{0pt}{3.3pt}}
%#################################################
% Component
%#################################################
\subfloat[
Effectiveness of components.
\label{tab:component}
]{
\centering
\begin{minipage}{0.44\linewidth}{\begin{center}
\tablestyle{2pt}{0.7}
%\begin{tabular}{x{20}x{20}x{10}x{25}x{30}}
\begin{tabular}{ccccc}
\shline
\noalign{\vskip 1pt}
Diffusion & JPMA & $H$ & Setting &MPJPE$\downarrow$ \\
\noalign{\vskip 1pt}
\shline
\noalign{\vskip 1pt}
&&  1 &N/A&41.0 \\
\checkmark && 1&N/A&40.0 \\
\checkmark && 20&P-Agg&39.9 \\
\checkmark &\checkmark&20&J-Agg& 39.5 \\
\noalign{\vskip 1pt}
\shline
\end{tabular}
\end{center}}
\end{minipage}
}
\hspace{0.05\linewidth}
%#################################################
% Multi-hypothesis aggregation
%#################################################
\subfloat[
Multi-hypothesis aggregation.
\label{tab:MHA}
]{
\begin{minipage}{0.44\linewidth}{
\begin{center}
\tablestyle{2pt}{0.7}
%\begin{tabular}{x{28}x{40}x{30}}
\begin{tabular}{ccc}
\shline
\noalign{\vskip 1pt}
Level & Method &MPJPE$\downarrow$ \\
\noalign{\vskip 1pt}
\shline
\noalign{\vskip 1pt}
pose & average &39.9 \\
pose & MLPs &42.5 \\
joint & MLPs &41.6 \\
pose & reproj. &39.7 \\
joint& reproj. & 39.5\\
%joint & JPMA-var. &39.5 \\
\noalign{\vskip 1pt}
\shline
\end{tabular}
\end{center}
}
\end{minipage}
}
% \hspace{0.01\linewidth}
% %#################################################
% % speed 
% %#################################################
% \subfloat[
% Accuracy\&speed. Setting: J-Agg. MACs is averaged over each frame. 
% \label{tab:speed}
% ]{
% \begin{minipage}{0.48\linewidth}{\begin{center}
% \tablestyle{1.5pt}{1.05}
% %\begin{tabular}{m{20pt}m{10pt}m{10pt}m{30pt}m{30pt}m{45pt}m{45pt}m{25pt}}
% \begin{tabular}{>{\centering\arraybackslash}m{20pt}>{\centering\arraybackslash}m{10pt}>{\centering\arraybackslash}m{10pt}>{\centering\arraybackslash}m{30pt}>{\centering\arraybackslash}m{35pt}>{\centering\arraybackslash}m{42pt}>{\centering\arraybackslash}m{42pt}>{\centering\arraybackslash}m{25pt}}

% %\begin{tabular}{cccccccc}
% \shline
% %\br{\# boxes}\tr & step & AP & AP$_{50}$ & AP$_{75}$ & FPS \\
% D3DP & $H$ & $K$ &MPJPE$\downarrow$ & Params(M)& Train. MACs(G)&Infer. MACs(G) & Infer. FPS \\
% \shline
% &1&1&41.0&33.6&0.57&1.14&4547 \\
% \checkmark&1&1&40.0&34.6&0.57&1.14&4364 \\
% \checkmark&5&1&39.9&34.6&0.57&5.70&886 \\
% \checkmark&1&5&40.0&34.6&0.57&5.70&1018 \\
% \checkmark&5&5&39.7&34.6&0.57&28.5&204 \\
% \bestcell{\checkmark}&\bestcell{20}&\bestcell{10}&\bestcell{39.5}&\bestcell{34.6}&\bestcell{0.57}&\bestcell{228.2}&\bestcell{29} \\

% \shline
% \end{tabular}
% \end{center}}
% \end{minipage}
% }

%% file: table/ablation_speed.tex
\begin{center}
  %\centering
  %\renewcommand\tabcolsep{4.0pt}
%\vspace{-0.1cm}
%\vspace{-0.4cm}
\renewcommand{\arraystretch}{0.7}
\resizebox{\linewidth}{!}{\begin{tabular}{>{\centering\arraybackslash}m{60pt}>{\centering\arraybackslash}m{8pt}>{\centering\arraybackslash}m{16pt}>{\centering\arraybackslash}m{28pt}>{\centering\arraybackslash}m{35pt}>{\centering\arraybackslash}m{35pt}>{\centering\arraybackslash}m{35pt}>{\centering\arraybackslash}m{18pt}}
\shline
\noalign{\vskip 1pt}
%\br{\# boxes}\tr & step & AP & AP$_{50}$ & AP$_{75}$ & FPS \\
Method & $H$ & $K$ &MPJPE$\downarrow$ & Params(M)& Train. MACs(G)&Infer. MACs(G) & Infer. FPS \\
\noalign{\vskip 1pt}
\shline
\noalign{\vskip 1pt}
PoseFormer~\cite{zheng20213d}&1&N/A&44.3&9.5&0.81&1.62&1952\\
MHFormer~\cite{li2022mhformer}&3&N/A&43.1&24.7&4.81&9.62&598\\
P-STMO~\cite{shan2022p}&1&N/A&42.8&6.7&0.87&1.74&3040\\
MixSTE~\cite{zhang2022mixste}&1&N/A&41.0&33.6&0.57&1.14&4547 \\
D3DP&1&1&40.0&34.6&0.57&1.14&4364 \\
% \checkmark&5&1&39.9&34.6&0.57&5.70&886 \\
% \checkmark&1&5&40.0&34.6&0.57&5.70&1018 \\
D3DP&5&5&39.7&34.6&0.57&28.5&204 \\
D3DP&20&10&39.5&34.6&0.57&228.2&29 \\

\noalign{\vskip 1pt}
\shline
\end{tabular}}
\end{center}
%\vspace{-0.6cm}

%% file: algorithms/training.tex
\small
\caption{\small D3DP Training 
}
\label{alg:train}
\algcomment{\fontsize{7.2pt}{0em}\selectfont \texttt{\emph{alpha\_cumprod}(t)}: cumulative product of $\alpha_i$, \ie, $\prod_{i=1}^t \alpha_i$
}
\definecolor{codeblue}{rgb}{0.25,0.5,0.5}
\definecolor{codegreen}{rgb}{0,0.6,0}
\definecolor{codekw}{RGB}{207,33,46}
\lstset{
  backgroundcolor=\color{white},
  basicstyle=\fontsize{7.5pt}{7.5pt}\ttfamily\selectfont,
  columns=fullflexible,
  breaklines=true,
  captionpos=b,
  commentstyle=\fontsize{7.5pt}{7.5pt}\color{codegreen},
  keywordstyle=\fontsize{7.5pt}{7.5pt}\color{codekw},
  escapechar={|}, 
}
\begin{lstlisting}[language=python, aboveskip=3pt, belowskip=3pt, emph={alpha_cumprod},emphstyle={\emph}]
def training_loss(2dp, 3dp_gt, T):
  # 2dp: [B, N, J, 2], 3dp_gt: [B, N, J, 3]
  # T: maximum number of timesteps
  # B: batch size, N: frame count, J: joint count

  # Signal scaling
  3dp_gt = 3dp_gt * scale  

  # Corrupt 3dp_gt
  t = randint(0, T)|~~~~~~~~~~~~~|# timestep
  noise = normal(mean=0, std=1)  # noise: [B, N, J, 3]
  alpha_cp = alpha_cumprod(t)
  3dp_crpt = sqrt(|~~~~|alpha_cp) * 3dp_gt + 
              |~~|sqrt(1 - alpha_cp) * noise

  # Denoise using a 3d pose estimator as backbone
  3dp_pred = denoiser(3dp_crpt, 2dp, t)

  # Set regression loss
  loss = MSE(3dp_pred, 3dp_gt)
  
  return loss
\end{lstlisting}

%% file: algorithms/inference.tex
\small
\caption{\small D3DP Inference 
}
\label{alg:sample}
\algcomment{\fontsize{7.2pt}{0em}\selectfont \texttt{\emph{linespace}}: generate evenly spaced values
}
\definecolor{codeblue}{rgb}{0.25,0.5,0.5}
\definecolor{codegreen}{rgb}{0,0.6,0}
\definecolor{codekw}{rgb}{0.85, 0.18, 0.50}
\lstset{
  backgroundcolor=\color{white},
  basicstyle=\fontsize{7.5pt}{7.5pt}\ttfamily\selectfont,
  columns=fullflexible,
  breaklines=true,
  captionpos=b,
  commentstyle=\fontsize{7.5pt}{7.5pt}\color{codegreen},
  keywordstyle=\fontsize{7.5pt}{7.5pt}\color{codekw},
  escapechar={|}, 
  escapeinside={<@}{@>}
}

\begin{lstlisting}[language=python, aboveskip=3pt, belowskip=3pt, emph={linespace},emphstyle={\emph}]
def inference(2dp, T, K, H):
  # 2dp: [B, N, J, 2], T: maximum number of timesteps
  # K, H: number of iterations and hypotheses
  
  # Initialize noisy 3d poses: [B, H, N, J, 3] 
  3dp_t = normal(mean=0, std=1)

  # Sample timesteps uniformly
  times = reversed(linespace(0, T, K + 1))
  
  # [(T*(1-k/K), T*(1-(k+1)/K))], k = 0,...,K-1
  time_pairs = list(zip(times[:-1], times[1:])

  for t_now, t_next in zip(time_pairs):   
    # Predict 3dp_0 from 3dp_t
    3dp_0 = denoiser(3dp_t, 2dp, t_now)

    # Diffusion flipping
    # Data augmentation using horizontal flipping
    if augment:
      2dp_hf = horiz_flipping(2dp)
      3dp_t_hf = horiz_flipping(3dp_t)
      3dp_0_hf = denoiser(3dp_t_hf, 2dp_hf, t_now)
      3dp_0 = (3dp_0 + horiz_flipping(3dp_0_hf)) / 2
    
    # Estimate 3dp_t at t_next
    3dp_t = ddim_step(3dp_t, 3dp_0, t_now, t_next)
    
  return 3dp_0
\end{lstlisting}

%\vspace{-10pt}
%<@\textcolor{codekw}{}

% latex
% \small
% \caption{\small D3DP Inference 
% }
% \label{alg:sample}
% \algcomment{\fontsize{7.2pt}{0em}\selectfont \texttt{linespace}: generate evenly spaced values
% }
% \definecolor{codeblue}{rgb}{0.25,0.5,0.5}
% \definecolor{codegreen}{rgb}{0,0.6,0}
% \definecolor{codekw}{rgb}{0.85, 0.18, 0.50}
% \lstset{
%   backgroundcolor=\color{white},
%   basicstyle=\fontsize{7.5pt}{7.5pt}\ttfamily\selectfont,
%   columns=fullflexible,
%   breaklines=true,
%   captionpos=b,
%   commentstyle=\fontsize{7.5pt}{7.5pt}\color{codegreen},
%   keywordstyle=\fontsize{7.5pt}{7.5pt}\color{codekw},
%   escapechar={|}, 
% }
% \begin{lstlisting}[language=python]
% def infer(pose_2d, K, T):

%   # data augmentation using horizontal flipping
%   pose_2d_flip = flipping(pose_2d)
%   return 0
% \end{lstlisting}

%\linespread{0.5}

%% file: table/ablation_D3DP.tex
\newcommand\tr{\rule{0pt}{10pt}}
\newcommand\br{\rule[-3.3pt]{0pt}{3.3pt}}
%#################################################
% Regression target
%#################################################
\subfloat[
Regression target.
\label{tab:reg}
]{
\centering
\begin{minipage}{0.15\linewidth}{\begin{center}
\tablestyle{2pt}{1.05}
% \begin{tabular}{x{43}x{30}}
\begin{tabular}{cc}
\shline
Target & MPJPE$\downarrow$ \\
\shline
noise & 40.2 \\
\bestcell{original data} & \bestcell{40.0} \\
\shline
\end{tabular}
\end{center}}
\end{minipage}
}
\hspace{0.01\linewidth}
%#################################################
% timestep embedding
%#################################################
\subfloat[
Location of the added timestep embedding.
\label{tab:timestep}
]{
\centering
\begin{minipage}{0.16\linewidth}{\begin{center}
\tablestyle{2pt}{1.05}
%\begin{tabular}{x{43}x{30}}
\begin{tabular}{cc}
\shline
Location & MPJPE$\downarrow$ \\
\shline
none    & 40.8 \\
\bestcell{first layer} & \bestcell{40.0} \\
all layers & 40.0 \\
\shline
\end{tabular}
\end{center}}
\end{minipage}
}
\hspace{0.01\linewidth}
%#################################################
% Data augmentation
%#################################################
\subfloat[
Data augmentation.
\label{tab:da}
]{
\begin{minipage}{0.18\linewidth}{
\begin{center}
\tablestyle{1.3pt}{1.05}
%\begin{tabular}{x{65}x{30}}
\begin{tabular}{cc}
\shline
Type & MPJPE$\downarrow$ \\
\shline
none & 40.6 \\
flipping-once & 40.3 \\
\bestcell{diffusion-flipping} & \bestcell{40.0}\\
\shline
\end{tabular}
\end{center}
}
\end{minipage}
}
\hspace{0.01\linewidth}
%#################################################
% 2D conditioning
%#################################################
\subfloat[
2D conditioning. IF: input fusion. EF: embedding fusion. CA: cross attention.
\label{tab:2Dcond}
]{
\centering
\begin{minipage}{0.2\linewidth}{\begin{center}
\tablestyle{2pt}{1.05}
%\begin{tabular}{x{20}x{30}x{30}}
\begin{tabular}{ccc}
\shline
Type & Method & MPJPE$\downarrow$ \\
\shline
\bestcell{IF} & \bestcell{concat}& \bestcell{40.0} \\
EF & concat & 40.9 \\
EF & add & 41.2 \\
EF & CA & 41.1  \\
\shline
\end{tabular}
\end{center}}
\end{minipage}
}
\hspace{0.01\linewidth}
%#################################################
% Maximum number of timesteps
%#################################################
\subfloat[
Maximum number of timesteps.
\label{tab:max_timesteps}
]{
\begin{minipage}{0.12\linewidth}{
\begin{center}
\tablestyle{1.3pt}{1.05}
%\begin{tabular}{x{65}x{30}}
\begin{tabular}{cc}
\shline
$T$ & MPJPE$\downarrow$ \\
\shline
100 & 40.8 \\
500 & 40.2 \\
\bestcell{1000} & \bestcell{40.0} \\
2000 & 40.3 \\
\shline
\end{tabular}
\end{center}
}
\end{minipage}
}

%% file: table/h36m_pmpjpe_new.tex
\begin{center}
  %\centering
  %\renewcommand\tabcolsep{4.0pt}
%\vspace{-0.5cm}
\resizebox{\textwidth}{!}{\begin{tabular}{cx{45}|ccccccccccccccc|c}
\hline
\noalign{\smallskip}
%\rowcolor[HTML]{DADADA}
\multicolumn{18}{c}{\textbf{\large{Deterministic Methods}}}\\
\noalign{\smallskip}
\hline
\noalign{\smallskip}
\multicolumn{2}{c|}{P-MPJPE} & Dir. & Disc. & Eat & Greet & Phone & Photo & Pose & Pur. & Sit & SitD. & Smoke & Wait & WalkD. & Walk & WalkT. & Avg \\
\noalign{\smallskip}
\hline
\noalign{\smallskip}
TCN~\cite{pavllo20193d} ($N$=243)*& CVPR'19 &34.1&36.1&34.4&37.2&36.4&42.2&34.4&33.6&45.0&52.5&37.4&33.8&37.8&25.6&27.3&36.5 \\
% Lin \textit{et al.}~\cite{lin2019trajectory} BMVC'19 ($N$=50)& 42.5&44.8&42.6&44.2&48.5&57.1&42.6&41.4&56.5&64.5&47.4&43.0&48.1&33.0&35.1&46.6 \\
% Xu \textit{et al.}~\cite{xu2020deep} CVPR'20 ($N$=9)
% &\textbf{37.4}&43.5&42.7&42.7&46.6&59.7&41.3&45.1&\textbf{52.7}&60.2&45.8&43.1&47.7&33.7&37.1&45.6\\
% Wang \textit{et al.}~\cite{wang2020motion} ECCV'20 ($N$=96)
% &41.3&43.9&44.0&42.2&48.0&57.1&42.2&43.2&57.3&61.3&47.0&43.5&47.0&32.6&31.8&45.6\\
% Attn-TCN~\cite{liu2020attention} CVPR'20 ($N$=243)& 41.8&44.8&41.1&44.9&47.4&54.1&43.4&42.2&56.2&63.6&45.3&43.5&45.3&31.3&32.2&45.1 \\
% SRNet~\cite{zeng2020srnet} ECCV'20 ($N$=243)*& 46.6&47.1&43.9&41.6&45.8&49.6&46.5&40.0&53.4&61.1&46.1&42.6&43.1&31.5&32.6&44.8 \\
% Zeng \textit{et al.}~\cite{zeng2021learning} ICCV'21 ($N$=9)
% &-&-&-&-&-&-&-&-&-&-&-&-&-&-&-&45.7\\
RIE~\cite{shan2021improving} ($N$=243)*& MM'21&
32.5&36.2&33.2&35.3&35.6&42.1&32.6&31.9&42.6&47.9&36.6&32.1&34.8&24.2&25.8&35.0\\
Anatomy~\cite{chen2021anatomy} ($N$=243)*& TCSVT'21&
32.6&35.1&32.8&35.4&36.3&40.4&32.4&32.3&42.7&49.0&36.8&32.4&36.0&24.9&26.5&35.0\\
PoseFormer~\cite{zheng20213d} ($N$=81)*& ICCV'21&
32.5&34.8&32.6&34.6&35.3&39.5&32.1&32.0&42.8&48.5&34.8&32.4&35.3&24.5&36.0&34.6\\
U-CDGCN~\cite{hu2021conditional} ($N$=96)*& MM'21&
\textcolor{red}{29.8}&34.4&31.9&\textcolor{blue}{31.5}&35.1&40.0&\textcolor{red}{30.3}&30.8&42.6&49.0&35.9&31.8&35.0&25.7&23.6&33.8\\
% Ray3D~\cite{zhan2022ray3d} CVPR'22 ($N=9$)*&
% 44.7&48.7&48.7&48.4&51.0&59.9&46.8&46.9&58.7&61.7&50.2&46.4&51.5&38.6&41.8&49.7\\
STE~\cite{li2022exploiting} ($N$=351)*& TMM'22&
32.7&35.5&32.5&35.4&35.9&41.6&33.0&31.9&45.1&50.1&36.3&33.5&35.1&23.9&25.0&35.2\\
P-STMO~\cite{shan2022p} ($N$=243)*& ECCV'22&
31.3&35.2&32.9&33.9&35.4&39.3&32.5&31.5&44.6&48.2&36.3&32.9&34.4&23.8&23.9&34.4\\
MixSTE~\cite{zhang2022mixste} ($N$=243)*& CVPR'22&
30.8&33.1&30.3&31.8&33.1&39.1&31.1&30.5&42.5&44.5&34.0&30.8&32.7&22.1&22.9&32.6\\
MixSTE~\cite{zhang2022mixste} ($N$=243)*$\ddagger$& CVPR'22&
30.8&\textcolor{blue}{32.7}&30.6&31.9&33.1&\textcolor{blue}{38.6}&\textcolor{blue}{30.8}&\textcolor{blue}{30.4}&42.4&46.4&34.2&\textcolor{blue}{30.7}&\textcolor{blue}{32.3}&\textcolor{blue}{21.8}&\textcolor{blue}{22.6}&32.6\\
DUE~\cite{zhang2022uncertainty} ($N$=300)*& MM'22&
\textcolor{blue}{30.3}&34.6&\textcolor{blue}{29.6}&31.7&\textcolor{red}{31.6}&38.9&31.8&31.9&\textcolor{red}{39.2}&\textcolor{red}{42.8}&\textcolor{red}{32.1}&32.6&\textcolor{red}{31.4}&25.1&23.8&\textcolor{blue}{32.5}\\

\bestcell{D3DP ($N$=243, $H$=1, $K$=1)*}&\bestcell{}&
\bestcell{30.6}&\bestcell{\textcolor{red}{32.5}}&\bestcell{\textcolor{red}{29.1}}&\bestcell{\textcolor{red}{31.0}}&\bestcell{\textcolor{blue}{31.9}}&\bestcell{\textcolor{red}{37.6}}&\bestcell{\textcolor{red}{30.3}}&\bestcell{\textcolor{red}{29.4}}&\bestcell{\textcolor{blue}{40.6}}&\bestcell{\textcolor{blue}{43.6}}&\bestcell{\textcolor{blue}{33.3}}&\bestcell{\textcolor{red}{30.5}}&\bestcell{\textcolor{red}{31.4}}&\bestcell{\textcolor{red}{21.5}}&\bestcell{\textcolor{red}{22.4}}&\bestcell{\textcolor{red}{31.7}}\\

\noalign{\smallskip}
\hline
\hline
\noalign{\smallskip}
%\rowcolor[HTML]{DADADA}
\multicolumn{18}{c}{\textbf{\large{Probabilistic Methods}}}\\
\noalign{\smallskip}
\hline
\noalign{\smallskip}
\multicolumn{2}{c|}{P-MPJPE} & Dir. & Disc. & Eat & Greet & Phone & Photo & Pose & Pur. & Sit & SitD. & Smoke & Wait & WalkD. & Walk & WalkT. & Avg \\
\noalign{\smallskip}
\hline
\noalign{\smallskip}

CVAE~\cite{sharma2019monocular} ($N$=1, $H$=200, P-Agg)& ICCV'19&
35.3&35.9&45.8&42.0&40.9&52.6&36.9&35.8&43.5&51.9&44.3&38.8&45.5&29.4&34.3&40.9\\
GAN~\cite{li2020weakly} ($N$=1, $H$=10, P-Agg)& BMVC'20&
42.1&44.7&45.4&51.0&49.3&51.5&41.2&46.2&57.5&70.8&48.7&44.1&50.8&42.1&43.7&48.7\\
GraphMDN~\cite{oikarinen2021graphmdn} ($N$=1, $H$=5, P-Agg)& IJCNN'21&
39.7&43.4&44.0&46.2&48.8&54.5&39.4&41.1&55.0&69.0&48.0&43.7&49.6&38.4&42.4&46.9\\
NF~\cite{wehrbein2021probabilistic} ($N$=1, $H$=1, P-Agg)& ICCV'21&
37.8&41.7&42.1&41.8&46.5&50.2&38.0&39.2&51.7&61.8&45.4&42.6&45.7&33.7&38.5&43.8\\
MHFormer~\cite{li2022mhformer} ($N$=351, $H$=3, P-Agg)*& CVPR'22&
\textcolor{blue}{31.5}&34.9&32.8&33.6&\textcolor{blue}{35.3}&39.6&32.0&32.2&43.5&48.7&36.4&32.6&34.3&\textcolor{blue}{23.9}&25.1&34.4
\\
\bestcell{D3DP ($N$=243, $H$=1, $K$=1, P-Agg)*}&\bestcell{}&
\bestcell{\textcolor{red}{30.6}}&\bestcell{\textcolor{blue}{32.5}}&\bestcell{\textcolor{red}{29.1}}&\bestcell{\textcolor{blue}{31.0}}&\bestcell{\textcolor{red}{31.9}}&\bestcell{37.6}&\bestcell{\textcolor{blue}{30.3}}&\bestcell{\textcolor{blue}{29.4}}&\bestcell{\textcolor{blue}{40.6}}&\bestcell{43.6}&\bestcell{\textcolor{blue}{33.3}}&\bestcell{\textcolor{blue}{30.5}}&\bestcell{\textcolor{blue}{31.4}}&\bestcell{\textcolor{red}{21.5}}&\bestcell{\textcolor{blue}{22.4}}&\bestcell{\textcolor{blue}{31.7}}\\
\bestcell{D3DP ($N$=243, $H$=20, $K$=10, P-Agg)*}&\bestcell{}&
\bestcell{\textcolor{red}{30.6}}&\bestcell{\textcolor{blue}{32.5}}&\bestcell{\textcolor{red}{29.1}}&\bestcell{\textcolor{red}{30.9}}&\bestcell{\textcolor{red}{31.9}}&\bestcell{\textcolor{blue}{37.5}}&\bestcell{\textcolor{red}{30.2}}&\bestcell{\textcolor{blue}{29.4}}&\bestcell{\textcolor{blue}{40.6}}&\bestcell{\textcolor{blue}{43.4}}&\bestcell{\textcolor{blue}{33.3}}&\bestcell{\textcolor{red}{30.4}}&\bestcell{\textcolor{blue}{31.4}}&\bestcell{\textcolor{red}{21.5}}&\bestcell{\textcolor{blue}{22.4}}&\bestcell{\textcolor{blue}{31.7}}
\\
\bestcell{D3DP ($N$=243, $H$=20, $K$=10, J-Agg)*}&\bestcell{}&
\bestcell{\textcolor{red}{30.6}}&\bestcell{\textcolor{red}{32.4}}&\bestcell{\textcolor{blue}{29.2}}&\bestcell{\textcolor{red}{30.9}}&\bestcell{\textcolor{red}{31.9}}&\bestcell{\textcolor{red}{37.4}}&\bestcell{\textcolor{red}{30.2}}&\bestcell{\textcolor{red}{29.3}}&\bestcell{\textcolor{red}{40.4}}&\bestcell{\textcolor{red}{43.2}}&\bestcell{\textcolor{red}{33.2}}&\bestcell{\textcolor{red}{30.4}}&\bestcell{\textcolor{red}{31.3}}&\bestcell{\textcolor{red}{21.5}}&\bestcell{\textcolor{red}{22.3}}&\bestcell{\textcolor{red}{31.6}}
\\
\noalign{\smallskip}
\hline
\noalign{\smallskip}

MDN~\cite{li2019generating} ($N$=1, $H$=5, P-Best$\sharp$)& CVPR'19&
35.5&39.8&41.3&42.3&46.0&48.9&36.9&37.3&51.0&60.6&44.9&40.2&44.1&33.1&36.9&42.6\\
CVAE~\cite{sharma2019monocular} ($N$=1, $H$=200, P-Best$\sharp$)& ICCV'19&
\textcolor{blue}{27.6}&\textcolor{red}{27.5}&34.9&32.3&33.3&42.7&28.7&\textcolor{blue}{28.0}&\textcolor{red}{36.1}&42.7&36.0&30.7&37.6&24.3&27.1&32.7\\
GAN~\cite{li2020weakly} ($N$=1, $H$=10, P-Best$\sharp$)& BMVC'20&
38.5&41.7&39.6&45.2&45.8&46.5&37.8&42.7&52.4&62.9&45.3&40.9&45.3&38.6&38.4&44.3\\
GraphMDN~\cite{oikarinen2021graphmdn} ($N$=1, $H$=200, P-Best$\sharp$)& IJCNN'21&
30.8&34.7&33.6&34.2&39.6&42.2&31.0&31.9&42.9&53.5&38.1&34.1&38.0&29.6&31.1&36.3\\
NF~\cite{wehrbein2021probabilistic} ($N$=1, $H$=200, P-Best$\sharp$)& ICCV'21&
27.9&31.4&29.7&\textcolor{blue}{30.2}&34.9&37.1&\textcolor{red}{27.3}&28.2&39.0&46.1&34.2&32.3&33.6&26.1&27.5&32.4\\

\bestcell{D3DP ($N$=243, $H$=1, $K$=1, P-Best$\sharp$)*}&\bestcell{}&
\bestcell{30.6}&\bestcell{32.5}&\bestcell{29.1}&\bestcell{31.0}&\bestcell{31.9}&\bestcell{37.6}&\bestcell{30.3}&\bestcell{29.4}&\bestcell{40.6}&\bestcell{43.6}&\bestcell{33.3}&\bestcell{30.5}&\bestcell{31.4}&\bestcell{21.5}&\bestcell{22.4}&\bestcell{31.7}\\

\bestcell{D3DP ($N$=243, $H$=20, $K$=10, P-Best$\sharp$)*}&\bestcell{}&
\bestcell{30.2}&\bestcell{32.1}&\bestcell{\textcolor{blue}{28.8}}&\bestcell{30.4}&\bestcell{\textcolor{blue}{31.5}}&\bestcell{\textcolor{blue}{37.0}}&\bestcell{29.8}&\bestcell{28.9}&\bestcell{39.9}&\bestcell{\textcolor{blue}{42.4}}&\bestcell{\textcolor{blue}{32.8}}&\bestcell{\textcolor{blue}{30.0}}&\bestcell{\textcolor{blue}{30.9}}&\bestcell{\textcolor{blue}{21.3}}&\bestcell{\textcolor{blue}{22.1}}&\bestcell{\textcolor{blue}{31.2}}
\\
\bestcell{D3DP ($N$=243, $H$=20, $K$=10, J-Best$\sharp$)*}&\bestcell{}&
\bestcell{\textcolor{red}{27.5}}&\bestcell{\textcolor{blue}{29.4}}&\bestcell{\textcolor{red}{26.6}}&\bestcell{\textcolor{red}{27.7}}&\bestcell{\textcolor{red}{29.2}}&\bestcell{\textcolor{red}{34.3}}&\bestcell{\textcolor{blue}{27.5}}&\bestcell{\textcolor{red}{26.2}}&\bestcell{\textcolor{blue}{37.3}}&\bestcell{\textcolor{red}{39.0}}&\bestcell{\textcolor{red}{30.3}}&\bestcell{\textcolor{red}{27.7}}&\bestcell{\textcolor{red}{28.2}}&\bestcell{\textcolor{red}{19.6}}&\bestcell{\textcolor{red}{20.3}}&\bestcell{\textcolor{red}{28.7}}
\\
\noalign{\smallskip}

\hline
\end{tabular}}
\end{center}

%% file: table/h36m_gt_new.tex
\begin{center}
  %\centering
  %\renewcommand\tabcolsep{4.0pt}
%\vspace{-0.5cm}
\resizebox{\textwidth}{!}{\begin{tabular}{cx{45}|ccccccccccccccc|c}
\hline
\noalign{\smallskip}
%\rowcolor[HTML]{DADADA}
\multicolumn{18}{c}{\textbf{\large{Deterministic Methods}}}\\
\noalign{\smallskip}
\hline
\noalign{\smallskip}
\multicolumn{2}{c|}{MPJPE} & Dir. & Disc. & Eat & Greet & Phone & Photo & Pose & Pur. & Sit & SitD. & Smoke & Wait & WalkD. & Walk & WalkT. & Avg \\
\noalign{\smallskip}
\hline
\noalign{\smallskip}
TCN~\cite{pavllo20193d} ($N$=243)& CVPR'19 &35.2&40.2&32.7&35.7&38.2&45.5&40.6&36.1&48.8&47.3&37.8&39.7&38.7&27.8&29.5&37.8 \\
% Lin \textit{et al.}~\cite{lin2019trajectory} BMVC'19 ($N$=50)& 42.5&44.8&42.6&44.2&48.5&57.1&42.6&41.4&56.5&64.5&47.4&43.0&48.1&33.0&35.1&46.6 \\
% Xu \textit{et al.}~\cite{xu2020deep} CVPR'20 ($N$=9)
% &\textbf{37.4}&43.5&42.7&42.7&46.6&59.7&41.3&45.1&\textbf{52.7}&60.2&45.8&43.1&47.7&33.7&37.1&45.6\\
% Wang \textit{et al.}~\cite{wang2020motion} ECCV'20 ($N$=96)
% &41.3&43.9&44.0&42.2&48.0&57.1&42.2&43.2&57.3&61.3&47.0&43.5&47.0&32.6&31.8&45.6\\
% Attn-TCN~\cite{liu2020attention} CVPR'20 ($N$=243)& 41.8&44.8&41.1&44.9&47.4&54.1&43.4&42.2&56.2&63.6&45.3&43.5&45.3&31.3&32.2&45.1 \\
SRNet~\cite{zeng2020srnet} ($N$=243)& ECCV'20 &34.8&32.1&28.5&30.7&31.4&36.9&35.6&30.5&38.9&40.5&32.5&31.0&29.9&22.5&24.5&32.0 \\
% Zeng \textit{et al.}~\cite{zeng2021learning} ICCV'21 ($N$=9)
% &-&-&-&-&-&-&-&-&-&-&-&-&-&-&-&45.7\\
Anatomy~\cite{chen2021anatomy} ($N$=243)& TCSVT'21&
-&-&-&-&-&-&-&-&-&-&-&-&-&-&-&32.3\\
PoseFormer~\cite{zheng20213d} ($N$=81)& ICCV'21&
30.0&33.6&29.9&31.0&30.2&33.3&34.8&31.4&37.8&38.6&31.7&31.5&29.0&23.3&23.1&31.3\\
RIE~\cite{shan2021improving} ($N$=243)& MM'21&
29.5&30.8&28.8&29.1&30.7&35.2&31.7&27.8&34.5&36.0&30.3&29.4&28.9&24.1&24.7&30.1\\
% U-CDGCN~\cite{hu2021conditional} ACMMM'21 ($N$=96)*&
% 38.0&43.3&39.1&\textcolor{blue}{39.4}&45.8&53.6&41.4&41.4&55.5&61.9&44.6&41.9&44.5&31.6&29.4&43.4\\
Ray3D~\cite{zhan2022ray3d} ($N$=9)& CVPR'22&
31.2&35.7&34.6&33.6&35.0&37.5&37.2&30.9&42.5&41.3&34.6&36.5&32.0&29.7&28.9&34.4\\
P-STMO~\cite{shan2022p} ($N$=243)& ECCV'22&
28.5&30.1&28.6&27.9&29.8&33.2&31.3&27.8&36.0&37.4&29.7&29.5&28.1&21.0&21.0&29.3\\
STE~\cite{li2022exploiting} ($N$=351)& TMM'22&
27.1&29.4&26.5&27.1&28.6&33.0&30.7&26.8&38.2&34.7&29.1&29.8&26.8&19.1&19.8&28.5\\
DUE~\cite{zhang2022uncertainty} ($N$=300)& MM'22&
22.1&23.1&\textcolor{blue}{20.1}&22.7&21.3&24.1&\textcolor{blue}{23.6}&\textcolor{blue}{21.6}&26.3&24.8&21.7&21.4&21.8&16.7&18.7&22.0\\
MixSTE~\cite{zhang2022mixste} ($N$=243)& CVPR'22&
\textcolor{blue}{21.6}&22.0&20.4&\textcolor{blue}{21.0}&\textcolor{blue}{20.8}&26.3&24.7&21.9&26.9&24.9&\textcolor{blue}{21.2}&21.5&20.8&\textcolor{blue}{14.7}&\textcolor{blue}{15.7}&21.6\\
MixSTE~\cite{zhang2022mixste} ($N$=243)$\ddagger$& CVPR'22&
22.9&\textcolor{blue}{21.7}&21.0&21.4&\textcolor{blue}{20.8}&\textcolor{blue}{23.5}&24.1&21.8&\textcolor{red}{25.3}&\textcolor{red}{23.5}&21.4&\textcolor{blue}{20.1}&\textcolor{blue}{19.5}&15.3&16.6&\textcolor{blue}{21.3}\\

\bestcell{D3DP ($N$=243, $H$=1, $K$=1)}&\bestcell{}&
\bestcell{\textcolor{red}{19.9}}&\bestcell{\textcolor{red}{19.6}}&\bestcell{\textcolor{red}{19.7}}&\bestcell{\textcolor{red}{19.3}}&\bestcell{\textcolor{red}{20.2}}&\bestcell{\textcolor{red}{22.7}}&\bestcell{\textcolor{red}{21.5}}&\bestcell{\textcolor{red}{19.2}}&\bestcell{\textcolor{blue}{25.5}}&\bestcell{\textcolor{blue}{24.0}}&\bestcell{\textcolor{red}{20.1}}&\bestcell{\textcolor{red}{18.9}}&\bestcell{\textcolor{red}{19.0}}&\bestcell{\textcolor{red}{14.0}}&\bestcell{\textcolor{red}{14.5}}&\bestcell{\textcolor{red}{19.9}}\\
\noalign{\smallskip}
\hline
\hline
\noalign{\smallskip}
%\rowcolor[HTML]{DADADA}
\multicolumn{18}{c}{\textbf{\large{Probabilistic Methods}}}\\
\noalign{\smallskip}
\hline
\noalign{\smallskip}
\multicolumn{2}{c|}{MPJPE} & Dir. & Disc. & Eat & Greet & Phone & Photo & Pose & Pur. & Sit & SitD. & Smoke & Wait & WalkD. & Walk & WalkT. & Avg \\
\noalign{\smallskip}
\hline
\noalign{\smallskip}
% MDN~\cite{li2019generating} CVPR'19 ($N$=1, $H$=5, P-Best$\sharp$)$\ddagger$&
% 43.8&48.6&49.1&49.8&57.6&64.5&45.9&48.3&62.0&73.4&54.8&50.6&56.0&43.4&45.5&52.7\\
% CVAE~\cite{sharma2019monocular} ICCV'19 ($N$=1, $H$=200, P-Agg)$\ddagger$&
% 48.6&54.5&54.2&55.7&62.6&72.0&50.5&54.3&70.0&78.3&58.1&55.4&61.4&45.2&49.7&58.0\\
% CVAE~\cite{sharma2019monocular} ICCV'19 ($N$=1, $H$=200, P-Best$\sharp$)$\ddagger$&
% 37.8&43.2&43.0&44.3&51.1&57.0&39.7&43.0&56.3&64.0&48.1&45.4&50.4&37.9&39.9&46.8\\
% GAN~\cite{li2020weakly} BMVC'20 ($N$=1, $H$=10, P-Agg)$\ddagger$&
% 67.9&75.5&71.8&81.8&81.4&93.7&75.2&81.3&88.8&114.1&75.9&79.1&83.3&74.3&79.0&81.1\\

GraphMDN~\cite{oikarinen2021graphmdn} ($N$=1, $H$=5, P-Agg)& IJCNN'21&
33.9&39.9&33.0&35.4&36.8&44.4&38.9&33.0&41.0&50.0&36.4&38.3&37.8&28.2&31.5&37.2\\
MHFormer~\cite{li2022mhformer}($N$=351, $H$=3, P-Agg)& CVPR'22& 
\textcolor{blue}{27.7}&32.1&29.1&28.9&30.0&33.9&33.0&31.2&37.0&39.3&30.3&31.0&29.4&\textcolor{blue}{22.2}&\textcolor{blue}{23.0}&30.5
\\
\bestcell{D3DP ($N$=243, $H$=1, $K$=1, P-Agg)}&\bestcell{}&
\bestcell{\textcolor{red}{19.9}}&\bestcell{19.6}&\bestcell{19.7}&\bestcell{19.3}&\bestcell{20.2}&\bestcell{22.7}&\bestcell{\textcolor{blue}{21.5}}&\bestcell{\textcolor{blue}{19.2}}&\bestcell{25.5}&\bestcell{24.0}&\bestcell{20.1}&\bestcell{\textcolor{blue}{18.9}}&\bestcell{19.0}&\bestcell{\textcolor{red}{14.0}}&\bestcell{\textcolor{red}{14.5}}&\bestcell{19.9}\\

\bestcell{D3DP ($N$=243, $H$=20, $K$=10, P-Agg)}&\bestcell{}&
\bestcell{\textcolor{red}{19.9}}&\bestcell{\textcolor{blue}{19.5}}&\bestcell{\textcolor{blue}{19.6}}&\bestcell{\textcolor{blue}{19.2}}&\bestcell{\textcolor{blue}{20.1}}&\bestcell{\textcolor{blue}{22.4}}&\bestcell{\textcolor{blue}{21.5}}&\bestcell{\textcolor{red}{19.1}}&\bestcell{\textcolor{blue}{25.4}}&\bestcell{\textcolor{blue}{23.7}}&\bestcell{\textcolor{blue}{20.0}}&\bestcell{\textcolor{blue}{18.9}}&\bestcell{\textcolor{blue}{18.8}}&\bestcell{\textcolor{red}{14.0}}&\bestcell{\textcolor{red}{14.5}}&\bestcell{\textcolor{blue}{19.8}}\\
\bestcell{D3DP ($N$=243, $H$=20, $K$=10, J-Agg)}&\bestcell{}&
\bestcell{\textcolor{red}{19.9}}&\bestcell{\textcolor{red}{19.4}}&\bestcell{\textcolor{red}{19.4}}&\bestcell{\textcolor{red}{19.0}}&\bestcell{\textcolor{red}{19.8}}&\bestcell{\textcolor{red}{22.0}}&\bestcell{\textcolor{red}{21.4}}&\bestcell{\textcolor{red}{19.1}}&\bestcell{\textcolor{red}{24.8}}&\bestcell{\textcolor{red}{23.2}}&\bestcell{\textcolor{red}{19.6}}&\bestcell{\textcolor{red}{18.7}}&\bestcell{\textcolor{red}{18.6}}&\bestcell{\textcolor{red}{14.0}}&\bestcell{\textcolor{red}{14.5}}&\bestcell{\textcolor{red}{19.6}}
\\

\noalign{\smallskip}
\hline
\noalign{\smallskip}

GAN~\cite{li2020weakly} ($N$=1, $H$=10, P-Best$\sharp$)& BMVC'20&
54.8&61.9&48.6&63.6&55.8&73.7&59.0&61.3&62.2&85.7&52.8&60.2&57.5&51.3&56.8&60.0\\
GraphMDN~\cite{oikarinen2021graphmdn} ($N$=1, $H$=5, P-Best$\sharp$)& IJCNN'21&
28.9&34.5&28.2&30.2&31.5&38.5&32.3&28.6&35.7&43.3&31.9&32.1&33.3&25.2&27.8&31.8\\
% NF~\cite{wehrbein2021probabilistic} ICCV'21 ($N$=1, $H$=1)$\vartriangle$&
% 52.4&60.2&57.8&57.4&65.7&74.1&56.2&59.1&69.3&78.0&61.2&63.7&67.0&50.0&54.9&61.8\\
% NF~\cite{wehrbein2021probabilistic} ICCV'21 ($N$=1, $H$=200, P-Best$\sharp$)$\vartriangle$&
% 38.5&42.5&39.9&41.7&46.5&51.6&39.9&40.8&49.5&56.8&45.3&46.4&46.8&37.8&40.4&44.3\\

\bestcell{D3DP ($N$=243, $H$=1, $K$=1, P-Best$\sharp$)}&\bestcell{}&
\bestcell{19.9}&\bestcell{19.6}&\bestcell{19.7}&\bestcell{19.3}&\bestcell{20.2}&\bestcell{22.7}&\bestcell{21.5}&\bestcell{19.2}&\bestcell{25.5}&\bestcell{24.0}&\bestcell{20.1}&\bestcell{18.9}&\bestcell{19.0}&\bestcell{14.0}&\bestcell{14.5}&\bestcell{19.9}\\

\bestcell{D3DP ($N$=243, $H$=20, $K$=10, P-Best$\sharp$)}&\bestcell{}&
\bestcell{\textcolor{blue}{19.7}}&\bestcell{\textcolor{blue}{19.3}}&\bestcell{\textcolor{blue}{19.2}}&\bestcell{\textcolor{blue}{18.8}}&\bestcell{\textcolor{blue}{19.6}}&\bestcell{\textcolor{blue}{21.9}}&\bestcell{\textcolor{blue}{21.2}}&\bestcell{\textcolor{blue}{18.6}}&\bestcell{\textcolor{blue}{24.7}}&\bestcell{\textcolor{blue}{23.0}}&\bestcell{\textcolor{blue}{19.4}}&\bestcell{\textcolor{blue}{18.4}}&\bestcell{\textcolor{blue}{18.4}}&\bestcell{\textcolor{blue}{13.9}}&\bestcell{\textcolor{blue}{14.4}}&\bestcell{\textcolor{blue}{19.4}}
\\
\bestcell{D3DP ($N$=243, $H$=20, $K$=10, J-Best$\sharp$)}&\bestcell{}&
\bestcell{\textcolor{red}{18.7}}&\bestcell{\textcolor{red}{18.2}}&\bestcell{\textcolor{red}{18.4}}&\bestcell{\textcolor{red}{17.8}}&\bestcell{\textcolor{red}{18.6}}&\bestcell{\textcolor{red}{20.9}}&\bestcell{\textcolor{red}{20.2}}&\bestcell{\textcolor{red}{17.7}}&\bestcell{\textcolor{red}{23.8}}&\bestcell{\textcolor{red}{21.8}}&\bestcell{\textcolor{red}{18.5}}&\bestcell{\textcolor{red}{17.4}}&\bestcell{\textcolor{red}{17.4}}&\bestcell{\textcolor{red}{13.1}}&\bestcell{\textcolor{red}{13.6}}&\bestcell{\textcolor{red}{18.4}}
\\
\noalign{\smallskip}

\hline
\end{tabular}}
\end{center}